\documentclass[sn-mathphys-num]{sn-jnl}

\usepackage{graphicx}%
\usepackage{multirow}%
\usepackage{amsmath,amssymb,amsfonts}%
\usepackage{amsthm}%
\usepackage{mathrsfs}%
\usepackage[title]{appendix}%
\usepackage{xcolor}%
\usepackage{textcomp}%
\usepackage{manyfoot}%
\usepackage{booktabs}%
\usepackage{algorithm}%
\usepackage{algorithmicx}%
\usepackage{algpseudocode}%
\usepackage{listings}%

\usepackage[utf8]{inputenc} 
\usepackage[T1]{fontenc}    
\usepackage{hyperref}       
\usepackage{url}            
\usepackage{booktabs}       
\usepackage{amsfonts}       
\usepackage{nicefrac}       
\usepackage{microtype}      

\usepackage{framed,multirow}

\usepackage{amssymb, amsmath}
\usepackage{latexsym}
\usepackage{graphicx}
\usepackage{algorithm} 
\usepackage{algpseudocode} 

\usepackage{url}
\usepackage{colortbl}
\usepackage[usenames,dvipsnames]{xcolor}
\definecolor{newcolor}{rgb}{.8,.349,.1}

\usepackage{hyperref}
\usepackage{subcaption}
\usepackage{makecell}
\usepackage{gensymb}
\usepackage{wrapfig}
\usepackage{multirow}
\usepackage{tabulary}
\usepackage[table]{xcolor}

\usepackage{float}
\usepackage{setspace}

\usepackage{soul} 

\usepackage{mathtools}

\DeclarePairedDelimiter{\norm}{\lVert}{\rVert} 
\DeclarePairedDelimiter{\abs}{\lvert}{\rvert}

\theoremstyle{thmstyleone}%
\theoremstyle{thmstyletwo}%

\theoremstyle{thmstylethree}%

\begin{document}

\title[Symmetria]{Symmetria: A Synthetic Dataset for Learning in Point Clouds}


\author*[1]{\fnm{Ivan} \sur{Sipiran}}\email{isipiran@dcc.uchile.cl}
\author[1]{\fnm{Gustavo} \sur{Santelices}}\email{gustavo.santelices@ing.uchile.cl}
\author[1]{\fnm{Lucas} \sur{Oyarzún}}\email{lucas.oyarzun@ing.uchile.cl}
\author[3]{\fnm{Andrea} \sur{Ranieri}}\email{andrea.ranieri@cnr.it}
\author[3]{\fnm{Chiara} \sur{Romanengo}}\email{chiara.romanengo@cnr.it}
\author[3]{\fnm{Silvia} \sur{Biasotti}}\email{silviamaria.biasotti@cnr.it}
\author[3]{\fnm{Bianca} \sur{Falcidieno}}\email{bianca.falcidieno@ge.imati.cnr.it}
\affil*[1]{\orgdiv{Department of Computer Science}, \orgname{University of Chile}, \orgaddress{\street{Av. Beauchef 851}, \city{Santiago}, \postcode{8370456}, \state{Región Metropolitana}, \country{Chile}}}
\affil[3]{\orgdiv{Istituto di Matematica Applicata e Tecnologie Informatiche "E. Magenes"}, \orgname{Consiglio Nazionale delle Ricerche}, \orgaddress{\street{Via De Marini 16}, \city{Genova}, \postcode{16149}, \country{Italy}}}


\abstract{Unlike image or text domains that benefit from an abundance of large-scale datasets, point cloud learning techniques frequently encounter limitations due to the scarcity of extensive datasets. To overcome this limitation, we present Symmetria, a formula-driven dataset that can be generated at any arbitrary scale. By construction, it ensures the absolute availability of precise ground truth, promotes data-efficient experimentation by requiring fewer samples, enables broad generalization across diverse geometric settings, and offers easy extensibility to new tasks and modalities. Using the concept of symmetry, we create shapes with known structure and high variability, enabling neural networks to learn point cloud features effectively. Our results demonstrate that this dataset is highly effective for point cloud self-supervised pre-training, yielding models with strong performance in downstream tasks such as classification and segmentation, which also show good few-shot learning capabilities. Additionally, our dataset can support fine-tuning models to classify real-world objects, highlighting our approach's practical utility and application. We also introduce a challenging task for symmetry detection and provide a benchmark for baseline comparisons. A significant advantage of our approach is the public availability of the dataset, the accompanying code, and the ability to generate very large collections, promoting further research and innovation in point cloud learning.}




\maketitle

\section{Introduction}
One of the main catalysts for the success of deep learning is the existence of large datasets. From the use of ImageNet~\cite{ImageNet2014} at the dawn of deep learning to LVD-142M for pre-training foundational models such as era DINOv2~\cite{Dinov2_2024}, data has been a boon for neural network models to extract patterns and learn high-level tasks. However, working with data to make it available and useful for learning has been a major challenge. In the 3D world, this situation is even more challenging: the creation and public availability of datasets has been considerably slower, limiting the study of both scalability and data-efficiency compared to 2D vision.

Pioneering datasets, such as ShapeNet~\cite{ShapeNet2015} and ModelNet~\cite{ModelNet2015}, grouped computer-designed 3D models of objects collected from the internet, categorizing and annotating them into categories of common everyday usage. These datasets are still important in the progress of 3D deep learning. 
The prevalence of these datasets is mainly due to the fact that it is complicated to obtain real 3D data; the digitization process is costly and time-consuming. Obviously, for the scale needed in machine learning, the most viable option for now is the one used to form ShapeNet and ModelNet. 

However, there are several challenges to obtaining reliable 3D data. First, data curation incurs a high cost. Privacy and copyright issues are latent concerns when downloading objects from the Internet. Second, tagging information related to third-party data is expensive. For example, although great effort has been put into preprocessing ShapeNet and aligning it to a standard coordinate system or labeling parts of an object, this procedure may not be scalable if the data set has to grow by a few orders of magnitude. More recently, Objaverse~\cite{Objaverse2023} has been presented as an interesting alternative that includes different modalities. However, the same questions remain regarding the feasibility of using data collected from the web for machine learning.

\begin{figure}[ht]
    \centering
    \includegraphics[width=\textwidth]{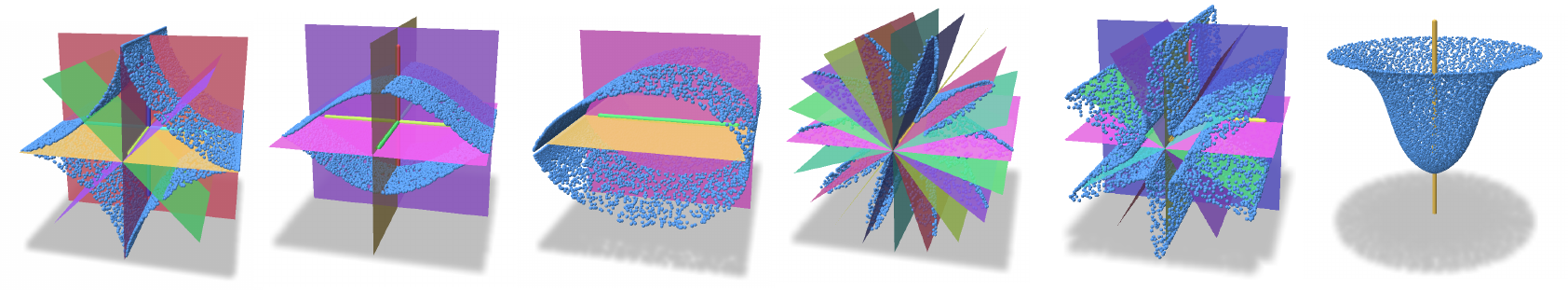}
    \caption{A glimpse into the Symmetria dataset: a formula-driven synthetic dataset composed of symmetric shapes generated from planar parametric curves.}
    \label{fig:teaser}
\end{figure}

This paper introduces Symmetria, a synthetic formula-driven dataset for learning symmetries in 3D point clouds. 
For images, formula-driven datasets have shown interesting performances, even surpassing ImageNet \cite{ImageNet2014} on some tasks \cite{Kataoka2022_Fractal}, such as understanding repeated patterns. Due to the scarcity of 3D models, the interest in synthetic datasets is even larger, and to the best of our knowledge, only the SHREC23 dataset \cite{shrec2023}, of which Symmetria is the evolution, investigated the creation of datasets using a formula-driven approach.
Our dataset originates from a set of planar curves with known geometry (see Fig.~\ref{fig:teaser}). One key observation in our design is that almost every object in the real world exhibits some symmetry; therefore, we hypothesize that a symmetry-based synthetic dataset is useful for learning 3D features, which can subsequently be useful for high-level tasks. 

Symmetry is a universal property in science, art, and nature. All scales in the physical world exhibit geometric symmetries and structural regularity, and symmetry governs many biochemical processes, leading to a richness of biological structures with strong regularity patterns. People have been inspired by the richness of symmetry in the natural world to use it to create tools, structures, artwork, and even musical compositions. In engineering and architecture, symmetry is frequently achieved through manufacturing efficiency and optimality principles, in addition to aesthetic considerations.
In the Computer Vision and Computer Graphics literature, there is a wealth of methods \cite{Mitra13} to detect global or partial symmetries, with application to object completion \cite{Sipiran_2017_ICCV},  model reduction \cite{Mitra06}, segmentation \cite{Simari06}, shape matching, and viewpoint selection \cite{Podolak06} that could benefit from this dataset. Symmetry analysis of 3D point clouds can also enable more robust object recognition and pose estimation: by understanding the symmetry properties of objects, robots can more accurately identify and localize them within the 3D environment, which is crucial for successful grasping and manipulation tasks. Symmetry is also an important indicator of the mass distribution of objects. Learning symmetries also simplifies the complexity of object manipulation by allowing the robot to infer missing or occluded parts of an object based on its visible symmetric features. When grasping an object, recognizing symmetric properties and the related mass distribution enables the robot to determine stable grasping points even if the object is partially occluded or viewed from an unfamiliar angle.

Symmetria is designed based on the following principles. 
\textbf{Large-scale exploration}: We design a procedural generation process that allows us to create any number of shapes with high variability. 
\textbf{Data-efficiency}: By providing rich and diverse structural cues through controlled procedural generation, Symmetria enables models to acquire transferable representations with substantially fewer samples than traditional curated datasets. This efficiency reduces the reliance on large-scale labor-intensive data collection while maintaining competitive downstream performance.
\textbf{Available ground truth}: Controlling the generation process, the ground truth comes for free. Information such as classification, symmetries, or parameters used to obtain a shape is easy to achieve. \textbf{Privacy}: Our formula-based dataset does not compromise copyright or sensitive information. With Symmetria, privacy is no longer a concern.
\textbf{Generality and extensibility}: The models are created by inserting arbitrary rotations. In its Erlangen Program, the prominent mathematician Klein formalised that exact symmetries in 2D and 3D can be classified and expressed in terms of symmetry groups. In our strategy, we have selected some symmetry groups, based on their high frequency in objects, to generate our dataset, but further classes of symmetries could be easily added.

We demonstrate the utility of Symmetria in two scenarios: pre-training of self-supervised point cloud models and symmetry detection.  Our first result is that pre-training with Symmetria is almost as effective as pre-training with ShapeNet. We observe effective performance in downstream tasks, such as classification, part segmentation, and few-shot learning. Our second result is that the variability of Symmetria poses a challenging task for neural symmetry detection, making it interesting to evaluate the capacity of point cloud neural networks in regression tasks.

\paragraph*{License}\label{sec:license}
The code and data generated in this paper are licensed under CC BY-NC-SA 4.0.

\section{Related works}
\label{sec:related}

\paragraph{3D datasets}
We are interested in object-centric datasets; therefore, we limit the presentation of related works to this type of data. For readers interested in scene-centric datasets and their use, we encourage the recent survey by Xiao et al.~\cite{Xiao2023}. Object-centric datasets are dominated by the pioneering ShapeNet~\cite{ShapeNet2015} and ModelNet~\cite{ModelNet2015}, which were proposed almost a decade ago and contain mainly computer-designed objects obtained from the Internet. Likewise, Thingi10K~\cite{Thingi10K2016} is a dataset of objects designed for 3D printing. The main difference from previous datasets is that 3D printed objects have particular geometric characteristics. Similarly, the ABC dataset~\cite{ABC2019} contains one million CAD-based objects with parametric representations useful for geometric learning. Recently, Objaverse~\cite{Objaverse2023} introduced a multimodal dataset containing 3D objects, images, and textual descriptions for multimodal learning. There are also real object options for object-centric data, such as ScanObjectNN~\cite{ScanObject2019}; however, their size is limited.

\paragraph{Synthetic datasets in machine learning}
The challenge of obtaining suitable and abundant data for machine learning has led to an increase in interest in synthetic data sets in various fields, such as biotechnology~\cite{BiologyMedicine2017,Helmrich2023}, robotic simulation~\cite{House3D,Orrb2019,Hypersim2021}, and natural language processing~\cite{SyntheticNLP2023}. In computer vision, synthetic data are created using computer-generated images or scenes through rendering techniques, aiding in pre-training models for images~\cite{Task2Sim2022,Yang2023}, videos~\cite{SynAPT2022,Herzig_2024_WACV,SynCDR2024}, and vision-language tasks~\cite{Cascante-Bonilla_2023_ICCV}. An alternative method involves diffusion-based generated images, which also aid in pre-training~\cite{Sariyildiz2023}. However, these methods require modeling or existing models, limiting their scalability. A more radical approach is procedural generation or formula-driven learning, exemplified by FractalDB~\cite{Kataoka2022_Fractal}, which uses fractal-generated images for pre-training convolutional networks, achieving competitive performance by leveraging geometric similarities with natural images. This approach has been extended to contour-generated images~\cite{Kataoka2022_contours} and, with the use of transformers~\cite{Nakashima2022,Takashima2023}, has shown effectiveness even without the inductive bias of translational invariance. Additionally, combining a few generated samples with data augmentation has proven effective for pre-training vision transformers~\cite{Nakamura2023}. The SHREC'2023 dataset for symmetry detection~\cite{shrec2023} is closely related to our dataset. Symmetria is the evolution of the SHREC'2023 dataset to include more variability, study the effect of scale, and evaluate the suitability of a synthetic dataset in a more general way.
Unlike SHREC'2023, Symmetria offers an expanded dictionary of parametric plane curves for surface generation, which includes ellipses, squares, and Bézier curves. Bézier curves are transformed into surfaces through a revolution process, while ellipses and squares produce simpler geometric forms, such as cylinders and cubes, through extrusion-based transformations. The dataset is hierarchically structured into four sub-datasets of 3D point clouds categorized by complexity: \textit{easy}, \textit{intermediate-1}, \textit{intermediate-2}, and \textit{hard}. Furthermore, two specialized subsets, \textit{SymSSL-10k} and \textit{SymSSL-50k}, are provided for self-supervised learning (SSL) applications. Finally, Symmetria provides more comprehensive ground truth annotations compared to SHREC'2023.

\section{The benchmark}
\label{sec:benchmark}

In this section, we describe the curves used to generate our dataset (Section \ref{sec:plane_curves}), the geometric rules to obtain 3D shapes from the plane curves (Section \ref{sec:3Dshapes}), the types of perturbations used (Section \ref{sec:perturbations}), the pipeline to generate the dataset (Section \ref{sec:dataset}) and the ground truth files associated with the point clouds of the training set (Section \ref{sec:GT}).

\subsection{Curve families}\label{sec:plane_curves}

The models of our dataset originate from a set of parametric planar curves that exhibit symmetries, which makes it easy to construct the ground truth of the symmetries of the resulting dataset. A generic parametric expression of these types of curves is

\[\mathbf{P}(t):=\begin{cases}
  x(t)=f(t,\mathbf{\Sigma}) \\ y(t)=g(t,\mathbf{\Sigma}))
\end{cases}
\label{eq:generic_curves}\]

where $f$ and $g$ are continuous functions of a common variable $t\in{}I\subseteq{}\mathbb{R}$, and $\mathbf{\Sigma}$ is a set of parameters that uniquely identify a curve in the selected family. Table \ref{tab:2Dcurves} details each family of curves, their parameters, and their symmetries. Note that our dataset also considers Bézier curves to generate shapes by revolution.

\begin{table*}[h!]
    \footnotesize
    \centering
\resizebox{.85\textwidth}{!}{
    \setlength{\leftmargini}{0.4cm}
    \begin{tabular}{|c|c|c|m{3cm}|}
    
        \hline
        \cellcolor{Aquamarine!15} \raisebox{-.5\height}{\rotatebox{90}{\makecell{Citrus \\ curve}}} & \raisebox{-.55\height}{\includegraphics[width=2.5cm]{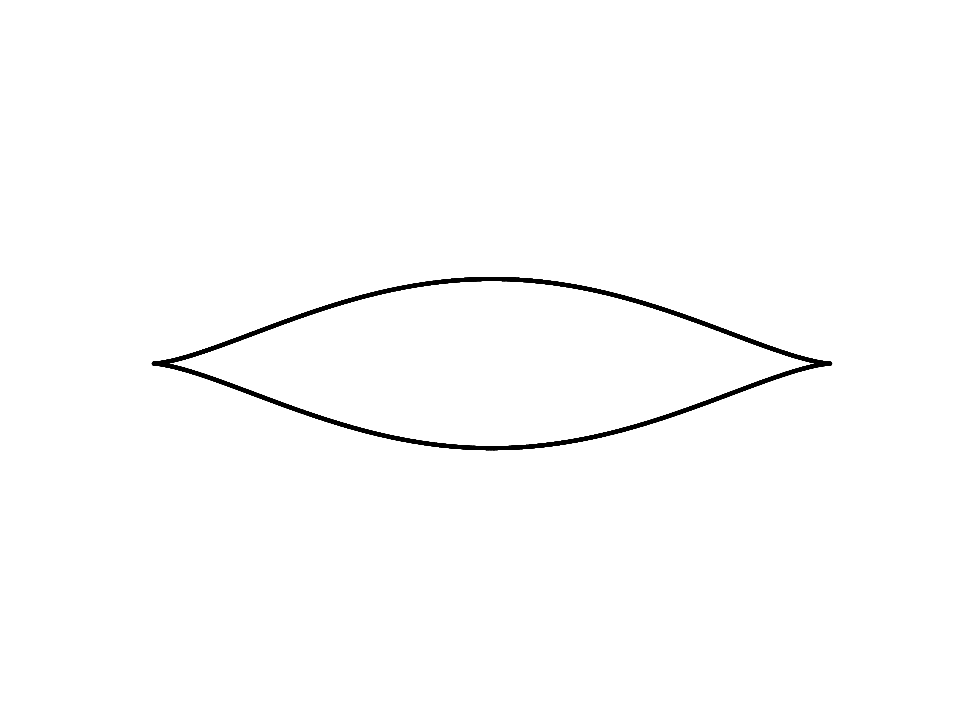}} & 
        
        $\mathbf{P}(t):=
            \begin{cases}
                x=t-\frac{a}{2} \\ y= \pm\sqrt{\frac{\left(a-t\right)^3t^3}{a^4b^2}}
            \end{cases} , \mbox{  } \mathbf{\Sigma}=\{a,b \in \mathbb{R}\}$ & 
        \begin{itemize}
            \item Ref: X axis
            \item Ref: Y axis
            \item Rot: $\pi$
        \end{itemize} \\
        \hline
        \cellcolor{Aquamarine!15} \raisebox{-.45\height}{\rotatebox{90}{$m-$Convexities}} & \raisebox{-.5\height}{\includegraphics[width=2.5cm]{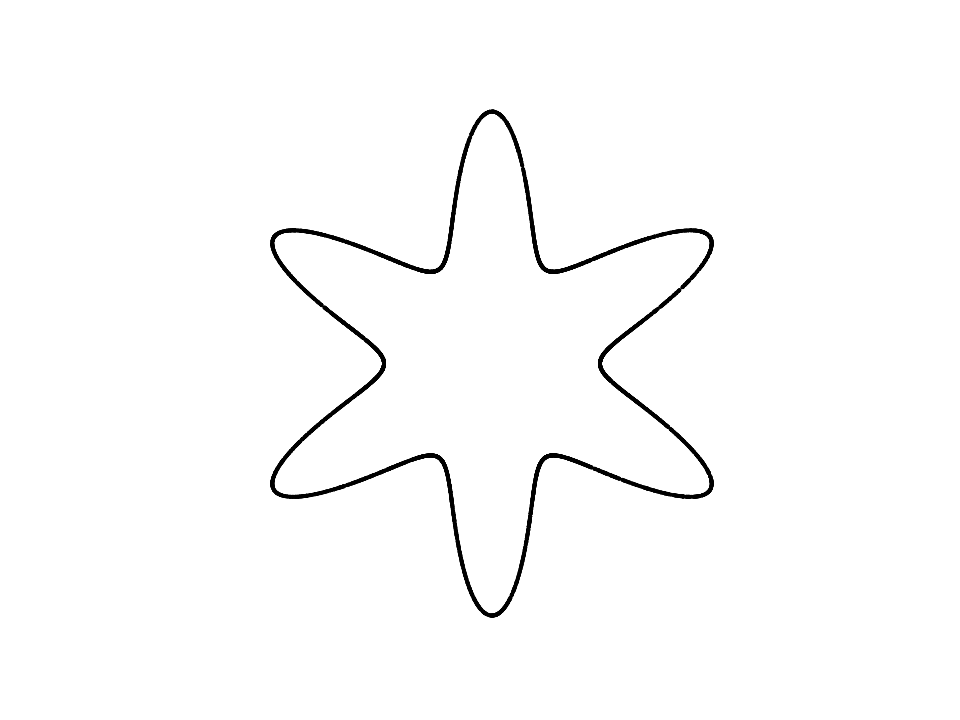}} &
         \makecell{
         $\mathbf{P}(t):=\begin{cases}
            x=\frac{a}{1+b\cos(mt)}\cos{t} \\ y=\frac{a}{1+b\cos(mt)}\sin{t}
            \end{cases},$ \\ \\
        $\mathbf{\Sigma}=\{a,b \in \mathbb{R}, m \in \mathbb{N}\}$} &
        
        \begin{itemize}
            \item Ref: lines with equations $x\sin{\frac{\pi}{m}k} - y\cos{\frac{\pi}{m}k}=0$ with $k=0,1,\ldots, m-1$
            \item Rot: $\frac{2\pi}{m}$
        \end{itemize} \\
        \hline
        \cellcolor{Aquamarine!15} \raisebox{-.5\height}{\rotatebox{90}{\makecell{Lemniscate \\ of Bernoulli}}} & 
        \raisebox{-.55\height}{\includegraphics[width=2.5cm]{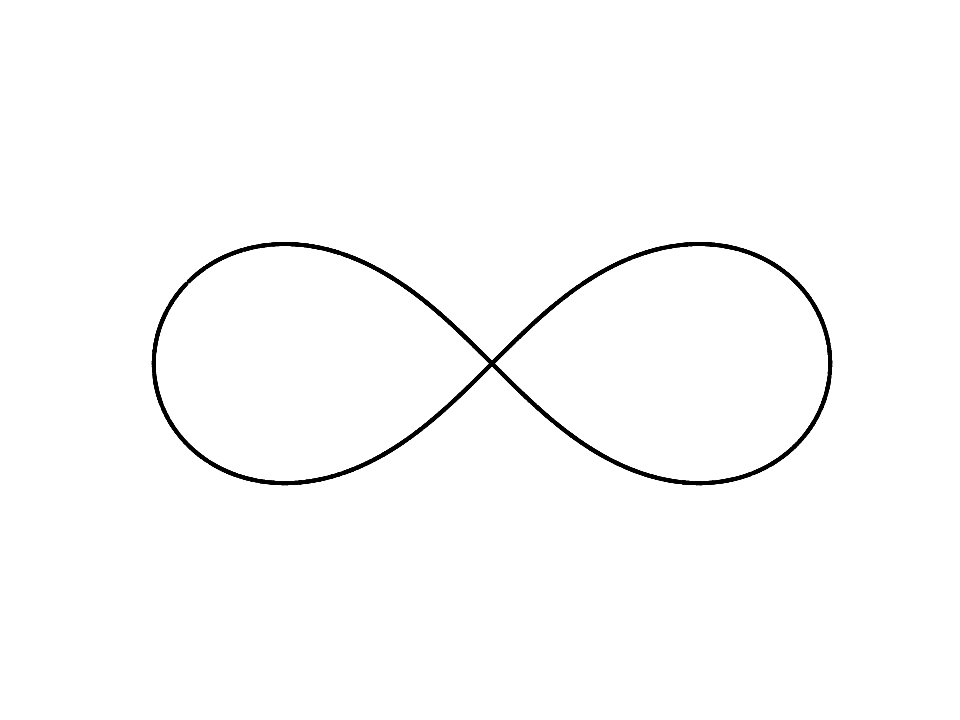}} & 
        
        $\mathbf{P}(t):=\begin{cases}
                x=a\frac{\sin{t}}{1+\cos^2t} \\ y=a\frac{\sin{t}\cos{t}}{1+\cos^2t}
        \end{cases}, \mbox{   } \mathbf{\Sigma}=\{a \in \mathbb{R}\}$ & 
        \begin{itemize}
            \item Ref: X axis
            \item Ref: Y axis
            \item Rot: $\pi$
        \end{itemize}\\
        \hline
        \cellcolor{Aquamarine!15} \raisebox{-.4\height}{\rotatebox{90}{\makecell{Egg of\\ Keplero}}} & 
        \raisebox{-.4\height}{\includegraphics[width=2.5cm]{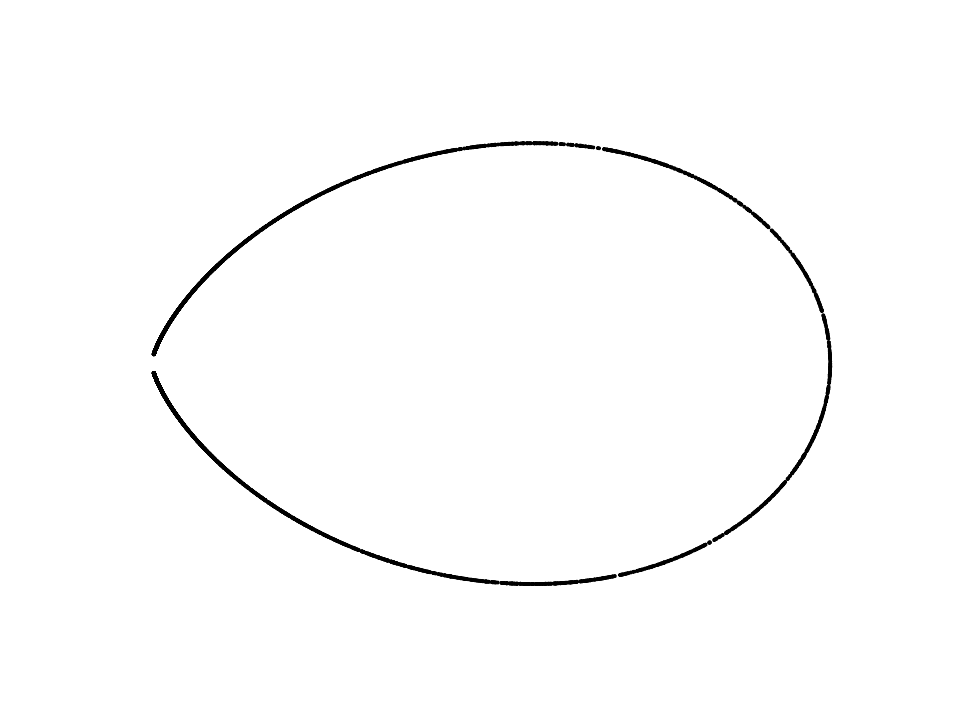}} & 
        
            $\mathbf{P}(t):=
            \begin{cases}
                x=\frac{a}{\left(1+t^2\right)^2}\\ y=\frac{at}{\left(1+t^2\right)^2}
            \end{cases}, \mbox{   } \mathbf{\Sigma}=\{a \in \mathbb{R}\}$ & 
        \begin{itemize}
            \item Ref: X axis
        \end{itemize}\\
        \hline
        \cellcolor{Aquamarine!15} \raisebox{-.3\height}{\rotatebox{90}{\makecell{Mouth\\ curve}}} & 
        \raisebox{-.4\height}{\includegraphics[width=2.5cm]{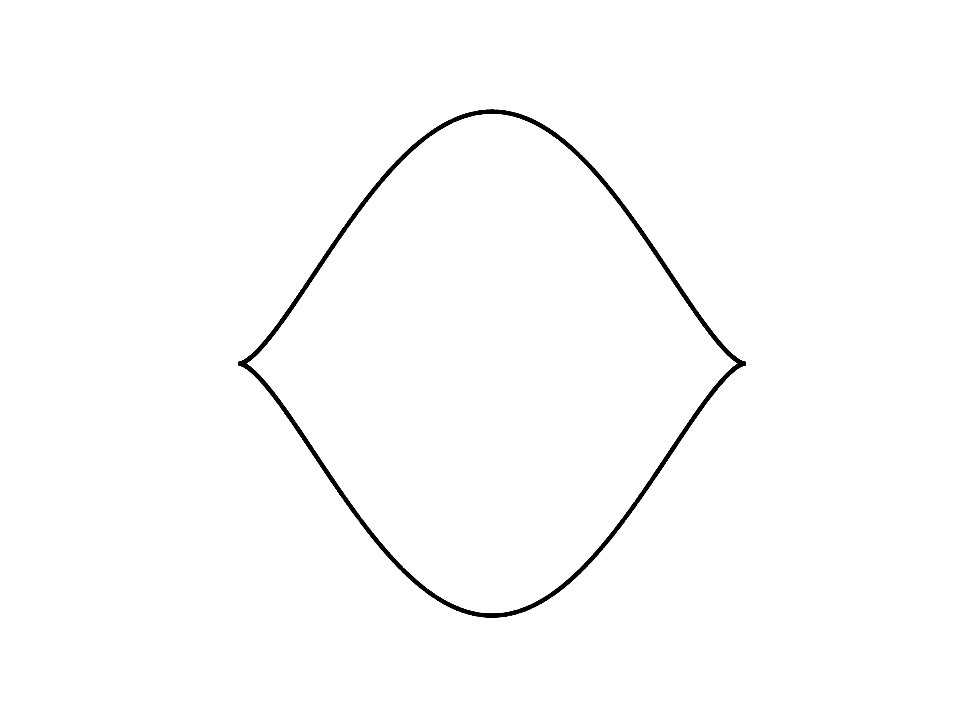}} & 
       
            $\mathbf{P}(t):\begin{cases}
                x=a\cos{t} \\ y=a\sin^3t
            \end{cases}, \mbox{   } \mathbf{\Sigma}=\{a \in \mathbb{R}\}$ & 
        \begin{itemize}
            \item Ref: X axis
            \item Ref: Y axis
            \item Rot: $\pi$
        \end{itemize}\\
        \hline
        \cellcolor{Aquamarine!15} \raisebox{-.3\height}{\rotatebox{90}{\makecell{Geometric\\ petal}}} & 
        \raisebox{-.5\height}{\includegraphics[width=2.5cm]{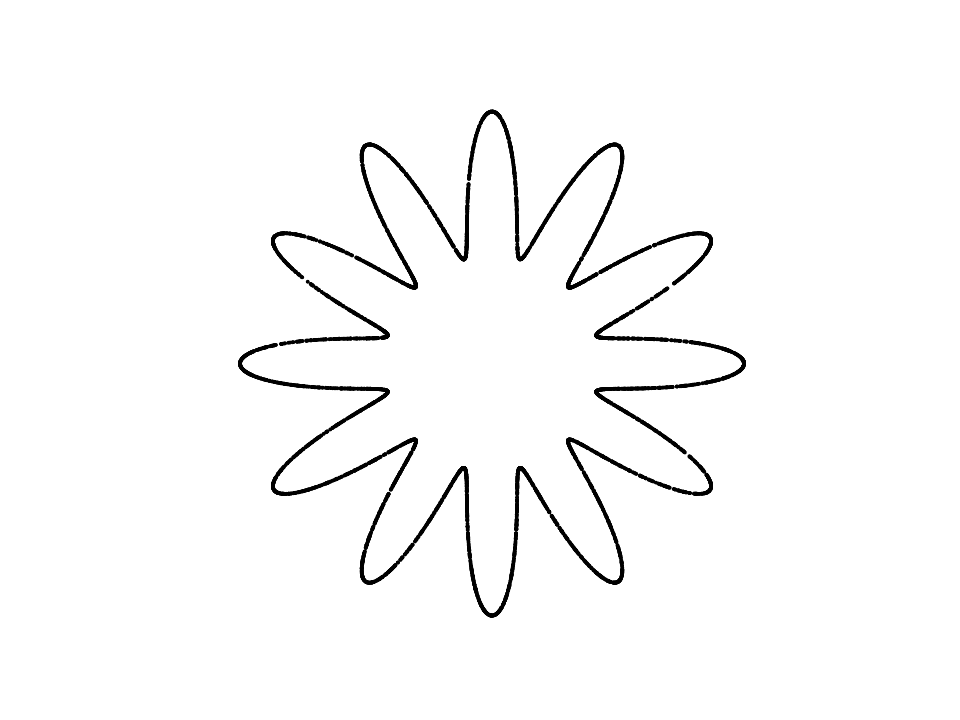}} & 
        \makecell{
            $\mathbf{P}(t):=
                \begin{cases}
                x=\left(a+b\cos mt\right)\cos{t} \\ y=\left(a+b\cos mt\right)\sin{t}
                \end{cases}$ \\ \\
        $\mathbf{\Sigma}=\{a,b \in \mathbb{R}, m \in \mathbb{N}\}$} & 
        \begin{itemize}
            \item Ref: lines with equations $x\sin{\frac{\pi}{m}k} - y\cos{\frac{\pi}{m}k}=0$ with $k=0,1,\ldots, m-1$
            \item Rot: $\frac{2\pi}{m}$
        \end{itemize}\\
        \hline
        \cellcolor{Aquamarine!15} \raisebox{-.7\height}{\rotatebox{90}{Astroid}} & 
        \raisebox{-.7\height}{\includegraphics[width=2.5cm]{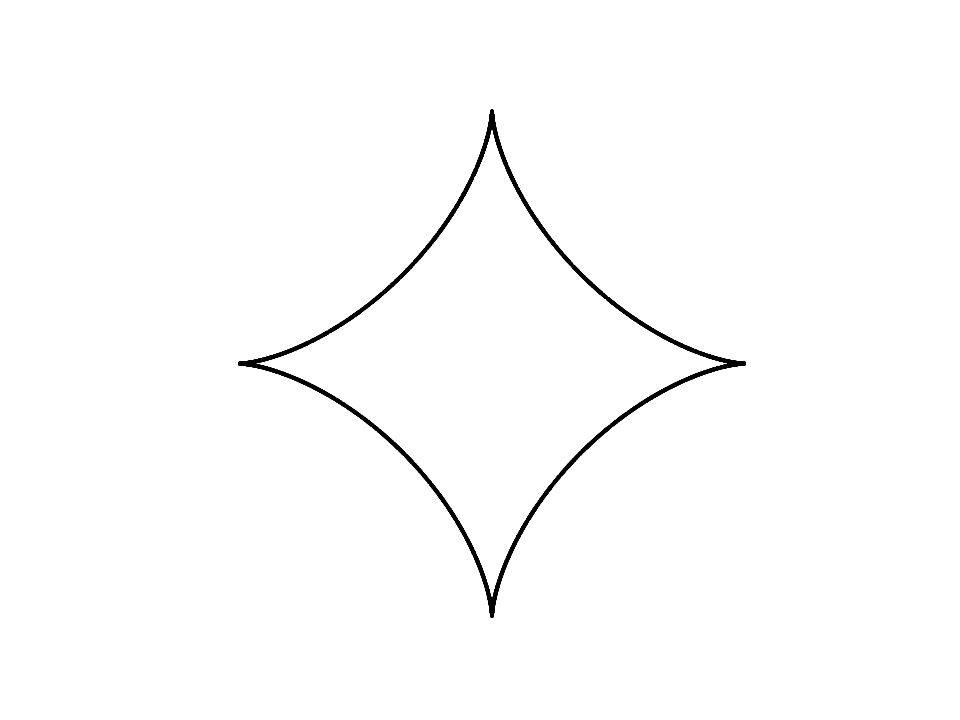}} & 
        
            $\mathbf{P}(t):=\begin{cases}
                x=a\cos^3t \\ y=a\sin^3t
            \end{cases}, \mbox{   } \mathbf{\Sigma}=\{a \in \mathbb{R}\} $ & 
        \begin{itemize}
            \item Ref: 
            \item Rot: $\pi/2$
        \end{itemize}\\
        \hline
        \cellcolor{Aquamarine!15} \raisebox{-.3\height}{\rotatebox{90}{Ellipse}} & 
        \raisebox{-.5\height}{\includegraphics[width=2.5cm]{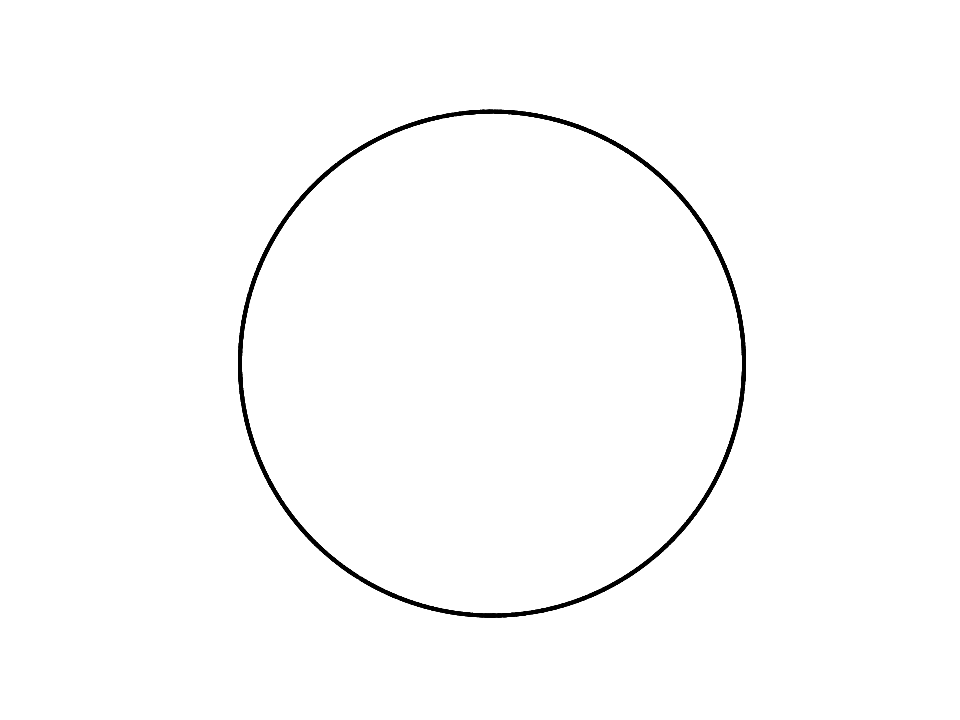}} & 
        
            $\mathbf{P}(t):=\begin{cases}
                x=a\cos{t} \\ y=b\sin{t}
            \end{cases}, \mbox{  } \mathbf{\Sigma}=\{a,b \in \mathbb{R}\}$ & 
        \begin{itemize}
            \item Rot: continuous when $a=b$
            \item Rot: $\pi/2$, otherwise
        \end{itemize}\\
        \hline
        \cellcolor{Aquamarine!15} \raisebox{-.3\height}{\rotatebox{90}{Square}} & 
        \raisebox{-.5\height}{\includegraphics[width=2.5cm]{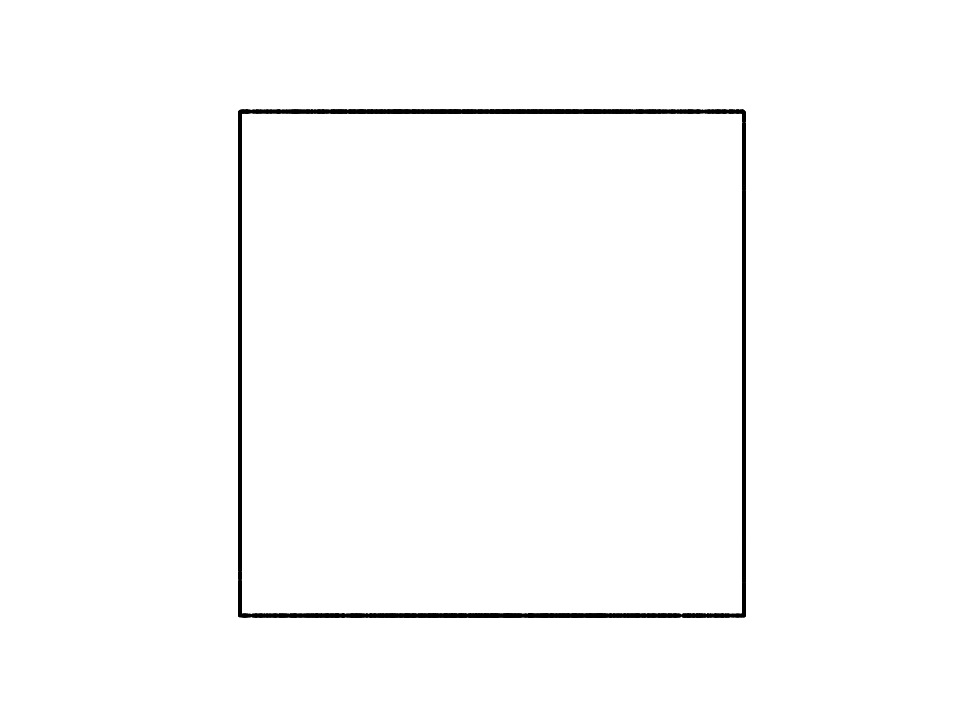}} & 
        
            $\mathbf{P}(t):=\begin{cases}
                x=\frac{a\cos{t}}{\max(|\cos{t}|,|\sin{t}|)} \\ y=\frac{b\sin{t}}{\max(|\cos{t}|,|\sin{t}|)}
            \end{cases}, \mbox{  } \mathbf{\Sigma}=\{a,b \in \mathbb{R}\}$ & 
        \begin{itemize}
            \item Ref: X axis
            \item Ref: Y axis
            \item Rot: $\pi/2$
        \end{itemize}\\
        \hline
        \cellcolor{Aquamarine!15} \raisebox{-.8\height}{\rotatebox{90}{Bézier}} & 
        \raisebox{-.8\height}{\includegraphics[width=2.5cm]{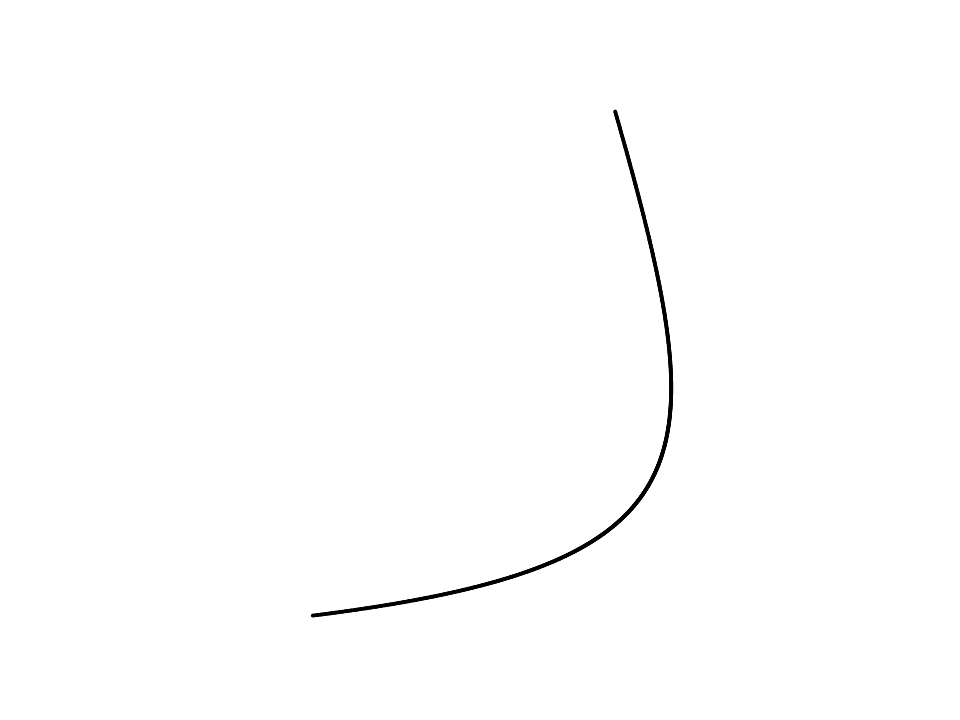}} & 

        \begin{tabular}{c}
        \\
            $\mathbf{P}(t):=\sum_{i=0}^3 \binom{3}{i}(1-t)^{3-i}t^i B_i,$   \\
             $\mathbf{\Sigma}=\{B_i \in \mathbb{R}^3, i=0,1,2,3\}$ \\
        \end{tabular}
            & 
        \begin{itemize}
            \item N.A.
        \end{itemize}\\
        \hline
    \end{tabular}
    }
    \caption{Families of plane curves used to generate the Symmetria dataset.}
    \label{tab:2Dcurves}
\end{table*}

\subsection{From curves to surfaces}\label{sec:3Dshapes}
We convert the curves into surfaces embedded in the Euclidean space $\mathbb{R}^3$ by applying an extrusion operation along the $Z$ axis, except for Bézier curves converted into surfaces by revolution. Next, we describe the conversion procedures.

\subsubsection{Extrusion}
We assume that the curves lie on the $XY$ plane. A straightforward extrusion would generate a cylinder along the $Z$ axis with the curve as its directrix. However, this procedure would leave lines along the $Z$ axis, which do not necessarily represent the 3D shape well. We opt for a more general approach to generating a well-distributed point cloud.

First, we generate $N^2$ random values for the curve parameter $t$. The number $N$ is a parameter of our generation method, with $N=80$ for all the experiments in this paper. Second, we randomly select the size $l_z$ of the 3D shape in the $Z$ direction. Third, each point in the original curve is randomly shifted along the axis $Z$. That is, given a point $p=(x,y)$ in $\mathbb{R}^2$, the corresponding point in $\mathbb{R}^3$ is $\hat{p} = (x,y,l_z \times z - l_z/2)$ for $z \sim \mathcal{U}(0,1)$. Therefore, the 3D shape is defined in the $[-l_z/2, l_z/2]$ interval in the $Z$ axis. Also, our procedure shifts the original points in the curve in the third dimension, so the final point cloud is well distributed on the surface of the extruded shape.

\begin{figure}[h!]
    \centering
    \includegraphics[width=8cm]{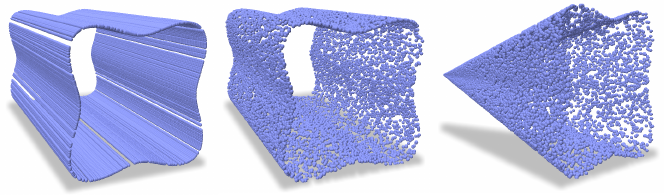}
  \caption{Extrusion examples}
\end{figure}

The previous procedure can be seen as a cylindrical extrusion of the curve. We also propose a conical extrusion to add more variability to the dataset. Given a point $p=(x,y)$, the corresponding conical extruded point is $\hat{p} = (z\times x,z\times y,l_z \times z - l_z/2)$ for $z \sim \mathcal{U}(0,1)$.

\subsubsection{Revolution}

Bézier curves are converted to surfaces through a revolution operation. To avoid poorly distributed points on the generated solid, we performed a randomized revolution, similar to the extrusion process described before. Given a point $p=(x,y)$ in $\mathbb{R}^2$, our method generates a set of points  in $\mathbb{R}^3$ obtained by rotating $p$ in the $Y$ axis with several angles, that is, $\hat{p}_j = R_Y(p, \sphericalangle(\theta_j + \sigma))$, where $\theta_j = 360\degree\times j / T$ and $\sigma \sim \mathcal{U}(-3.4\degree, 3.4\degree)$. In all of our experiments, we set $T = 100$ for the number of rotations applied to each point. 

\begin{figure}[h!]
    \centering
    \includegraphics[width=4cm]{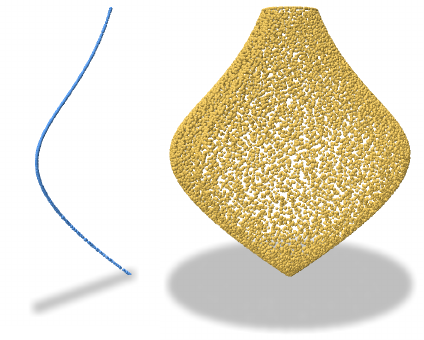}
  \caption{Revolution example.}
\end{figure}

\subsubsection{Equivalences between planar and spatial symmetries}
Each planar curve has symmetries as described in Table~\ref{tab:2Dcurves}. Therefore, the resulting 3D shapes have equivalent symmetries in space, as described next.

\begin{itemize}
    \item 2D reflection on $X$ axis $\longrightarrow$ 3D reflection on XZ plane
    \item 2D reflection on $Y$ axis $\longrightarrow$ 3D reflection on YZ plane
    \item 2D reflection on lines with equations $x\sin{\frac{\pi}{m}k} - y\cos{\frac{\pi}{m}k}=0$ with $k=0,1,\ldots, m-1$ $\longrightarrow$ 3D reflection on planes $x\sin{\frac{\pi}{m}k} - y\cos{\frac{\pi}{m}k}=0$ with $k=0,1,\ldots, m-1$.
    \item 2D rotation with angle $\alpha$ 
    $\longrightarrow$ 3D rotation on Z axis with angle $\alpha$. 
    \item Exceptionally, the Bézier curves get a continuous rotational symmetry on the Y-axis after the revolution.
\end{itemize}

\subsection{Perturbations}\label{sec:perturbations}
To add variability, we performed some perturbations in the 3D shapes.
More specifically, we consider one of the following variations. 1) \textbf{Clean}. The point cloud is not perturbed. 2) \textbf{Uniform noise}. A random percentage, between $30\%$ and $80\%$, of the point cloud is perturbed by applying a uniform noise, obtained by sampling uniform distributions of the form $\mathcal{U}(-\frac{1}{n},\frac{1}{n})$, being $n$ a random value among $15,\,17,\,19,\,20$. 3)  \textbf{Gaussian noise}. A random percentage, between $30\%$ and $80\%$ of the point cloud, is perturbed by applying a Gaussian noise, obtained by sampling Gaussian distributions of the form $\mathcal{N}(0,\frac{1}{n^2})$, being $n$ a random value among $20,\,23,\,27,\,30$. 4) \textbf{Undersampling}. A random percentage, between $30\%$ and $80\%$, of the point cloud is removed. 5) \textbf{Uniform noise $+$ undersampling}. 6) \textbf{Gaussian noise $+$ undersampling.}($P_5$). Regarding noise transformations, we first decide how many points will be perturbed with a probability between 30\% and 80\%. For each point to perturb, we generate a random shift in the distribution (uniform or Gaussian) with the value of $n$ also chosen randomly. Table \ref{tab:pertubations} summarizes the type and the entity of the perturbations as distributed in the dataset.
\begin{table}[h!]
    \centering
    \begin{tabular}{|c|c|c|c|c|}
    \hline
    \textbf{Perturbation type} &  \textbf{Clean}   &  \textbf{Uniform noise} & \textbf{Gaussian noise}  & \textbf{Undersampling} \\
        & & $\mathcal{U}(-\frac{1}{n},\frac{1}{n})$ &  $\mathcal{N}(0,\frac{1}{n^2})$  & \\
    \hline
    $n$ & $-$ & 15, 17, 19, 20 & 20, 23, 27, 30 & $-$\\
    \hline
    percentage & $-$ & from 30\% to 80\%  & from 30\% to 80\% & from 30\% to 80\%\\
    of perturbed points & & &  &\\
    \hline          
    \end{tabular}
    \caption{Summary of the type and size of perturbations applied to the mathematical surfaces.}
    \label{tab:pertubations}
\end{table}

\subsection{Surface generation}\label{sec:shape_generation}
The point clouds provided in the Symmetria dataset were generated using the following procedure (see Algorithm \ref{alg:pipeline_dataset}): for each perturbation presented in Section \ref{sec:perturbations}, we first randomly choose one of the closed plane curves described in Section \ref{sec:plane_curves} and randomly select the parameters' values in $\mathbf{\Sigma}$. Specifically, we set the parameters of each curve as follows:
 \begin{itemize}
    \item Citrus: $a = 1$ and $b$ randomly chosen in the interval $[1,13], \,b\in \mathbb{N}$.
    \item m-Convexities: $a$ randomly chosen in the interval $[0.5, 1.1],$ $a \in \mathbb{R}$, $b$ randomly chosen in the interval $[0.2, 0.9],$ $ b \in \mathbb{R}$, and $m$ randomly chosen in the interval $[3,9],$ $ m \in \mathbb{N}$.
    \item Geometric petal: $a$ randomly chosen in the interval $[1.0, 2.0], \, a \in \mathbb{R}$, $b$ randomly chosen in the interval $[1, 6], \, b \in \mathbb{N}$, and $m$ randomly chosen in the interval $[1,6], \, m \in \mathbb{N}$.
    \item Lemniscate of Bernoulli: $a = 1$.
    \item Egg of Keplero: $a = 1$.
    \item Mouth curve: $a = 1$.
    \item Astroid: $a = 1$.
    \item Square: $a = 1$ and $b$ randomly chosen in the interval $[0.5, 1.5]$.
    \item Cylinder: $a = 1$ and $b$ randomly chosen in the interval $[0.5, 1.5]$.
    \item Revolution: Four control points are randomly chosen the following way: $C_0=(x_0, 1, 0), C_1=(x_1,y_1,0), C_2=(x_2,y_2,0), C_3=(0, 0, 0)$, where $x_0,x_1,y_1,x_2$ are randomly chosen in the interval $[0.1, 1.0]$ and $y_2$ is randomly chosen in the interval $[0, 1]$. These control points ensure the generation of curves with no self-intersections and good variability.
 \end{itemize}

 Then, we generate the 3D point cloud by randomly selecting one of the shapes at our disposal, and we apply the selected type of perturbation. For conciseness, Table \ref{tab:parameters_curves} summarizes the curve parameters chosen for the creation of our dataset.

  \begin{table}[h!]
    \centering
  \resizebox{.99\textwidth}{!}{  \begin{tabular}{|c|c|c|c|c|c|c|c|c|}
    \hline
    \textbf{Citrus} &  \textbf{m-Conv.}   &  \textbf{Geom. petal} & \textbf{Lemnisc.}  & \textbf{Egg} & \textbf{Mouth} & \textbf{Astroid} & \textbf{Square} & \textbf{Cylinder}\\
    \hline
 $a=1,\,$ & $a\in[0.5,\,1.1],\, a\in\mathbb{R},$ & $a\in[1.0,\,2.0],\, a\in\mathbb{R},$ & $a=1$ & $a=1$ & $a=1$ & $a=1$ & $a=1$ & $a=1$ \\
  $b\in[1,13],\,b\in\mathbb{N}$ & $b\in[0.2,\,0.9],\, b\in\mathbb{R},$ & $b\in[1,\,6],\, b\in\mathbb{N},$ & & & & & $b\in[0.5,\,1.5]$ & $b\in[0.5,\,1.5]$\\
& $m\in[3,\,9],\, m\in\mathbb{N},$ & $m\in[1,\,6],\, m\in\mathbb{N},$  & & & & & &  \\
    \hline
    \end{tabular}}
    \caption{Summary of the parameters chosen to generate the plane curves used to create the dataset.}
    \label{tab:parameters_curves}
\end{table}
 
 Since these shapes have the rotational axis coincident with the $z$-axis, we randomly apply translations and/or rotations to the point cloud to make its position/orientation more generic. We first apply a translation with a 0.8 probability and subsequently a rotation with a variable probability from 0.5 on the $x$-axis to 1.0 on all three axes, according to the difficulty of the sub-dataset. Note that we do not apply a scale normalization of the objects and let the methods decide on this transformation. The size of objects varies depending on the parameters used during the generation.

\subsection{Generation of the dataset}\label{sec:dataset}
The dataset, \textit{Symmetria}, is created using a publicly available Python script\footnote{\url{https://github.com/ivansipiran/symmetria}}. It consists of three-dimensional simple shapes represented as XYZ point clouds, provided in xz-compressed text files, and ground truth (GT) data in uncompressed text files. The dataset is organized into four sub-datasets of increasing complexity: \textit{easy}, \textit{intermediate-1}, \textit{intermediate-2}, and \textit{hard}, plus two datasets, \textit{SymSSL-10k} and \textit{SymSSL-50k}, for self-supervised learning (SSL) experiments. Each sub-dataset is split into training (70\%), validation (20\%), and test (10\%) sets, with files grouped by class for convenience. Shapes in the sub-datasets undergo random transformations as presented in Section~\ref{sec:perturbations}, with a probability of 0.8, and are shifted uniformly in the interval [0, 1) with the same probability. The probability of generating a conic extrusion is 0.5. Further details on SSL and supervised learning datasets are provided in Section~\ref{sec:ssl} and Section~\ref{sec:symmetrydetection}, respectively.

The generation process was designed to have a balanced dataset in terms of the type of perturbations described in Section~\ref{sec:perturbations}. Additionally, the parameters $\Sigma$ for the curves were randomly selected to ensure high variability. Details regarding the parameters used and the overall generation procedure are provided in the Appendix.

The pseudocode of Algorithm \ref{alg:pipeline_dataset} shows the steps necessary to generate the 3D shapes and to choose the perturbations to be applied to create a single sample that belongs to the Symmetria dataset.

\begin{algorithm}[hb!]
	\caption{Dataset generation pseudocode}
	\begin{algorithmic}[1]
            \State Let $P=\{P_0,P_1,P_2,P_3,P_4,P_5\}$ be the set of perturbations
            \State Let $S$ be the set of curve families (citrus, $m-$convexities, etc.)
            \State Let $Pr_{conic}$ be the probability of generating a conic extrusion
            \State Let $Pr_{R}$ be the probability of a random rotation
            \State Let $Pr_{T}$ be the probability of a random translation
            \State $D=\emptyset$
            
		\For {every perturbation $P_i \in P$}
		      \State type\_shape = random choice of $S$
                \State Random selection of parameters $\Sigma$
                \State shape = generate\_shape(type\_shape, $\Sigma$, $Pr_{conic}$)
                \State Apply perturbation $P_i$ to shape
                \State Apply random rotation to shape with probability $Pr_{R}$
                \State Apply random translation to shape with probability $Pr_{T}$
                \State $D = D \cup \{$shape$\}$

            return $D$
            \EndFor
	\end{algorithmic} 
    \label{alg:pipeline_dataset}
\end{algorithm}

\subsubsection{Ground truth}\label{sec:GT}

Each point cloud in training, validation, and test sets is provided with its ground truth text file containing the planar and axial symmetries of the 3D shape. The unique ID of the shape, the class, and the transformations applied to the point cloud can be obtained from the file name (e.g., \textit{008706-revolution-undersampling+uniform.xz} and the corresponding ground truth file "\textit{-sym.txt}"). The detailed format of the data within the ground truth files is available in the Appendix. The \textbf{Symmetria} dataset can be downloaded at the following website: \url{http://deeplearning.ge.imati.cnr.it/symmetria}.

\paragraph{Groundtruth data format}\label{sec:dataset_detailed_gt}
For each \textit{-sym.txt} file:
\begin{itemize}
    \item the first line of the text file contains a single integer in the range $[1-14]$ representing the overall number of symmetries, both planar and rotational, of the shape;
    \item for each of these symmetries, the ground truth file contains an additional line in the following format.
    \begin{itemize}
        \item \textit{Planar symmetry}: the line begins with the word \textit{plane} and is followed by 6 floats. The first group of 3 represents the normal vector of the symmetry plane, while the second group of 3 represents the coordinates of the point on the plane from which the normal vector originates.
        \item \textit{Axial symmetry}: the line begins with the word \textit{axis} and is followed by 7 floats. The first 6 represent the normal and the point of origin as for planar symmetries, and the 7th represents the periodicity expressed in radians along the rotational axis. So, for instance, \textit{Astroids} will only have axes with periodicity $\pi/2$ or $\pi$ (repetitions every quarter of a rotation along the "longitudinal" symmetry axis or every half rotation along the "transversal" axes of symmetry) while the \textit{Revolutions} will have a single axis of symmetry with infinite periodicity as they are "true" solids of rotation.
    \end{itemize}
\end{itemize}

\subsubsection{Dataset variability}

All shapes are generated starting from parametric equations and subsequently subjected to extrusion or rotation operations. Table \ref{tab:shapes} provides a view of the intraclass variability in the data set when the perturbations described in Section \ref{sec:perturbations} are applied. For each row, some examples of the surfaces within a class are shown. For instance, intra-class variation in the citrus class exhibits an important variation in the surface eccentricity. In the m-Convexities class, also the number of convexities varies, thus presenting a different number of symmetry axes within the same class. Similarly, from the geometric petal curve we have generated shapes with different symmetries and even different sizes of the petals. In addition, surfaces were perturbed with noise and arbitrarily rotated.

\begin{table*}[h!]
    \centering
    \resizebox{.99\textwidth}{!}{
    \begin{tabular}{|c|c|}
    \hline
    & \\
    \multirow{2}{*}{\begin{tabular}{c} \textbf{Citrus} \\ $a = 1$ \\ $b \in [1, 13], b \in \mathbb{N}$ \end{tabular}} & \includegraphics[width=10cm,height=1.25cm]{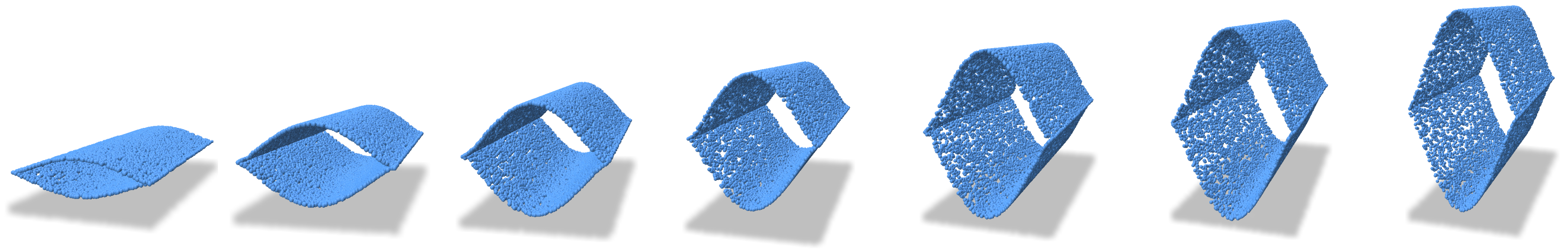} \\
     & \includegraphics[width=10cm,height=1.25cm]{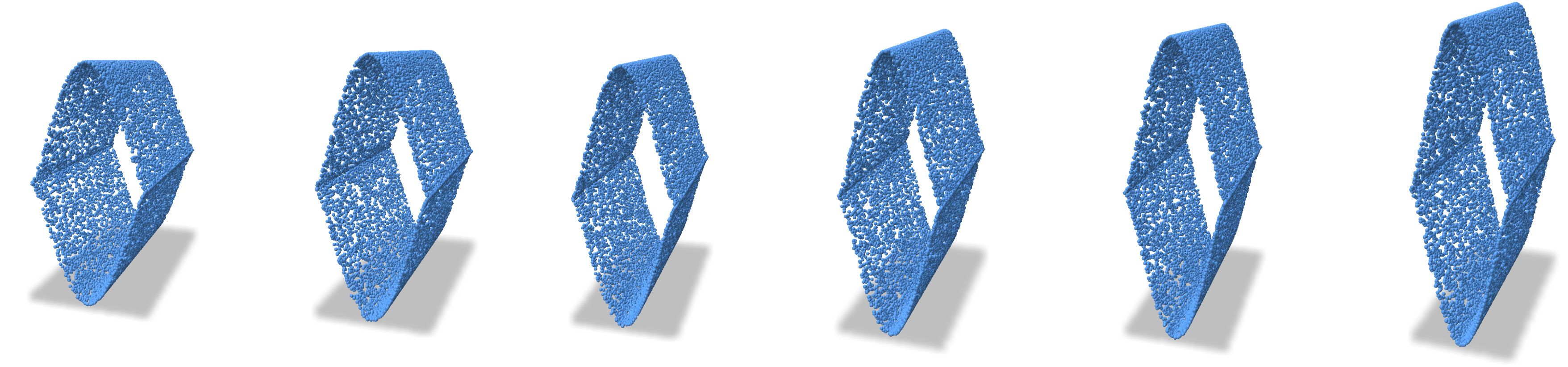} \\
    \hline
      & \\
    \multirow{2}{*}{\begin{tabular}{c} \textbf{m-Convexities} \\ Top: $a = 0.5$, $b = 0.2$ \\ Bottom: $a = 1.1$, $b = 0.9$ \\ $m \in [3,9], m \in \mathbb{N}$ \end{tabular}} & \includegraphics[width=10cm]{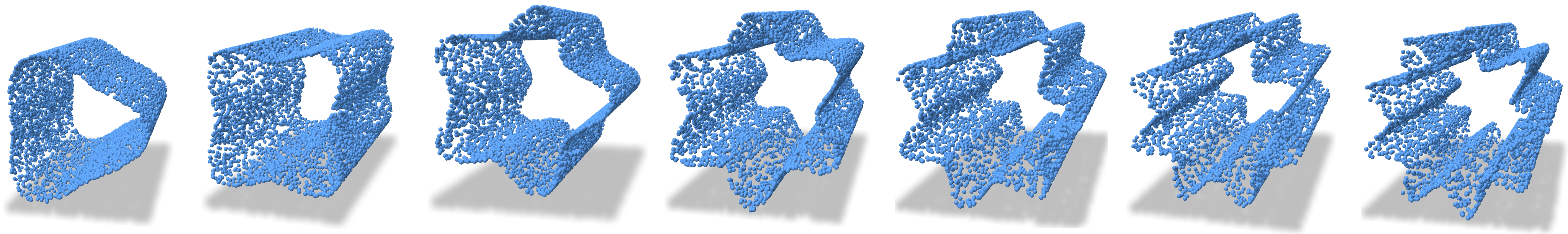} \\
     & \includegraphics[width=10cm]{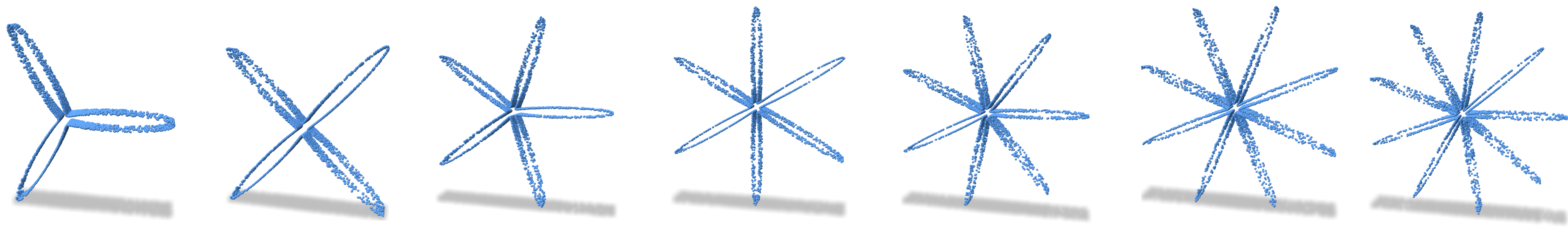} \\
    \hline
      & \\
    \multirow{2}{*}{\begin{tabular}{c} \makecell{\textbf{Geometric petal} \\ Top: $a = 1.0$, $b = 1$ \\ Bottom: $a = 2.0$, $b = 6$ \\ $m \in [1,6], m \in \mathbb{N}$} \end{tabular}} & \includegraphics[width=10cm]{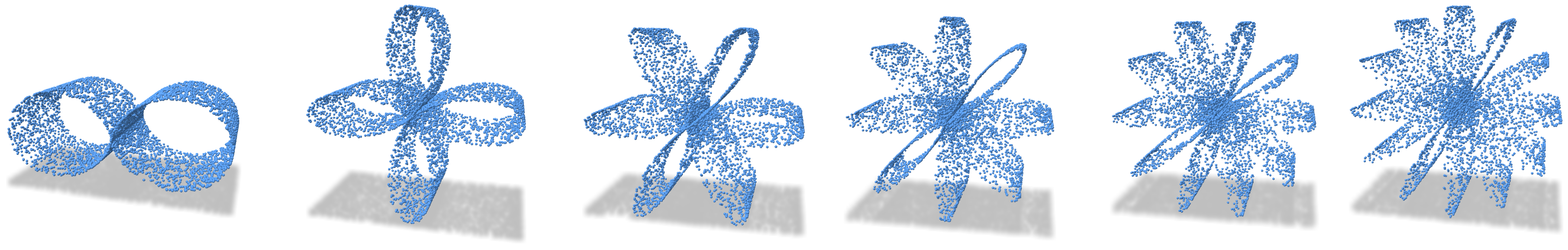} \\
    & \includegraphics[width=10cm]{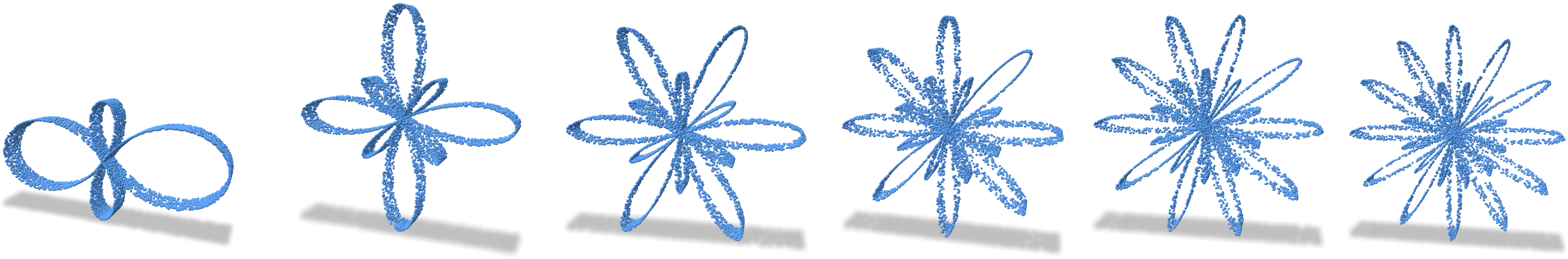} \\
    \hline
      & \\
    \begin{tabular}{c} \textbf{Lemniscate of  Bernoulli} \\ \textbf{Egg of Keplero} \\ \textbf{Mouth curve} \\ \textbf{Astroid} \end{tabular} & \raisebox{-0.9cm}{\includegraphics[width=10cm]{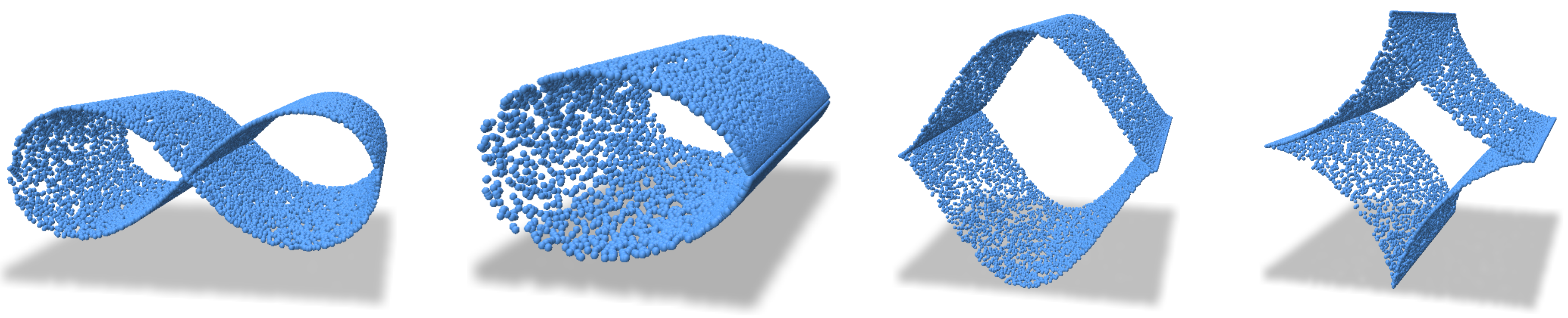}} \\
    \hline
    \end{tabular}
    }
    \caption{Samples of surfaces available in the Symmetria dataset.}
    \label{tab:shapes}
 \end{table*}

\section{Experiments}
\subsection{Self-supervised learning on point clouds}\label{sec:ssl}
The first objective of our research is to investigate whether the proposed dataset is useful for learning embeddings in a self-supervised manner. The typical approach in the literature is to use ShapeNet as a dataset for pre-training and to evaluate the effectiveness of pre-trained models with more specialized tasks, such as classification and part segmentation, over other datasets, such as ModelNet and ShapeNet-part. We drastically changed this approach and used our dataset for the pre-training of the self-supervised models. We hypothesized that the proposed dataset contains unique features that make it useful for learning self-supervised embeddings. For comparison, in all of our results, we show the variation in performance with respect to the results of the models pre-trained with ShapeNet. The reported experiments were run on three NVIDIA RTX A5000 GPUs with 24GB of memory each. All experiments (pre-training, fine-tuning, and testing) took approximately 120 hours.

\paragraph{Pre-training}
For experimenting with SSL methods, we employed SymSSL-10K and SymSSL-50K, containing 10,000 and 50,000 point clouds, respectively. We selected Point-MAE~\cite{PointMAE2022} and PointGPT~\cite{PointGPT2023} as self-supervised architectures for this task. Point-MAE uses a masked autoencoder strategy: groups of points are hidden, and the network is trained to reconstruct the input. PointGPT is an autoregressive model trained to predict a shape sequentially by generating sequences of points that are similar to the training data. Both models are state-of-the-art in pre-training for 3D point clouds, thus allowing us to study the effect of using different datasets for pre-training. Note that we used the default settings reported in the literature for each method. For PointGPT, we used the small version PointGPT-S. 

\paragraph{Classification on scanned objects}
In this experiment, the pre-trained networks are fine-tuned to classify the ScanObjectNN dataset~\cite{ScanObject2019}. The dataset comes in three different versions: OBJ-BG (objects with background), OBJ-ONLY (isolated objects), and PB-T50-RS (objects perturbed with transformations). Table~\ref{Tab:ScanObject} shows our results. Despite a general performance gap compared to ShapeNet, the results demonstrate that SymSSL can serve as a viable alternative for pre-training, especially in scenarios where access to large-scale curated datasets is limited. For Point-MAE, pre-training on SymSSL results in only modest accuracy drops (e.g., -0.5\% on OBJ-ONLY with SymSSL-10K), showing that it can still yield competitive performance. Increasing the size of the SymSSL dataset to 50K further narrows the performance gap, suggesting that scaling SymSSL improves generalization. Notably, for more challenging variants like PB-T50-RS, both Point-MAE and PointGPT-S retain strong performance even when pre-trained on SymSSL, indicating that SymSSL is particularly effective for learning robust representations under geometric perturbations. These findings highlight the potential of SymSSL as a cost-effective and scalable alternative for pre-training 3D models.

\begin{table}[h]
    \centering\footnotesize
    \begin{tabular}{cclll}
    \hline
    \textbf{Model} & \textbf{\makecell{Pre-training\\dataset}} & \textbf{OBJ-BG} & \textbf{OBJ-ONLY} & \textbf{PB-T50-RS} \\ \hline
    Point-MAE & ShapeNet & 90.0 & 88.3 & 85.2 \\
    Point-MAE & SymSSL-10K & 86.1 \textcolor{blue}{(-3.9)} & 87.8 \textcolor{blue}{(-0.5)}& 82.4 \textcolor{blue}{(-2.8)} \\
    Point-MAE & SymSSL-50K & 87.8 \textcolor{blue}{(-2.2)}& 86.6 \textcolor{blue}{(-1.7)} & 83.0 \textcolor{blue}{(-2.2)} \\ \hline
    PointGPT-S & ShapeNet & 91.6 & 90.0 & 86.9 \\
    PointGPT-S & SymSSL-10K & 73.8 \textcolor{blue}{(-17.8)}  & 79.0 \textcolor{blue}{(-11.00)} & 84.3 \textcolor{blue}{(-2.6)} \\
    PointGPT-S & SymSSL-50K & 74.4 \textcolor{blue}{(-17.2)} & 79.4 \textcolor{blue}{(-10.6)} & 84.9 \textcolor{blue}{(-2.0)}\\ \hline
    \end{tabular}
   \caption{Classification accuracy on ScanObjectNN across three variants. Point-MAE shows minor drops with SymSSL pre-training compared to ShapeNet, while PointGPT-S exhibits larger declines (percentages in \textcolor{blue}{blue}), particularly on OBJ-BG and OBJ-ONLY. Results highlight that SymSSL pre-training transfers reasonably well to real-world data, though effectiveness varies by model and scenario.}
    \label{Tab:ScanObject}
\end{table}

\paragraph{Classification on clean objects}
In this experiment, we fine-tune the pre-trained models to classify the ModelNet40 dataset~\cite{ModelNet2015}. Table~\ref{Tab:ModelNet} shows our results. We evaluate two scenarios: linear probing (classification directly from embeddings extracted by the pre-trained network) and fine-tuning (training with a multi-layer perceptron after the pre-trained encoder). For the fine-tuning evaluation, we follow the voting strategy~\cite{ModelNetVoting2019} to make our comparison fair with previous methods. 

Table~\ref{Tab:ModelNet} presents the classification performance on ModelNet40 under three evaluation protocols: k-NN, SVM, and fine-tuning. Overall, the results show that SymSSL, despite being a synthetic dataset, provides competitive pre-training signals compared to ShapeNet.

For Point-MAE, pre-training with SymSSL-10K leads to a notable improvement in k-NN accuracy (+3.3\%) compared to ShapeNet, indicating stronger local feature representations. While SVM and fine-tuning accuracy slightly decrease, the gap remains small (within 1.8\% and 1.3\%), showing that SymSSL can effectively support downstream learning, even with limited supervision. SymSSL-50K further improves fine-tuning performance to 93.4\%, nearly matching the 93.8\% of ShapeNet, demonstrating that increasing the size of SymSSL enhances generalization.

For PointGPT-S and -B, SymSSL-pre-trained models show a larger drop in linear evaluation (k-NN and SVM), particularly with the smaller 10K set. However, fine-tuning performance remains robust, with only a minor accuracy drop of 1.0\% (SymSSL-10K) and 0.6\% (SymSSL-300K) compared to ShapeNet. This suggests that while initial embeddings may be less discriminative, the representations learned from SymSSL still retain sufficient structure to support strong performance after task-specific tuning.

In summary, SymSSL proves to be a viable and scalable alternative to ShapeNet for pre-training, especially in settings where full supervision or curated datasets are unavailable. Its strengths are particularly evident in k-NN evaluations for Point-MAE and the consistently strong fine-tuning results across both models, highlighting its potential for cost-effective and flexible 3D representation learning. Furthermore, Symmetria's strong data efficiency allows to reach accuracies close to those obtained by pre-training on ShapeNet, but with only 10k samples instead of 50k.

\begin{table}[h]
    \centering\footnotesize
    \begin{tabular}{cclll}
    \hline
    \textbf{Model} & \textbf{\makecell{Pre-training\\dataset}} & \textbf{\makecell{k-NN\\Acc.}} & \textbf{\makecell{SVM\\Acc.}} & \textbf{\makecell{Fine-tuning\\Acc.}} \\ \hline
    Point-MAE & ShapeNet & 78.4 & 90.3 & 93.8 \\
    Point-MAE & SymSSL-10K & 81.7 \textcolor{blue}{(+3.3)} & 90.0 \textcolor{blue}{(-0.3)} & 92.5 \textcolor{blue}{(-1.3)}\\
    Point-MAE & SymSSL-50K & 78.0 \textcolor{blue}{(-0.4)} & 88.5 \textcolor{blue}{(-1.8)} & 93.4 \textcolor{blue}{(-0.4)} \\ \hline
    PointGPT-S & ShapeNet & 81.6 & 88.7 & 94.0 \\
    PointGPT-S & SymSSL-10K & 66.5 \textcolor{blue}{(-15.1)} & 76.4 \textcolor{blue}{(-12.3)} & 93.0 \textcolor{blue}{(-1.0)} \\
    PointGPT-S & SymSSL-50K &  72.4 \textcolor{blue}{(-9.2)} & 79.4 \textcolor{blue}{(-9.3)} & 93.2 \textcolor{blue}{(-0.8)} \\
    PointGPT-B & SymSSL-300K &  73.8 \textcolor{blue}{(-7.8)} & 82.7 \textcolor{blue}{(-6.0)} & 93.4 \textcolor{blue}{(-0.6)} \\
    PointGPT-B & SymSSL-1M &  71.5 \textcolor{blue}{(-10.1)} & 83.9 \textcolor{blue}{(-4.8)} & 93.1 \textcolor{blue}{(-0.9)} \\ \hline    
    \end{tabular}
   \caption{Classification accuracy on ModelNet using k-NN, SVM, and fine-tuning. SymSSL-10K outperforms ShapeNet in k-NN for Point-MAE, while PointGPT-S shows larger drops with SymSSL, especially in non-parametric settings. On the contrary, PointGPT-B pre-trained on SymSSL-300k and -1M shows a partially improving trend in both SVM classification and fine-tuning, although still inferior to that of ShapeNet.} Results highlight architecture-dependent effects of synthetic pre-training.
    \label{Tab:ModelNet}
\end{table}

\paragraph{Few-shot learning}
The few-shot learning experiment consists of evaluating the capacity of a pre-trained network to be fine-tuned and tested with limited data. We follow the setup proposed in previous works~\cite{PointMAE2022,PointBert2022}. Four sub-datasets are derived from the ModelNet dataset with the name $w-$way, $s-$shot for $w\in\{5,10\}$, and $s \in \{10,20\}$. Table~\ref{Table:FewShot} shows the results. The few-shot learning results demonstrate that SymSSL is a strong alternative to ShapeNet for pre-training, even under limited data scenarios. Models pre-trained on SymSSL consistently achieve competitive performance across all few-shot setups, with accuracy differences typically within 1–2\% of ShapeNet. Notably, SymSSL-10K already provides strong results, highlighting its data efficiency, while SymSSL-50K shows further improvements, indicating scalability. These findings confirm that SymSSL-pre-trained networks learn robust and transferable features, making SymSSL a cost-effective and scalable solution for pre-training.

\begin{table}[h]
    \centering\footnotesize
    \begin{tabular}{cccccc}
    \hline
        \textbf{Model} & \textbf{\makecell{Pre-training\\dataset}} & \textbf{\makecell{5-way\\10-shot}} & \textbf{\makecell{5-way\\20-shot}} & \textbf{\makecell{10-way\\10-shot}} & \textbf{\makecell{10-way\\20-shot}} \\ \hline
        Point-MAE & ShapeNet & 96.3 $\pm$ 2.5 & 97.8 $\pm$ 1.8 & 92.6 $\pm$ 4.1  & 95.0 $\pm$ 3.0\\
        Point-MAE & SymSSL-10K & 95.0 $\pm$ 2.0 & 97.0 $\pm$ 2.9 & 90.7 $\pm$ 5.1 &  93.7 $\pm$ 4.6\\
        Point-MAE & SymSSL-50K & 95.0 $\pm$ 2.4 & 97.0 $\pm$ 2.5 & 89.5 $\pm$ 4.8 & 94.0 $\pm$ 3.7 \\ \hline
        PointGPT-S & ShapeNet & 96.8 $\pm$ 2.0 & 98.6 $\pm$ 1.1 & 92.6 $\pm$ 4.6 & 95.2 $\pm$ 3.4\\
        PointGPT-S & SymSSL-10K & 94.0 $\pm$ 3.5 &  98.0 $\pm$ 2.1 & 90.5 $\pm$ 5.6 & 94.0 $\pm$ 4.2\\
        PointGPT-S & SymSSL-50K & 95.7 $\pm$ 3.2 & 98.0 $\pm$ 1.5 & 91.1 $\pm$ 4.9 & 94.3 $\pm$ 4.0\\ \hline
    \end{tabular}
    \caption{Few-shot classification accuracy on ModelNet for Point-MAE and PointGPT-S pre-trained on ShapeNet and SymSSL. SymSSL pre-training achieves competitive results across all settings, showing strong transferability of synthetic representations even in low-data regimes.}
    \label{Table:FewShot}
\end{table}

\paragraph{Part segmentation}
In this experiment, we fine-tune the pre-trained models to predict a pointwise classification label for segmentation. We use the ShapeNetPart~\cite{ShapeNetPart2016} dataset that contains 16,881 objects annotated in 16 segment classes. The metric used in our evaluation is mean Intersection over Union (mIoU) averaged by class (Cls. mIoU) and averaged by instance (Inst. mIoU). Table~\ref{Tab:PartSegmentation} shows the results. 

The results demonstrate that SymSSL-pre-trained models perform comparably to ShapeNet-pre-trained models in the task of part segmentation on ShapeNetPart. For Point-MAE, the instance-level mIoU drops by only 0.2\% when switching from ShapeNet to SymSSL, with virtually no performance difference between the 10K and 50K versions. Although class-level mIoU for the ShapeNet baseline is not available, SymSSL-pre-trained models still achieve strong results (84.2–84.4\%), suggesting good class-wise generalization.

For PointGPT-S, the differences between ShapeNet and SymSSL pre-training are minimal. The instance mIoU decreases by just 0.4\%, and class mIoU drops by only 0.6\%. These small gaps confirm that SymSSL enables the learning of fine-grained, transferable features that generalize well to dense prediction tasks like segmentation.

Importantly, these results highlight that SymSSL provides high-quality pre-training signals without the need for curated datasets, achieving near-identical downstream segmentation performance. The consistent performance across both models and both dataset sizes further underscores SymSSL’s robustness and scalability for 3D representation learning.

In summary, SymSSL is a strong, scalable, and practical alternative to ShapeNet for pre-training, maintaining high segmentation accuracy while reducing reliance on manually annotated datasets.

\begin{table}[h]
    \centering\footnotesize
    \begin{tabular}{ccll}
    \hline
    \textbf{Model} & \textbf{Pre-training dataset} & \textbf{Cls. mIoU(\%)} & \textbf{Inst. mIoU(\%)} \\ \hline
    Point-MAE & ShapeNet & - & 86.1\\ 
    Point-MAE & SymSSL-10K & 84.2 \textcolor{blue}{(N.A.)} & 85.9 \textcolor{blue}{(-0.2)}\\
    Point-MAE & SymSSL-50K & 84.4 \textcolor{blue}{(N.A.)} & 85.9 \textcolor{blue}{(-0.2)}\\
    \hline
    PointGPT-S & ShapeNet & 84.1 & 86.2 \\ 
    PointGPT-S & SymSSL-10K & 83.5 \textcolor{blue}{(-0.6)} & 85.8 \textcolor{blue}{(-0.4)} \\
    PointGPT-S & SymSSL-50K & 83.6 \textcolor{blue}{(-0.6)}& 85.8 \textcolor{blue}{(-0.4)}\\ \hline
    \end{tabular}
    \caption{Part segmentation results comparing pre-training on ShapeNet and SymSSL (10K and 50K) using two models: Point-MAE and PointGPT-S. We report class-averaged (Cls. mIoU) and instance-averaged (Inst. mIoU) mean Intersection over Union on ShapeNetPart. Results in blue indicate the difference relative to the ShapeNet baseline. SymSSL pre-training yields competitive performance across both models, with minimal drop in instance mIoU ($\leq 0.4$), even when using synthetic data exclusively.}
    \label{Tab:PartSegmentation}
\end{table}


\paragraph{Understanding the Role of Symmetry in Pre-training}
Previous experiments demonstrate that our dataset enables effective pre-training of point cloud neural networks, yielding competitive downstream performance compared to models pre-trained on ShapeNet. However, an open question remains: \emph{what is the relationship between the symmetries embedded in our dataset and the network's ability to discriminate between symmetric objects in downstream tasks?} To investigate this, we designed an experiment aimed at clarifying the correlation between dataset symmetry and learned model representations.

We selected the Point-MAE model pre-trained with SymSSL-10K for this analysis. The network was frozen, and we evaluated class-wise classification performance on the ModelNet dataset using a k-NN classifier applied directly to the raw embeddings extracted from the pre-trained encoder. This linear evaluation protocol allows us to assess the quality of the learned features without any further fine-tuning, providing insight into the information encoded in the latent space. Notably, as shown in Table 6, this setup revealed that SymSSL-10K embeddings achieved 3.3\% higher classification accuracy than those from the ShapeNet-pre-trained model, highlighting the strength of the learned representations.

\begin{figure}[h!]
    \centering
    \includegraphics[width=\textwidth]{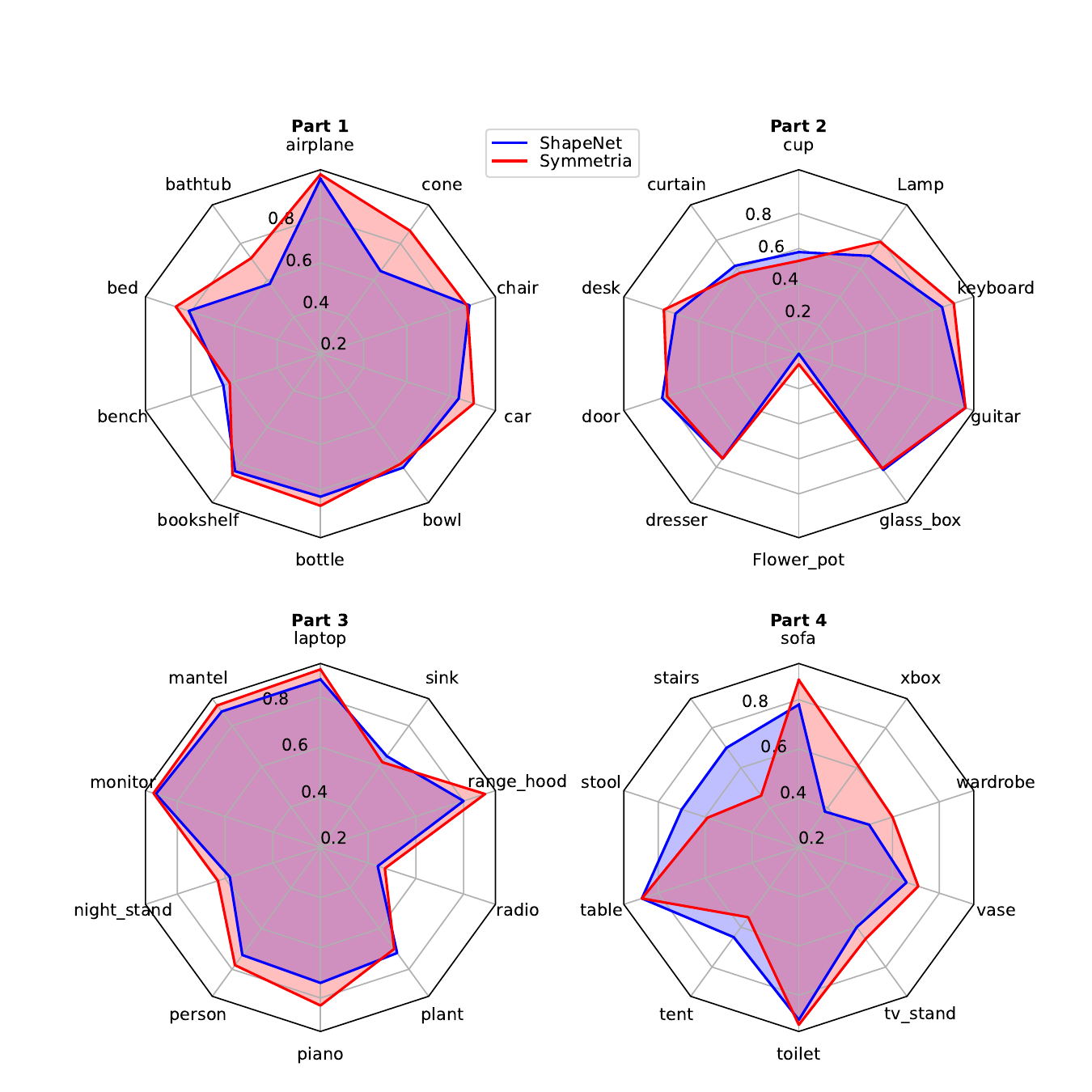}
  \caption{Class-level F1 measure for test set in ModelNet. The base model is Point-MAE. The model is pre-trained with ShapeNet (blue) and SymSSL-10K (red).}
  \label{Fig:radar}
\end{figure}

To further explore whether this improvement is linked to the presence of symmetrical structures in our dataset, we conducted an analysis based on per-class F1-score on ModelNet. Figure~\ref{Fig:radar} presents these results. We observe that certain classes, such as cone, bathtub, and xbox, are better classified when models are pre-trained with SymSSL, while others, like curtain, cup, stool, and stairs, benefit more from ShapeNet pre-training. For the remaining classes, performance differences are minor, though SymSSL generally outperforms ShapeNet.

To dig into the factors influencing these class-level differences, we designed an experiment to estimate the correlation between classification improvement and the presence of symmetry in the test classes. Given that ModelNet objects are approximately aligned with canonical 3D axes, their principal symmetry planes are often aligned with the XY, YZ, or XZ planes. We define an asymmetry score that quantifies the geometric discrepancy between a point cloud and its mirrored versions across these canonical planes. For a point cloud $P$, the asymmetry score is defined as:

\begin{equation}
	assymetry(P) = \min( {d(P, Ref_{XY}(P)), d(P, Ref_{YZ}(P)), d(P, Ref_{XZ}(P)))}
\end{equation}

\noindent where $d$ is the Chamfer distance, and $Ref$ denotes the reflection of the point cloud across a given plane. Since the objects are well-aligned, the minimum Chamfer distance among the three reflections provides a good approximation of how symmetric the object is.

\begin{figure}[h!]
    \centering
    \includegraphics[width=\textwidth]{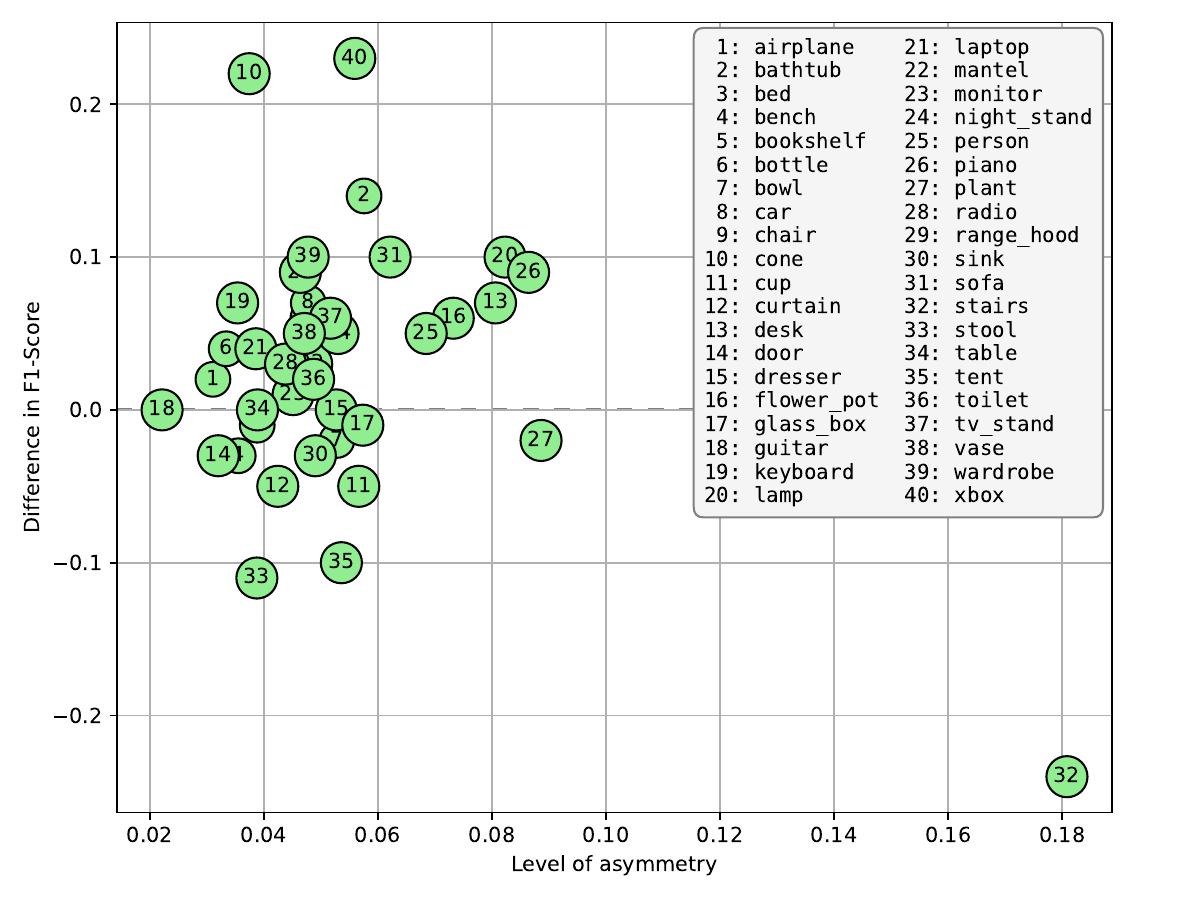}
  \caption{Per-class relationship between level of asymmetry and difference in F1 measure (Symmetria vs ShapeNet). The F1 score is computed on the ModelNet's test set.}
  \label{Fig:asymmetry}
\end{figure}

We computed the average asymmetry score for each class in the ModelNet test set, thereby estimating the class-level symmetry. Figure~\ref{Fig:asymmetry} plots the relationship between the difference in F1-score (SymSSL vs. ShapeNet) and the average asymmetry level of each class. Positive F1-score differences indicate better performance with SymSSL pre-training. The results provide a clear explanation for the trends observed in Figure~\ref{Fig:radar}: classes like bathtub, cone, and xbox, exhibit very low asymmetry scores (i.e., they are highly symmetric), and show substantial performance gains with SymSSL pre-training. In contrast, classes like stairs and plant, which are more asymmetric, do not benefit as much.

These findings suggest that SymSSL pre-training enables neural networks to learn features that capture object symmetry, which can be leveraged in downstream tasks like classification. This behavior is particularly valuable since many real-world, human-designed objects tend to exhibit some degree of symmetry, making SymSSL a practical and effective dataset for pre-training 3D models.

\paragraph{Ablation Study on Curve Families and Perturbations}

We also evaluated the impact of the selected curve families and the decision to apply geometric perturbations to the objects. As a baseline, we used the Point-MAE model pre-trained on the full SymSSL-10K dataset, as described earlier in this section. In addition, we generated three alternative versions of the dataset to isolate the effects of curve complexity and perturbations:

\begin{itemize}
    \item \textbf{SymSSL-10K – only clean – all curves}: This version contains 10,000 randomly generated objects using all available curve families but without any perturbations. It serves to evaluate the effect of training on clean, unmodified shapes.
    \item \textbf{SymSSL-10K – all transf. – subset simple}: This version also contains 10,000 randomly generated objects but includes all perturbations while limiting the curve families to a subset with simpler geometries: astroid, citrus, cylinder, egg of Kepler, mouth curve, and square. These curves have fewer geometric features and are more regular in shape.
    \item \textbf{SymSSL-10K – all transf. – subset complex}: This version includes 10,000 objects with all perturbations applied, but only uses a subset of curves with more complex geometries: m-convexities, geometric petal, and lemniscate. These curve families exhibit higher geometric variability and structural richness.
\end{itemize}

After pre-training Point-MAE on each of these dataset variants, we evaluated the classification accuracy on the ModelNet test set to analyze how the composition and complexity of the pre-training data affect downstream performance. Table~\ref{Tab:ModelNet_Ablation} shows the results.

\begin{table}[h]
    \centering
    
    \begin{tabular}{ccccc}
    \hline
    \textbf{Model} &
    \textbf{\makecell{Pre-training\\dataset}} & \textbf{\makecell{k-NN\\Acc.}} & \textbf{\makecell{SVM\\Acc.}} & \textbf{\makecell{Fine-tuning\\Acc.}} \\ \hline
    Point-MAE & ShapeNet & 78.4 & 90.3 & 93.8 \\
    Point-MAE & SymSSL-10K & 81.7 & 90.0 & 92.5 \\
    Point-MAE & SymSSL-10K - only clean - all curves & 78.6 & 88.9 & 93.2 \\ 
    Point-MAE & SymSSL-10K - all transf. - subset simple & 78.6 & 88.0 & 92.9 \\ 
    Point-MAE & SymSSL-10K - all transf. - subset complex & 79.6 & 89.1 & 93.0 \\ \hline
    \end{tabular}
    \caption{Ablation study on the effects of curve family diversity and geometric perturbations in SymSSL pre-training. We report classification accuracy on ModelNet using k-NN, SVM, and fine-tuning. The full SymSSL-10K dataset (all curves + all transformations) yields the highest k-NN accuracy, highlighting the benefits of combining geometric complexity with data variability during pre-training.}
    \label{Tab:ModelNet_Ablation}
\end{table}

The results of the ablation study show a clear trend: pre-training with the full SymSSL-10K dataset—comprising all curve families and all perturbations—leads to the most robust and generalizable representations, particularly when evaluated through non-parametric methods like k-NN classification.

Examining the dataset variants helps clarify why the full SymSSL-10K is most effective:

\begin{itemize}
    \item The "only clean – all curves" version yields much lower k-NN and SVM accuracies (78.6\% and 88.9\%, respectively), suggesting that the absence of perturbations limits the model’s ability to generalize to real-world shape variations. Although its fine-tuning accuracy remains high (93.2\%), this suggests that downstream adaptation can compensate for some limitations of the initial embeddings.
    \item The "all transf. – subset simple" variant performs similarly in k-NN accuracy (78.6\%) and slightly worse in SVM accuracy (88.0\%), despite including transformations. This indicates that limiting the geometric complexity of curve families reduces the diversity and richness of the learned features.
    \item The "all transf. – subset complex" variant improves upon the simple subset (79.6\% k-NN and 89.1\% SVM), supporting the hypothesis that curve complexity contributes positively to representation learning, especially when combined with geometric variability.
\end{itemize}

Overall, these findings suggest that the synergy between curve diversity and geometric perturbations is critical. Pre-training on a dataset that includes both factors—as in the full SymSSL-10K—enables the model to develop more robust, invariant, and transferable representations, leading to superior performance in classification tasks with minimal supervision.

To further evaluate the generality of representations learned through SymSSL pre-training, we conducted experiments on the part segmentation task. Table~\ref{Tab:PartSegmentation_Ablation} reports both class-averaged and instance-averaged mIoU on the ShapeNetPart dataset.

Ablation variants reveal consistent trends with our classification analysis. Removing geometric perturbations (“only clean – all curves”) slightly degrades performance (83.7\% class mIoU, 85.6\% instance mIoU), suggesting that training on unmodified shapes limits robustness. Interestingly, both the “subset simple” and “subset complex” variants achieve similar results (84.1\% class mIoU, 85.8\% instance mIoU), indicating that, unlike in classification, curve complexity has a reduced impact on segmentation performance. This may reflect the more localized nature of part segmentation, which relies less on global shape structure and more on fine-grained geometric cues.

Overall, these results reinforce the conclusion that combining geometric variability with synthetic diversity yields transferable representations, capable of supporting both global (classification) and local (segmentation) 3D understanding.

\begin{table}[h]
    \centering
    \begin{tabular}{cccc}
    \hline
    \textbf{Model} &
    \textbf{Pre-training dataset} & \textbf{Cls. mIoU(\%)} & \textbf{Inst. mIoU(\%)} \\
    \hline
    Point-MAE & SymSSL-10K & 84.2 & 85.9 \\
    Point-MAE & SymSSL-10K - only clean - all curves & 83.7 & 85.6 \\
    Point-MAE & SymSSL-10K - all transf. - subset simple & 84.1 & 85.8 \\
    Point-MAE & SymSSL-10K - all transf. - subset complex & 84.1 & 85.8 \\
    \hline
    \end{tabular}
    \caption{Part segmentation results on ShapeNetPart. We report class-averaged (Cls. mIoU) and instance-averaged (Inst. mIoU) mean Intersection over Union. Ablation variants reveal the importance of geometric perturbations, while curve complexity has a limited impact on segmentation performance.}
    \label{Tab:PartSegmentation_Ablation}
\end{table}

\paragraph{Effect of Dataset Scale on Pre-training Performance}

To assess the impact of dataset scale on the quality of learned representations, we conducted a scaling experiment using four versions of our synthetic SymSSL dataset: 10K, 50K, 300K, and 1M objects. These were used to pre-train the Point-MAE model under the same configuration. We then evaluated the resulting models on two downstream tasks: classification on ModelNet (Table~\ref{Tab:ModelNet_Scale}) and part segmentation on ShapeNetPart (Table~\ref{Tab:PartSegmentation_Scale}). This study aims to understand how the volume of pre-training data interacts with model performance across different types of tasks.

\begin{table}[h]
    \centering
    \begin{tabular}{ccccc}
    \hline
    \textbf{Model} & \textbf{\makecell{Pre-training\\dataset}} & \textbf{\makecell{k-NN\\Acc.}} & \textbf{\makecell{SVM\\Acc.}} & \textbf{\makecell{Fine-tuning\\Acc.}} \\ \hline
    Point-MAE & ShapeNet & 78.4 & 90.3 & 93.8 \\
    Point-MAE & SymSSL-10K & 81.7 & 90.0 & 92.5 \\
    Point-MAE & SymSSL-50K & 78.6 & 88.9 & 93.2 \\ 
    Point-MAE & SymSSL-300K & 74.5 & 82.4 & 92.3 \\ 
    Point-MAE & SymSSL-1M & 73.3 & 86.1 & 92.8 \\ \hline
    \end{tabular}
    \caption{ModelNet classification results at different SymSSL dataset scales.}
    \label{Tab:ModelNet_Scale}
\end{table}

The classification results reveal a non-linear relationship between dataset size and downstream performance:
\begin{itemize}
    \item The model pre-trained on SymSSL-10K achieves the best k-NN accuracy (81.7\%), outperforming ShapeNet (78.4\%) and all larger-scale SymSSL variants.
    \item As dataset size increases beyond 10K, performance in both k-NN and SVM classification declines, reaching 73.3\% and 86.1\%, respectively, for SymSSL-1M.
    \item Fine-tuning accuracy remains relatively stable (between 92.3\%, and 93.2\%), indicating that the model retains sufficient adaptability with additional supervision.
\end{itemize}

This performance drop in linear evaluation (k-NN/SVM) suggests that larger datasets do not automatically lead to better embeddings. A likely explanation is that, without proper scaling of model capacity, the network may underfit the growing diversity and complexity introduced by larger datasets. With a fixed model size (Point-MAE), the network may not have sufficient representational power to fully benefit from the increased shape variability and geometric richness present in SymSSL-300K and SymSSL-1M.

\begin{table}[h]
    \centering
    \begin{tabular}{cccc}
    \hline
    \textbf{Model} & \textbf{Pre-training dataset} & \textbf{Cls. mIoU(\%)} & \textbf{Inst. mIoU(\%)} \\ \hline
    Point-MAE & SymSSL-10K & 84.2 & 85.9 \\
    Point-MAE & SymSSL-50K & 83.7 & 85.6 \\
    Point-MAE & SymSSL-300K & 83.6 & 85.9 \\
    Point-MAE & SymSSL-1M & 84.5 & 85.9 \\
    \hline
    \end{tabular}
    \caption{Part segmentation results on ShapeNetPart across SymSSL dataset scales.}
    \label{Tab:PartSegmentation_Scale}
\end{table}

In contrast to classification, segmentation results remain robust across dataset scales:

\begin{itemize}
    \item The instance-level mIoU (Inst. mIoU) stays consistently high (~85.9\%) across all dataset sizes, showing no degradation with scale.
    \item The class-level mIoU (Cls. mIoU) remains stable, with SymSSL-1M even showing a slight improvement (84.5\%) over smaller datasets.
\end{itemize}

This robustness in segmentation performance suggests that dense prediction tasks can better exploit large-scale pre-training data, possibly because they benefit from the broader distribution of object parts and local geometric configurations. However, the gains are relatively modest, pointing again to potential capacity limitations.

The degradation in classification accuracy at higher scales indicates a mismatch between dataset complexity and model capacity. While small-scale datasets like SymSSL-10K may be well-suited to compact encoders like Point-MAE, larger datasets such as SymSSL-1M likely contain richer, more diverse geometric structures that require models with greater capacity to fully exploit.

This observation aligns with trends in the vision and language domains, where scaling data and model size together is crucial for performance gains. Thus, the current architecture may be underpowered for 1M-scale pre-training, leading to underfitting and diminished embedding quality in shallow evaluations (k-NN, SVM). Future work should explore scaling the architecture proportionally, either by increasing depth, width, or attention capacity, to better match the information density of large-scale synthetic datasets like SymSSL.

\subsection{Symmetry Detection}\label{sec:symmetrydetection}

The second goal of our research is to introduce Symmetria as a challenging dataset for symmetry detection. To accomplish this, we created four versions of the dataset, each with increasing complexity, to evaluate the performance of symmetry detection algorithms. We trained a baseline model from scratch on each version (\textit{easy}, \textit{intermediate-1}, \textit{intermediate-2}, \textit{hard}), and we present findings that demonstrate the progressive challenge posed by each dataset variation in the plane symmetry detection task.

\subsubsection{Employed datasets}\label{sec:symmetrydetectiondatasets}

As introduced in \ref{sec:dataset}, each of the four sub-datasets generated for the supervised learning experimentation comes in two distinct sizes: \textit{10k} (small) which represents a reduced version of the dataset suitable for exploratory data analysis or for quick experimentation and \textit{100k} (large) to be used for experimenting with established architectures to be trained up to the maximum possible level of accuracy. These numbers refer to the \textbf{total number of samples} within each sub-dataset. The key differences between these four sub-datasets are the following:

\begin{itemize}
    \item \textbf{Easy}: this basic portion of the dataset includes 7 classes: \textit{Astroid}, \textit{Citrus}, \textit{Egg of Keplero}, \textit{Geometric Petal}, \textit{Lemniscate of Bernoulli}, \textit{m-Convexities} and  \textit{Mouth curve}. Random rotations are applied with a probability of 0.5 and exclusively around the x-axis.
    \item \textbf{Intermediate-1}: the only difference between the \textit{easy} and the \textit{intermediate-1} sub-datasets is that two classes are added to the latter: \textit{Square} and \textit{Cylinder} (with the related conics).
    \item \textbf{Intermediate-2}: the shapes in the \textit{intermediate-2} sub-dataset are subjected to the same transformations as in \textit{intermediate-1} but are rotated with a probability of 0.75 around two axes, $x$ and $y$, instead of one.
    \item \textbf{Hard}: \textit{hard} sub-dataset has one more class, \textit{Revolution}, for a total of $10$ classes, and its shapes are rotated around all three axes with probability 1.0.
\end{itemize}

\subsubsection{Network architecture}\label{sec:symmetrydetectionnetwork}

The architecture for our supervised learning experiments is built on a PointNet encoder~\cite{qi2017pointnet}, which generates a 1024-dimensional feature representation that is then processed by multiple specialized PredictionHead modules. Each PredictionHead reduces this representation to 256 dimensions via a three-layer MLP and outputs values specific to the task, such as plane symmetry regression, which uses a 4-neuron output layer. Focusing on plane normals, 32 PredictionHeads were found optimal, providing enough capacity to learn features effectively. Each PredictionHead outputs three values for the normal vector and a confidence score, while a separate CenterPredictionHead predicts the central point coordinates for each normal using a single hidden layer and a 3-neuron output layer.

\paragraph{Loss function}
Detecting, predicting, and regressing the parameters of symmetry planes is a notably intricate and difficult task. We tackled this issue by adopting the methodology outlined in \textit{SymmetryNet} \cite{shi_symmetrynet_2020}. To match predicted symmetries to real ones, we employed the optimal assignment strategy utilized by SymmetryNet. The overall loss employs a multitask learning framework, aiming for the model to not only predict symmetries but also estimate the confidence in its predictions. The details on architecture and loss function are provided in the
Appendix (\ref{sec:sym_det_loss_decomposition}).

\subsubsection{Evaluation measures}\label{sec:evaluationmeasures}
We used the following metrics defined in the SHREC2023 dataset~\cite{shrec2023}: \textit{mean Average Precision} (mAP) and \textit{Precision at Highest Confidence} (PHC). Briefly, mAP quantifies how well the predicted symmetry planes match the planes in the ground truth. If the normal and position deviations of a prediction are below a certain threshold with respect to the ground-truth plane, the predicted symmetry is considered to be a true positive. Perfect matching gives an mAP of one, and no matching gives an mAP of zero. In contrast, PHC measures the ratio of shapes for which at least one prediction is correct.

\subsubsection{Experimental Results}\label{sec:experimental_results_supervised}

In our experimentation, we employed PyTorch-Lightning as the primary machine learning library for model training, automatically ensuring a very good level of reproducibility of our results. The hyperparameters selected for our models were as follows: learning rate (LR) of \(1 \times 10^{-3}\), epsilon (eps) of \(1 \times 10^{-8}\), no weight decay. For the training of our final baseline models, we utilized two different computational infrastructures based on the dataset size: \textit{i)} models trained on the Symmetry-10k sub-datasets were executed on an NVIDIA RTX 2070 Super GPU with 8 GB of memory. Each training epoch took approximately 20 minutes to complete, with a validation period lasting around 4 minutes per epoch. \textit{ii)} models trained on the Symmetry-100k sub-datasets were run on two NVIDIA RTX 3090 GPUs with 24 GB of memory each. The total training for these experiments took approximately 80 hours.

In Table \ref{tab:supervisedresults}, we report the results of the best training runs for the different Symmetria sub-datasets; for further details,  refer to the extended analysis of the results in the Appendix (Sec. \ref{sec:extendedanalysis}). 

\begin{table}[h]
\centering\footnotesize
\begin{tabular}{|c|lll|lll|cc|}
\hline
\multirow{2}{*}{\textbf{Dataset}} &
  \multicolumn{3}{c|}{\textbf{Train}} &
  \multicolumn{3}{c|}{\textbf{Validation}} &
  \multicolumn{2}{c|}{\textbf{Test}} \\
 &
  \multicolumn{1}{c|}{\textbf{mAP}} &
  \multicolumn{1}{c|}{\textbf{PHC}} &
  \multicolumn{1}{c|}{\textbf{Loss}} &
  \multicolumn{1}{c|}{\textbf{mAP}} &
  \multicolumn{1}{c|}{\textbf{PHC}} &
  \multicolumn{1}{c|}{\textbf{Loss}} &
  \multicolumn{1}{c|}{\textbf{mAP}} &
  \textbf{PHC} \\ \hline
Easy-10k &
  \multicolumn{1}{l|}{0.850} &
  \multicolumn{1}{l|}{0.861} &
  0.345 &
  \multicolumn{1}{l|}{0.865} &
  \multicolumn{1}{l|}{0.890} &
  0.302 &
  \multicolumn{1}{c|}{0.877} &
  0.872 \\ 
Intermediate-1-10k &
  \multicolumn{1}{l|}{0.775} &
  \multicolumn{1}{l|}{0.791} &
  0.326 &
  \multicolumn{1}{l|}{0.807} &
  \multicolumn{1}{l|}{0.831} &
  0.329 &
  \multicolumn{1}{c|}{0.787} &
  0.816 \\ 
Intermediate-2-10k &
  \multicolumn{1}{l|}{0.213} &
  \multicolumn{1}{l|}{0.205} &
  0.887 &
  \multicolumn{1}{l|}{0.226} &
  \multicolumn{1}{l|}{0.211} &
  0.869 &
  \multicolumn{1}{c|}{0.228} &
  0.216 \\ 
Hard-10k &
  \multicolumn{1}{l|}{0.004} &
  \multicolumn{1}{l|}{0.001} &
  1.062 &
  \multicolumn{1}{l|}{0.004} &
  \multicolumn{1}{l|}{0.003} &
  1.047 &
  \multicolumn{1}{c|}{0.002} &
  0.000 \\ \hline
Easy-100k &
  \multicolumn{1}{l|}{0.839} &
  \multicolumn{1}{l|}{0.844} &
  0.227 &
  \multicolumn{1}{l|}{0.852} &
  \multicolumn{1}{l|}{0.871} &
  0.189 &
  \multicolumn{1}{c|}{0.840} &
  0.858 \\ 
Intermediate-1-100k &
  \multicolumn{1}{l|}{0.769} &
  \multicolumn{1}{l|}{0.764} &
  0.265 &
  \multicolumn{1}{l|}{0.788} &
  \multicolumn{1}{l|}{0.800} &
  0.232 &
  \multicolumn{1}{c|}{0.759} &
  0.740 \\ 
Intermediate-2-100k &
  \multicolumn{1}{l|}{0.292} &
  \multicolumn{1}{l|}{0.261} &
  0.521 &
  \multicolumn{1}{l|}{0.320} &
  \multicolumn{1}{l|}{0.279} &
  0.467 &
  \multicolumn{1}{c|}{0.320} &
  0.280 \\ 
Hard-100k &
  \multicolumn{1}{l|}{0.002} &
  \multicolumn{1}{l|}{0.000} &
  1.069 &
  \multicolumn{1}{l|}{0.004} &
  \multicolumn{1}{l|}{0.000} &
  1.034 &
  \multicolumn{1}{l|}{0.002} &
  \multicolumn{1}{l|}{0.000} \\ 
  \hline
\end{tabular}
\caption{Best training, validation and test metrics for each of the training runs on the different Symmetria sub-datasets.}
\label{tab:supervisedresults}
\end{table}

\paragraph{Discussion of results}

Empirical analysis of the baseline models on the proposed multitask learning framework reveals a critical dependency on the complexity of shape rotations. While exceptional performance is achieved on the \textit{Easy} and \textit{Intermediate-1} sub-datasets (mAP/PHC > 0.8), characterized by single-axis rotations, a significant performance degradation is observed on the \textit{Intermediate-2} sub-dataset (mAP/PHC > 0.2) with two-axis rotations. This suggests that the current problem formulation might be insufficient for handling more complex rotations. Furthermore, the \textit{Hard} sub-dataset, exhibiting arbitrary rotations along three axes, proved to be virtually intractable under the current problem formulation, necessitating the exploration of more elaborate and advanced strategies.

\subsubsection{Ablation study}\label{sec:ablation_study_supervised}


In this section, we present our ablation experiments aimed at identifying the fundamental differences between the \textit{Symmetria Easy} and \textit{Hard} datasets and understanding why the latter poses significant challenges for the baseline \textit{PointNet} architecture. All experiments were conducted on single-class datasets containing 2,000 samples each, following our standard 70/20/10 train/validation/test split. Training was performed on 1,400 samples per dataset. The \textit{Symmetria} benchmark comprises 10 classes: \textit{Astroid}, \textit{Citrus}, \textit{Cylinder}, \textit{Egg of Keplero}, \textit{Geometric Petal}, \textit{Lemniscate}, \textit{m-Convexities}, \textit{Mouth curve}, \textit{Revolution}, and \textit{Square}. We evaluated performance using \textit{validation loss}, \textit{validation Average Precision} (mAP), and \textit{validation Precision at Highest Confidence} (PHC) as metrics. The following paragraphs describe three experimental groups: 1) subsampling/noise transformations with standard PointNet, 2) rotation transformations with standard PointNet, and 3) rotation transformations with an upscaled PointNeXt XXL encoder.

Six classes (\textit{Astroid}, \textit{Citrus}, \textit{Geometric Petal}, \textit{Lemniscate}, \textit{Mouth curve}, \textit{Square}) exhibit similar trends - we show \textit{Astroid} as representative. The remaining four classes (\textit{Cylinder}, \textit{Egg of Keplero}, \textit{m-Convexities}, \textit{Revolution}) display unique behaviors and are analyzed individually. The complete results, metrics, and the script used to generate the graphs in this section are available at: \url{http://deeplearning.ge.imati.cnr.it/symmetria}.

\paragraph{Undersampling and noise}

\newcommand{\includegraphicsundersampling}[2][]{{\includegraphics[width=#1\textwidth,trim={0.0cm 0cm 0.0cm 0cm},clip]{./#2}}}
\begin{figure*}[tbp]
\centering
\begin{tabular}{cc}
\includegraphicsundersampling[0.95]{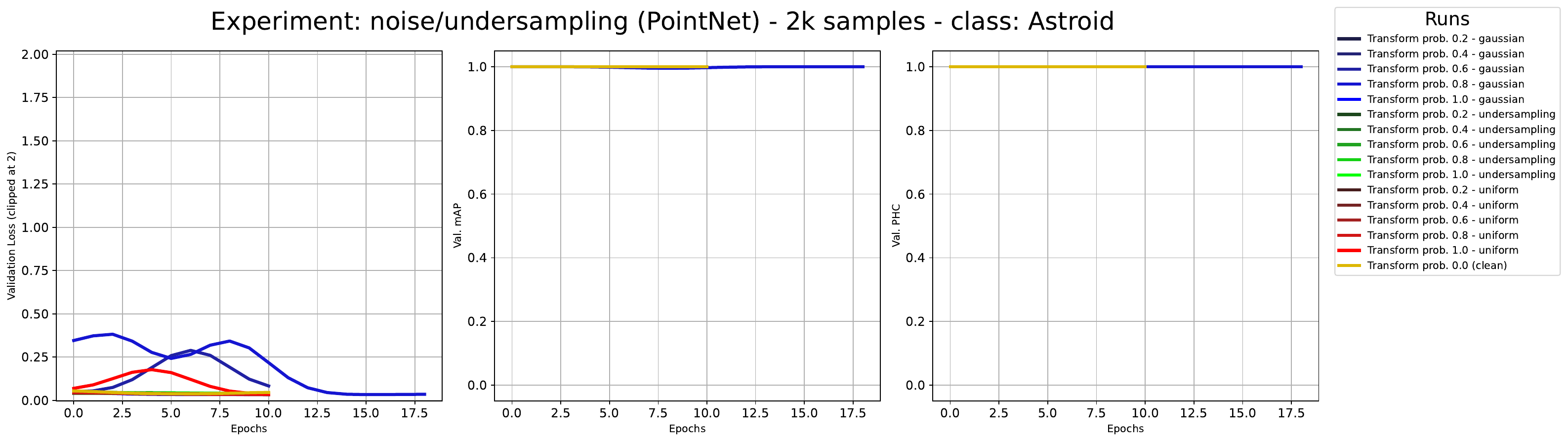}       \\
\includegraphicsundersampling[0.95]{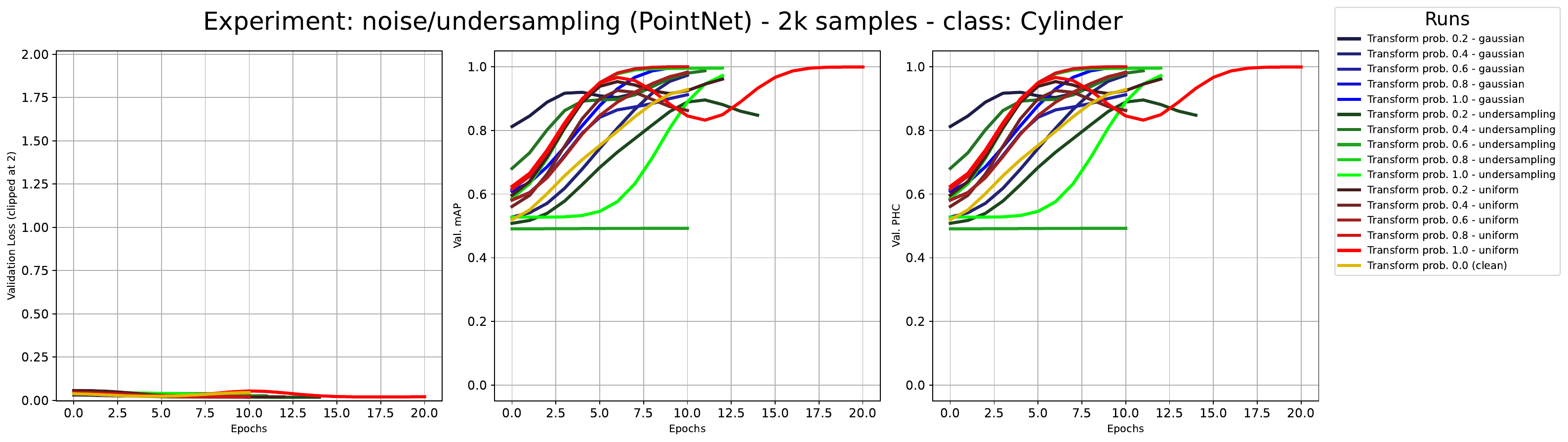}      \\
\includegraphicsundersampling[0.95]{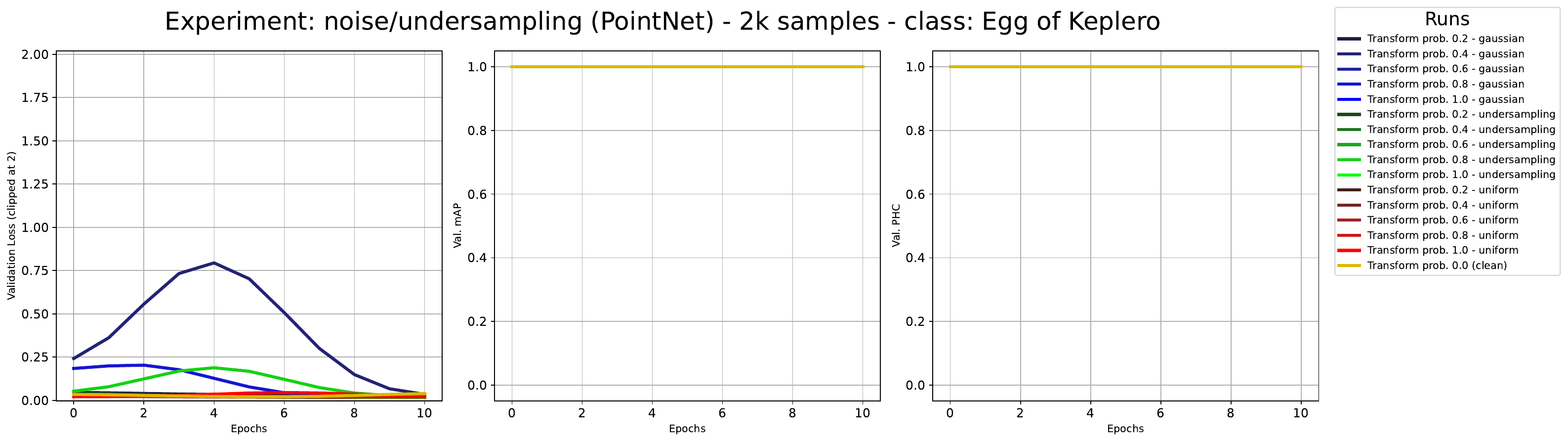}   \\
\includegraphicsundersampling[0.95]{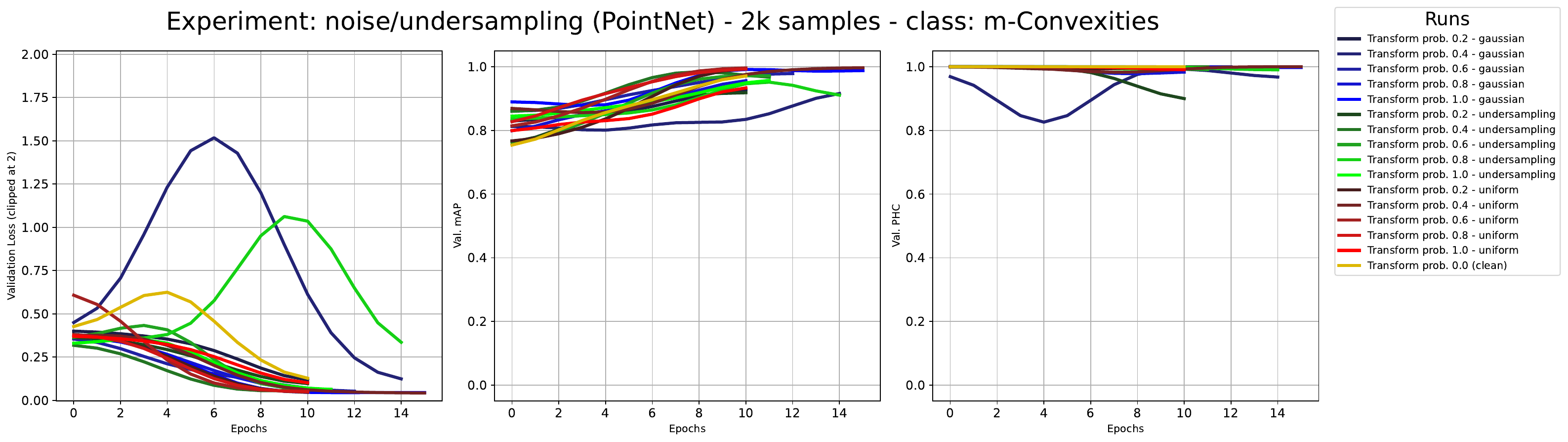} \\
\includegraphicsundersampling[0.95]{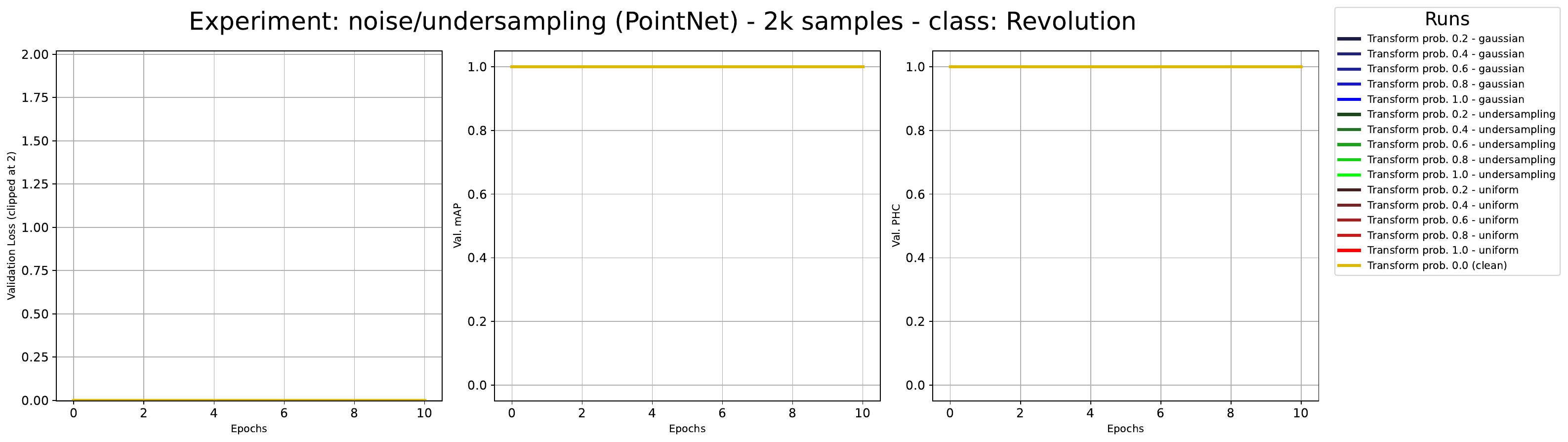}    \\
\end{tabular}
\caption{Ablation study: training on noisy (Uniform, Gaussian) or undersampled data using a standard \textit{PointNet} encoder. Classes: Astroid, Cylinder, Egg of Keplero, m-Convexities, Revolution.}
\label{tab:ablation_noise_undersampling_grouped}
\end{figure*}


The first experiment evaluates the impact of uniform/Gaussian noise and point cloud subsampling. As shown in Fig.~\ref{tab:ablation_noise_undersampling_grouped}, we generated datasets with fixed size (2,000 samples) but varying transformation probabilities (0.0 to 1.0). Transformations include \textit{undersampling}, \textit{uniform noise}, and \textit{Gaussian noise}. The 0.0 probability case (clean dataset) is shown in gold. Red, blue, and green lines denote Gaussian noise, uniform noise, and subsampling, respectively. Color intensity indicates transformation probability (darker: low, brighter: high).


Fig.~\ref{tab:ablation_noise_undersampling_grouped} reveals that noise and subsampling minimally impact model performance, even at maximum probability. Loss curves (left panels) remain near zero, while mAP and PHC (center/right panels) stay close to 1.0. Minor deviations occur for \textit{Cylinder}, \textit{Geometric Petal}, and \textit{m-Convexities}. The clean \textit{Cylinder} dataset achieves maximum mAP/PHC $\approx$ 0.9, likely due to its single symmetry (degenerate case). \textit{Geometric Petal} and \textit{m-Convexities} classes (containing up to 14 and 12 symmetries, respectively) show slightly reduced performance, suggesting higher symmetry complexity challenges network learning.

\paragraph{Rotations with standard PointNet encoder}

\newcommand{\includegraphicsrotations}[2][]{{\includegraphics[width=#1\textwidth,trim={0.0cm 0cm 0.0cm 0cm},clip]{./#2}}}
\begin{figure*}[tbp]
\centering
\begin{tabular}{cc}
\includegraphicsrotations[0.95]{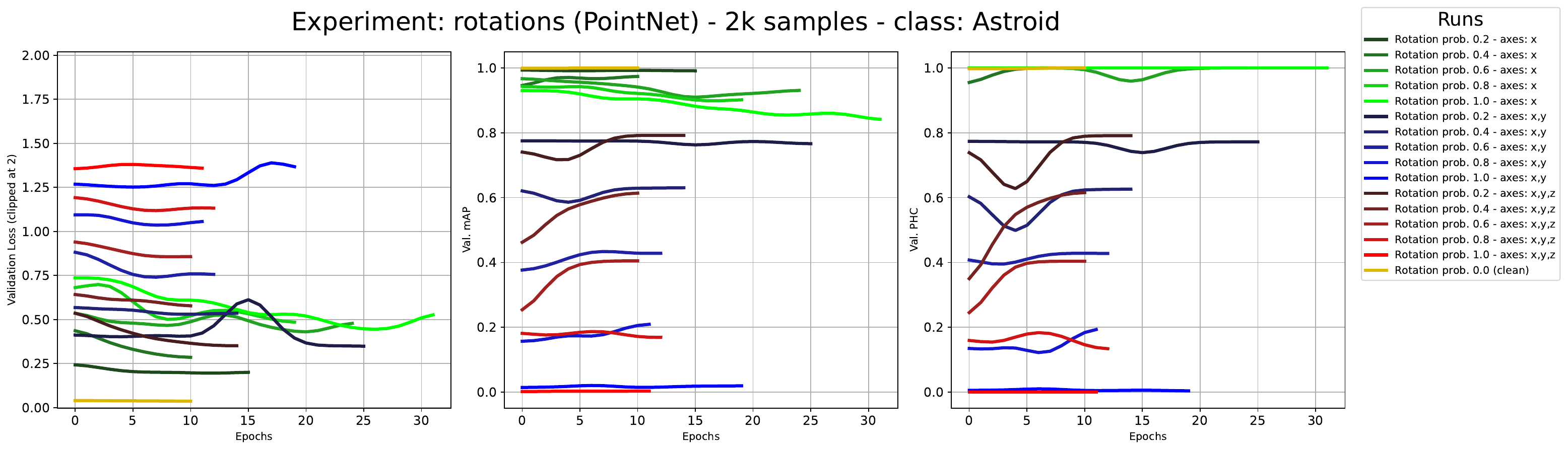}       \\
\includegraphicsrotations[0.95]{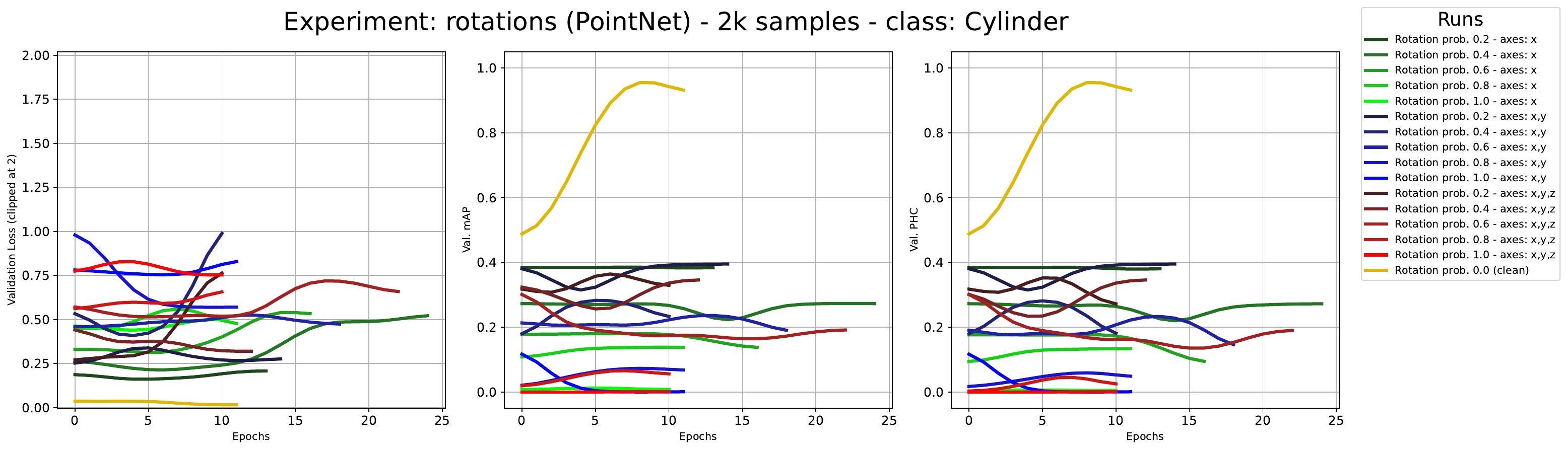}      \\
\includegraphicsrotations[0.95]{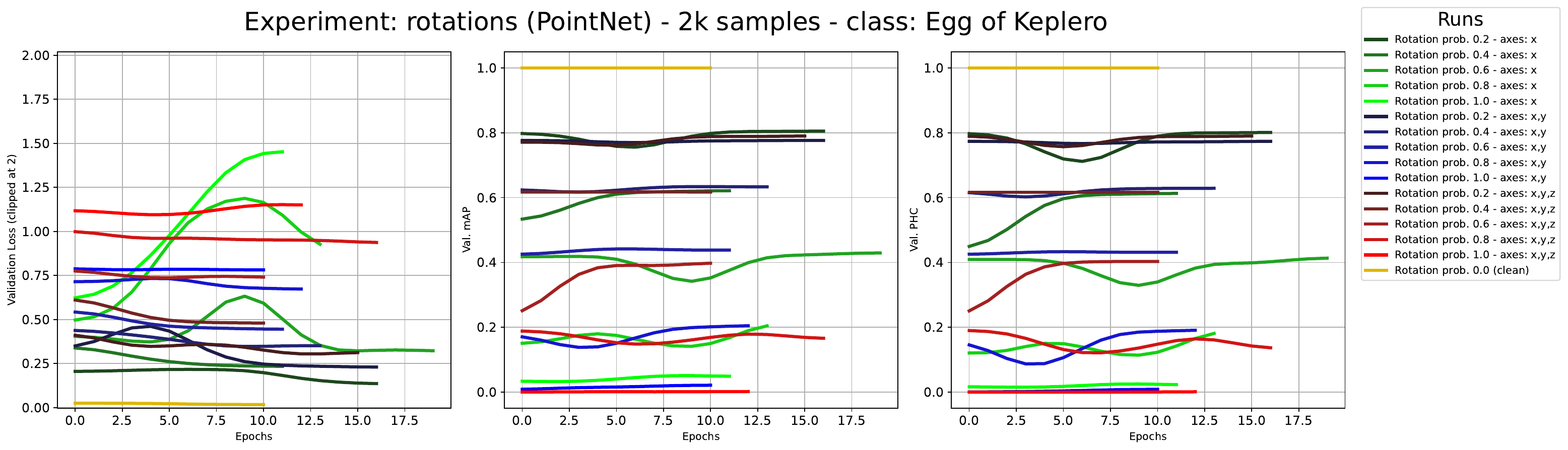}   \\
\includegraphicsrotations[0.95]{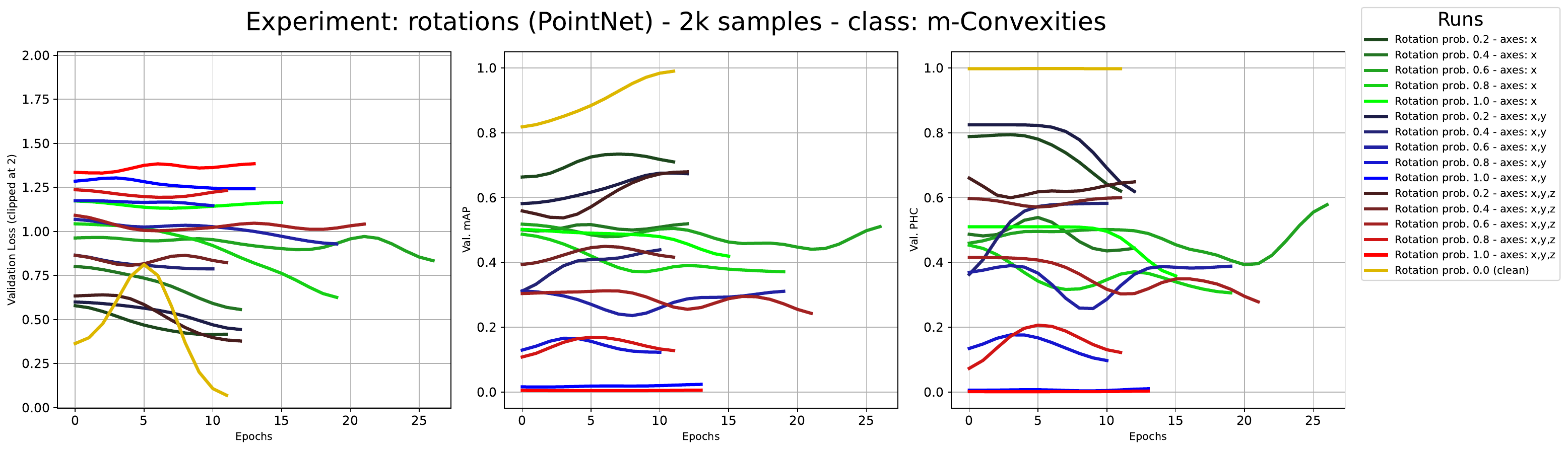} \\
\includegraphicsrotations[0.95]{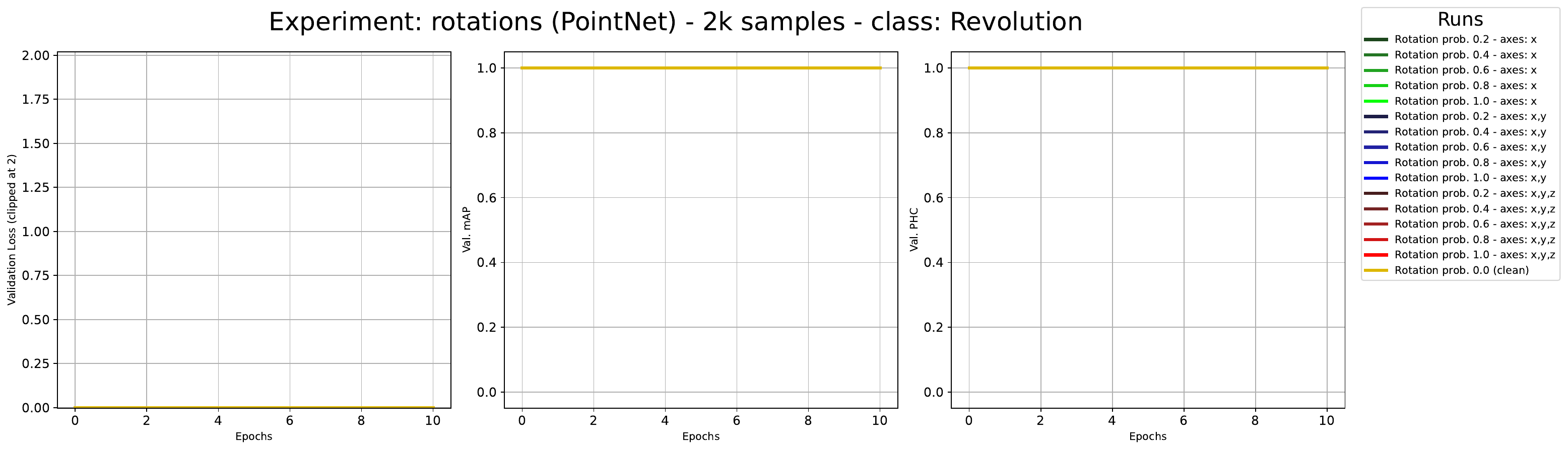}    \\
\end{tabular}
\caption{Ablation study: training on shapes rotated with varying probabilities using a standard \textit{PointNet} encoder. Classes: Astroid, Cylinder, Egg of Keplero, m-Convexities, Revolution.}
\label{tab:ablation_rotations_grouped}
\end{figure*}

The second experiment analyzes rotations about x, y, and z axes. Following the same protocol, transformation probabilities range from 0.0 (clean dataset, gold line) to 1.0. Green, blue, and red lines denote x-axis, x/y-axis, and x/y/z-axis rotations, respectively. Color intensity again encodes transformation probability.

Fig.~\ref{tab:ablation_rotations_grouped} demonstrates that rotations critically affect model performance. For the \textit{Astroid} group, the clean dataset is learned perfectly (gold line stops early due to maximum mAP triggering early stopping). x-axis rotations achieve near-perfect accuracy (mAP/PHC $\geq$ 0.9, decreasing slightly with higher probabilities). However, x/y and x/y/z rotations cause mAP/PHC to inversely correlate with rotation probability. Notably, y-axis rotations alone significantly degrade performance for \textit{Astroid}, while additional z-axis rotations provide marginal additional impact.

\textit{Cylinder} exhibits distinct behavior: the clean dataset reaches mAP/PHC $\approx$ 0.9 (matching noise experiment results in Fig.~\ref{tab:ablation_noise_undersampling_grouped}), and any rotation (even x-axis) degrades performance. \textit{Egg of Keplero} shows similar trends but with a higher baseline (clean mAP/PHC = 1.0), likely due to its variable symmetry count (1 or 3). \textit{m-Convexities} displays intermediate behavior - clean datasets achieve near-perfect scores (mAP $\approx$ 1.0), but x-axis rotations reduce mAP to $\approx$ 0.5. x/y and x/y/z rotations exhibit expected inverse probability correlations. The symmetry-free \textit{Revolution} class (exclusive to \textit{Symmetria Hard}) is learned perfectly, constituting a degenerate case.

\paragraph{Rotations with scaled-up PointNeXt XXL encoder}

\newcommand{\includegraphicsrotationsxxl}[2][]{{\includegraphics[width=#1\textwidth,trim={0.0cm 0cm 0.0cm 0cm},clip]{./#2}}}
\begin{figure*}[tbp]
\centering
\begin{tabular}{cc}
\includegraphicsrotationsxxl[0.95]{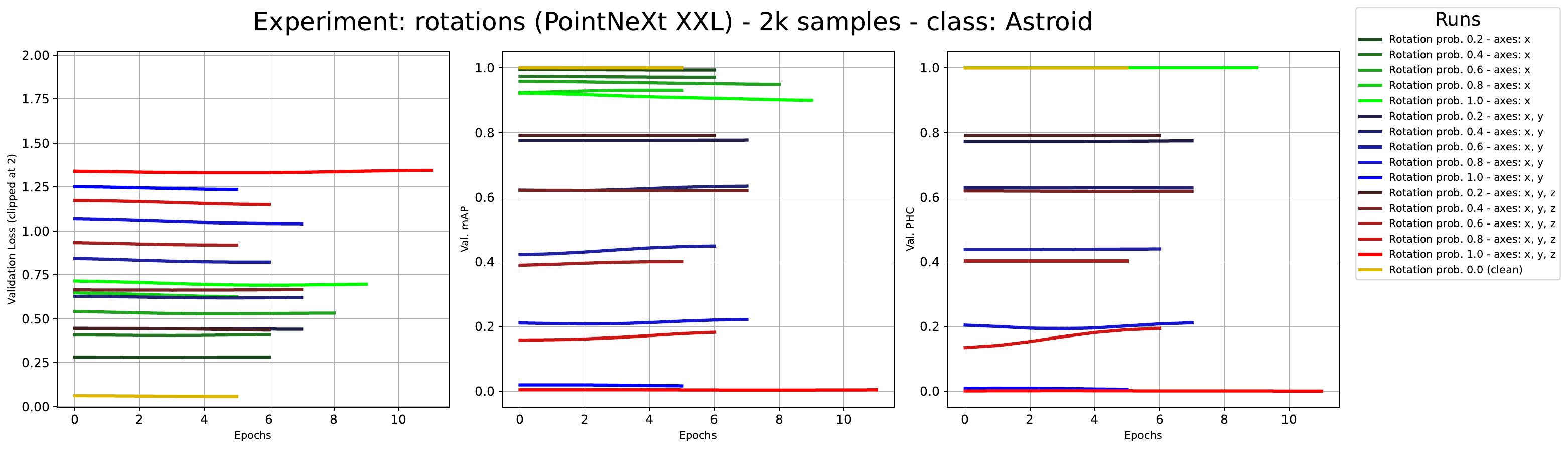}       \\
\includegraphicsrotationsxxl[0.95]{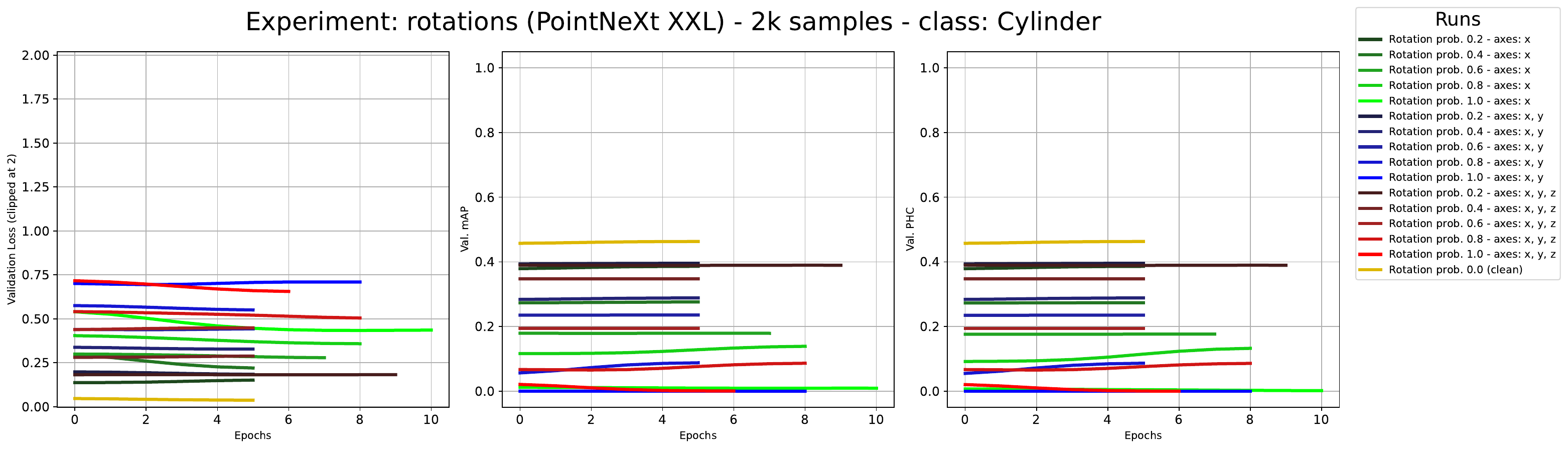}      \\
\includegraphicsrotationsxxl[0.95]{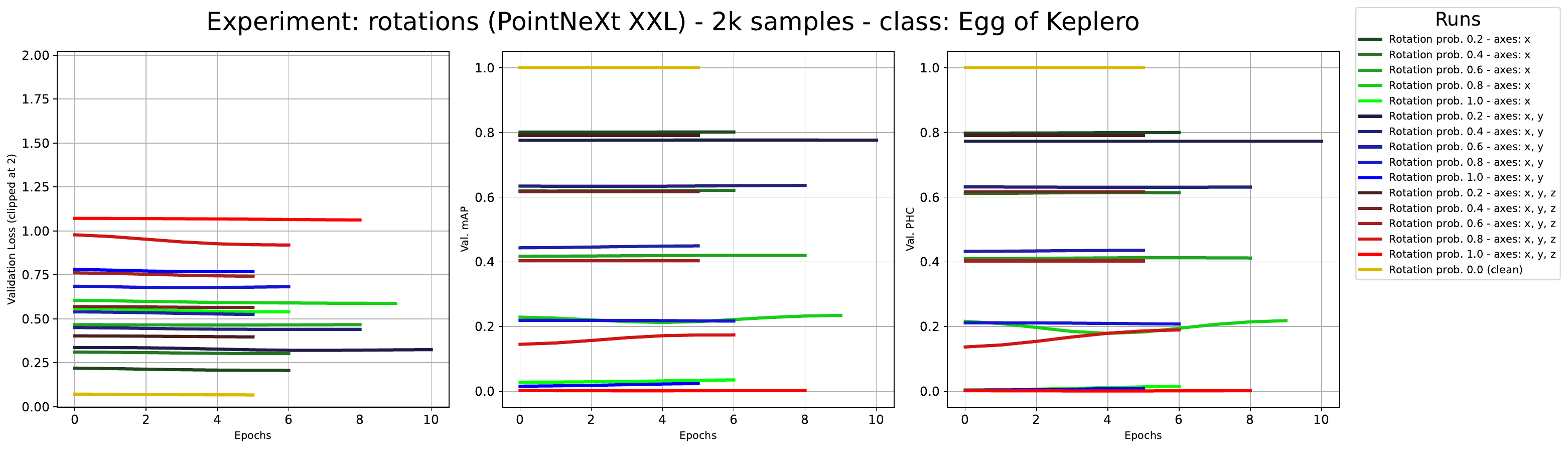}   \\
\includegraphicsrotationsxxl[0.95]{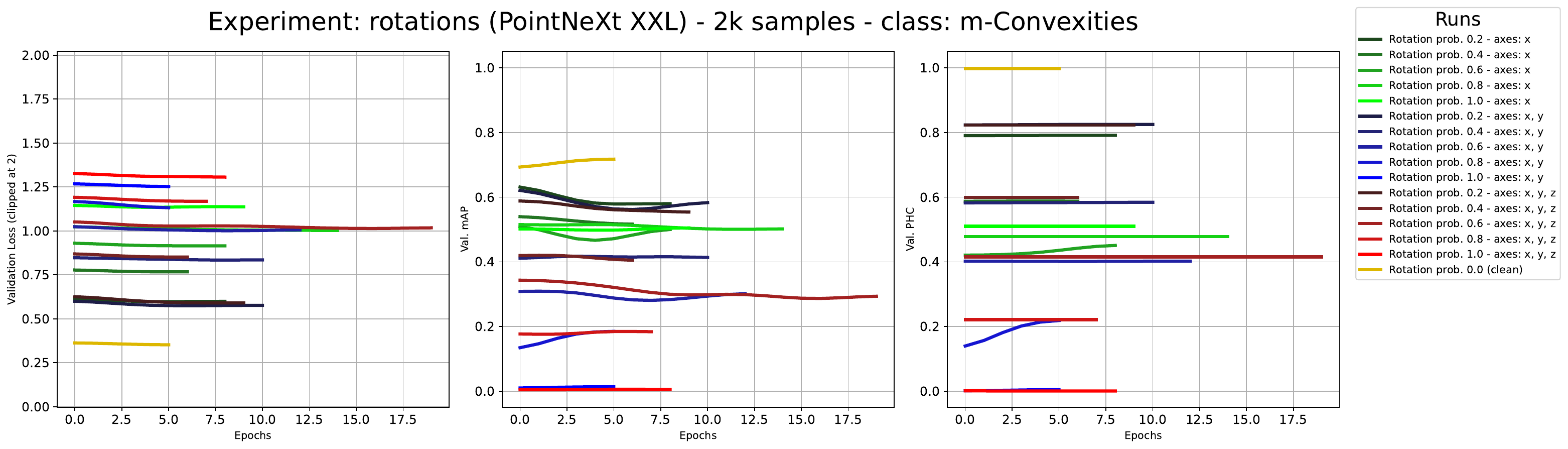} \\
\includegraphicsrotationsxxl[0.95]{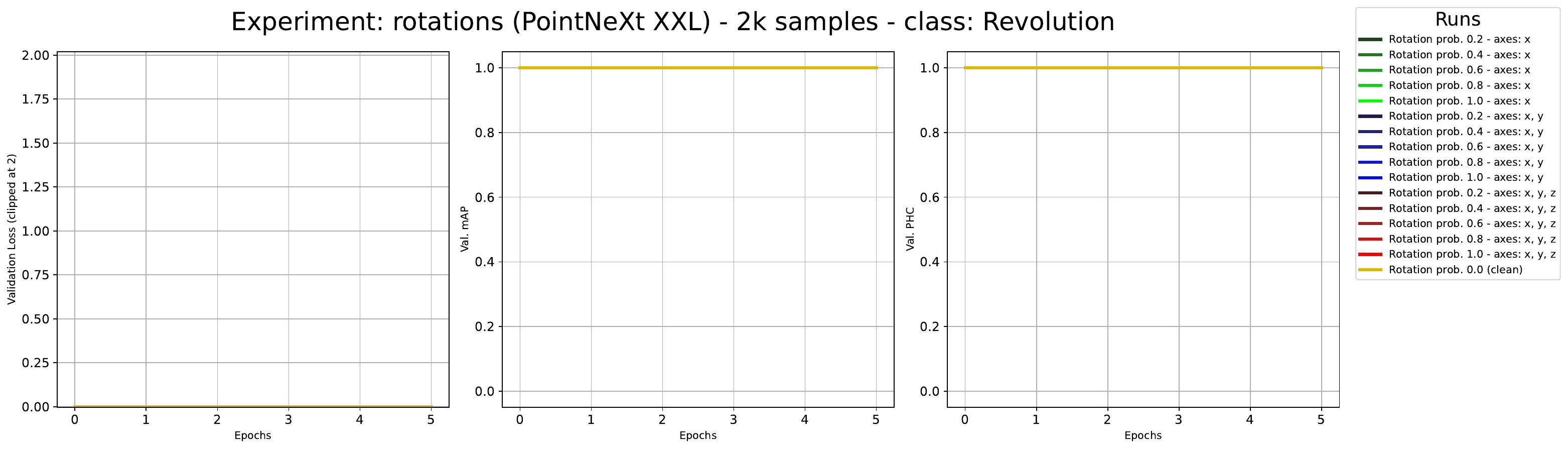}    \\
\end{tabular}
\caption{Ablation study: training on shapes rotated with varying probabilities using a scaled-up \textit{PointNeXt XXL} encoder. Classes: Astroid, Cylinder, Egg of Keplero, m-Convexities, Revolution.}
\label{tab:ablation_rotations_xxl_grouped}
\end{figure*}

Experiments using the 73-million parameter \textit{PointNeXt XXL} encoder for rotation transformations closely mirror those conducted with the standard 24-million parameter \textit{PointNet} encoder. For the \textit{Astroid} group (\textit{Astroid}, \textit{Citrus}, \textit{Geometric Petal}, \textit{Lemniscate}, \textit{Mouth curve}, \textit{Square}), results are nearly identical to \textit{PointNet} baselines but exhibit significantly smoother training curves. As shown in Fig.~\ref{tab:ablation_rotations_xxl_grouped}, clean and x-axis rotations datasets for this group achieve near-zero and very low validation loss, with mAP = 1.0 and PHC $\geq$ 0.9. However, a few peculiar differences emerge for other classes. The \textit{PointNeXt XXL} model trained on the clean \textit{Cylinder} dataset significantly underperforms its \textit{PointNet} counterpart (mAP < 0.5 vs. $~$0.9 for PointNet). A similar degradation occurs for \textit{m-Convexities}, where \textit{PointNeXt XXL} achieves mAP $\approx$ 0.7 compared to \textit{PointNet}’s near-perfect 1.0. In contrast, \textit{Egg of Keplero} and \textit{Revolution} maintain performance parity with \textit{PointNet} but exhibit smoother validation loss, mAP, and PHC trajectories.

\section{Conclusions}\label{sec:conclusions}
In this study, we present Symmetria, a novel synthetic dataset designed to advance the field of 3D point cloud learning. Our research reveals two significant findings. 

Firstly, Symmetria serves as an effective pre-training resource for self-supervised models, achieving performance levels nearly equivalent to those derived from ShapeNet pre-training across various downstream tasks. This is not completely unexpected because it confirms what is already observed for images, for instance, in \cite{Kataoka2022_Fractal} and motivates further studies in the use of synthetically generated datasets of 3D models.
In particular, Symmetria accomplishes this without compromising privacy or copyright issues, offering a distinct advantage over existing datasets. Nevertheless, to advance beyond pure scaling, we envision future work to systematically explore architectural variations, hyperparameter configurations and dataset design trade-offs through AutoML techniques and further ablation studies.

Secondly, our dataset underscores the complexity of discerning structural features within 3D objects, positioning symmetry detection as a crucial benchmark for evaluating the capabilities of neural network models. It is pertinent to mention that traditional downstream tasks, such as ModelNet classification, have nearly reached their performance limits. Thus, Symmetria introduces new avenues for exploration in the realm of 3D point cloud learning.

\paragraph{Limitations and Ethical Impact}
While synthetic datasets like Symmetria offer controlled environments for experimentation, they may lack certain features present in real-world objects, potentially leading to challenges in model learning, as observed in some of our PointGPT experiments. Nonetheless, fine-tuning generally yields satisfactory results. Future research could focus on enhancing the dataset's variability and scale to further investigate these limitations. We also recognize that incorporating non-symmetrical objects into future datasets is crucial for further improving both the dataset scale and variety. Training models on both symmetrical and non-symmetrical objects can help them learn to distinguish between the two, enhancing their ability to generalize and accurately detect symmetry when present or recognize its absence when not. Exposing models to a broader variety of asymmetrical shapes should grant them a better ability to handle real-world scenarios where objects may exhibit partial, broken, or approximate symmetry.

Importantly, we have identified no negative ethical implications associated with our dataset. On the contrary, Symmetria contributes positively by ensuring the protection of privacy and copyright, two critical considerations in contemporary data-driven research.

\section*{Acknowledgements}
The work of Ivan Sipiran has been supported by ANID Chile - Fondecyt Regular N° 1251263 and by National Center for Artificial Intelligence CENIA FB210017, Basal ANID, Chile. The work of Andrea Ranieri has been partially funded within the framework of the activities of the National Recovery and Resilience Plan (NRRP) M4C2 Inv. 1.4 -- CN MOST -- Sustainable Mobility Center, Spoke 7, whose financial support is gratefully acknowledged. 
C. Romanengo and S. Biasotti are members of the RAISE Innovation Ecosystem, funded by the European Union - NextGenerationEU and by the Ministry of University and Research (MUR), National Recovery and Resilience Plan (NRRP), Mission 4, Component 2, Investment 1.5, project ``RAISE - Robotics and AI for Socio-economic Empowerment'' (ECS00000035).

\section*{Data Availability Statement}
The datasets used in this paper are available at \url{ }. The source code for the generation of the dataset is available at \url{https://github.com/ivansipiran/symmetria}. The source code for the symmetry detection experiments is \url{https://github.com/QwagPerson/symmetria-symmetry-detection}. The code and data generated in this paper are licensed under CC BY-NC-SA 4.0.

\bibliography{refsfinal}

\newpage
\appendix

\section{Appendix}

\subsection{Symmetry Detection loss function decomposition}\label{sec:sym_det_loss_decomposition}
Our loss function can be decomposed into the following sub-tasks:

\begin{itemize}
    \item Normal Loss ($L^{normal}$): This loss quantifies the minimization of the angular difference between the true symmetries and their corresponding predicted matches, defined as follows:
    \begin{equation}
    L^{normal} = \frac{1}{K} \sum_{i=0}^{K} 1 - \abs{\hat{n_i} \cdot n_i}
    \end{equation}
    where $K$ is the number of real symmetries, $\hat{n_i}$ is the true normal, and $n_i$ is the normal of the matched predicted symmetry.
    \item Distance Loss ($L^{distance}$): This loss minimizes the distance between the predicted point on the plane and the actual point, given by:
    \begin{equation}
        L^{distance} = \norm{\hat{c} - c}
    \end{equation}
    where $\hat{c}$ is the predicted center and $c$ is the real one.
    
    \item Reflection Symmetry Distance Loss ($L^{RSD}$): This loss measures the discrepancies between the point cloud after applying the true symmetry transformation and the matched predicted transformation, similar to the Symmetry Distance Loss presented in PRS-Net \cite{prsnet}. It is calculated as:
    \begin{equation}
        L^{RSD} = \frac{1}{K} \sum_{i=0}^{K} SDE(P, \hat{y_i}, y_i)
    \end{equation}
    where $P$ is the associated point cloud, $\hat{y_i}$ represents the parameters of the predicted symmetry plane for the $i$-th symmetry, and $y_i$ represents the parameters of the true symmetry plane. The Reflection Symmetry Distance ($RSD$) between the predicted and true planes is defined as:
    \begin{equation}
        RSD(P, \hat{y_i}, y_i) = \frac{1}{3} \sum_{k=1}^3 \left\|  \left( \text{reflect}(P, y_i) - \text{reflect}(P, \hat{y_i}) \right)\right\|
    \end{equation}
    Here, $\text{reflect}(P, y_i)$ is the reflection of point cloud $P$ across the true symmetry plane $y_i$, and $\text{reflect}(P, \hat{y_i})$ is the reflection of point cloud $P$ across the predicted symmetry plane $\hat{y_i}$. The summation over $k$ indicates that the norm is computed for each of the three spatial components (x, y, z) individually instead of the usual summation over each point.

    \item Confidence Loss ($L^{confidence}$): This is a cross-entropy loss between a vector $O$, where $O_i=1$ if the $i$-th predicted symmetry corresponds to an actual symmetry and $O_i=0$ otherwise. This loss encourages the model to estimate the confidence of its symmetry predictions accurately. It is defined as:
    \begin{equation}
        L^{confidence} = -\sum_{i=1}^{K} O_i \log(\hat{O}_i) + (1 - O_i) \log(1 - \hat{O}_i)
    \end{equation}
    where $\hat{O}_i$ is the predicted confidence for the $i$-th symmetry.
\end{itemize}

The total loss function for the model combines these individual losses in a weighted sum, promoting both accurate symmetry prediction and reliable confidence estimation:
\begin{equation}
    L_{\text{total}} = \alpha L^{normal} + \beta L^{distance} + \gamma L^{RSD} + \delta L^{confidence}
\end{equation}

\begin{figure*}[b!]
\centering
\begin{tabular}{cc}
    \includegraphics[width=0.475\textwidth,trim={0.0cm 0cm 0.0cm 0cm},clip]{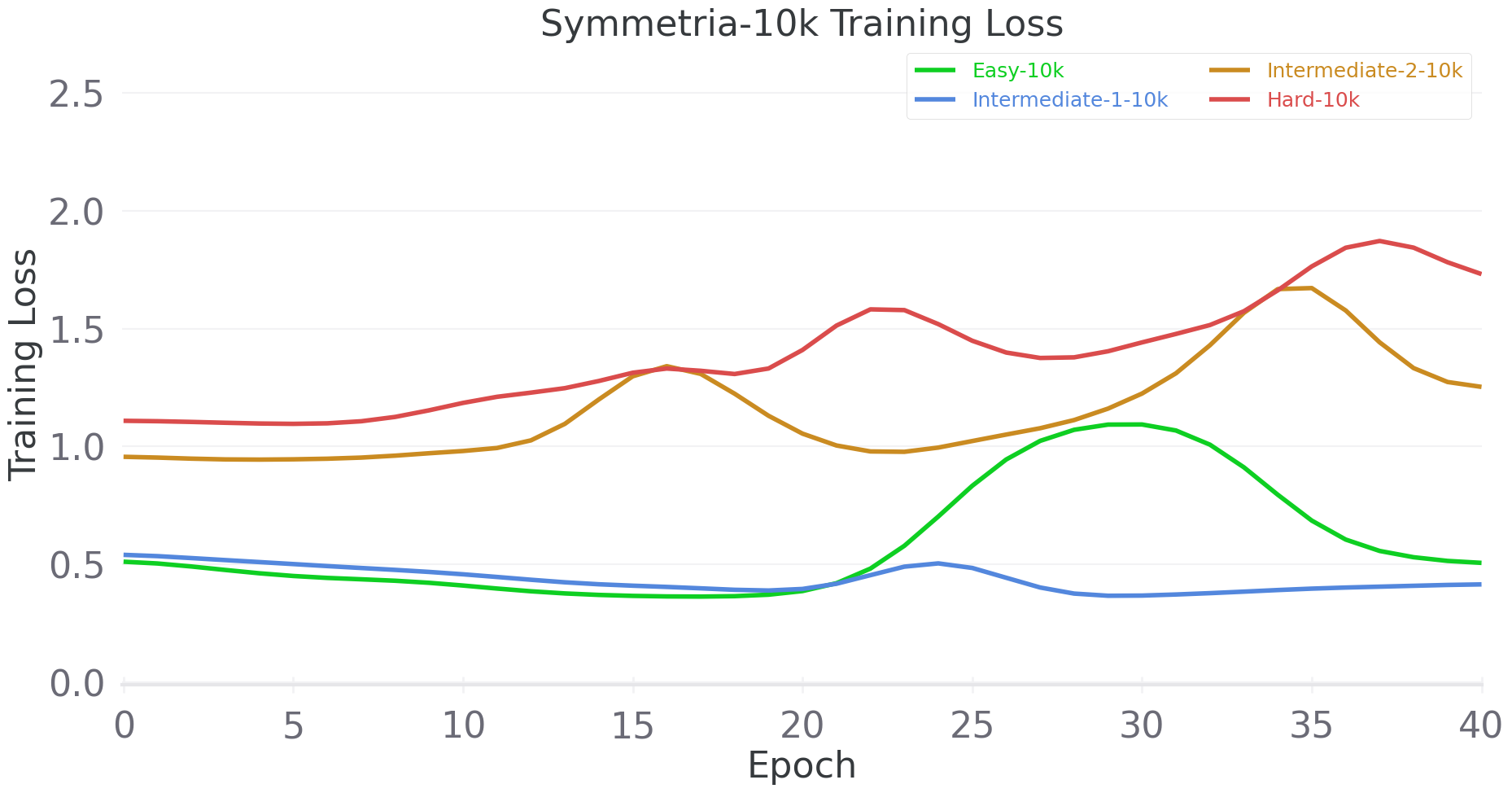} &
    \includegraphics[width=0.475\textwidth,trim={0.0cm 0cm 0.0cm 0cm},clip]{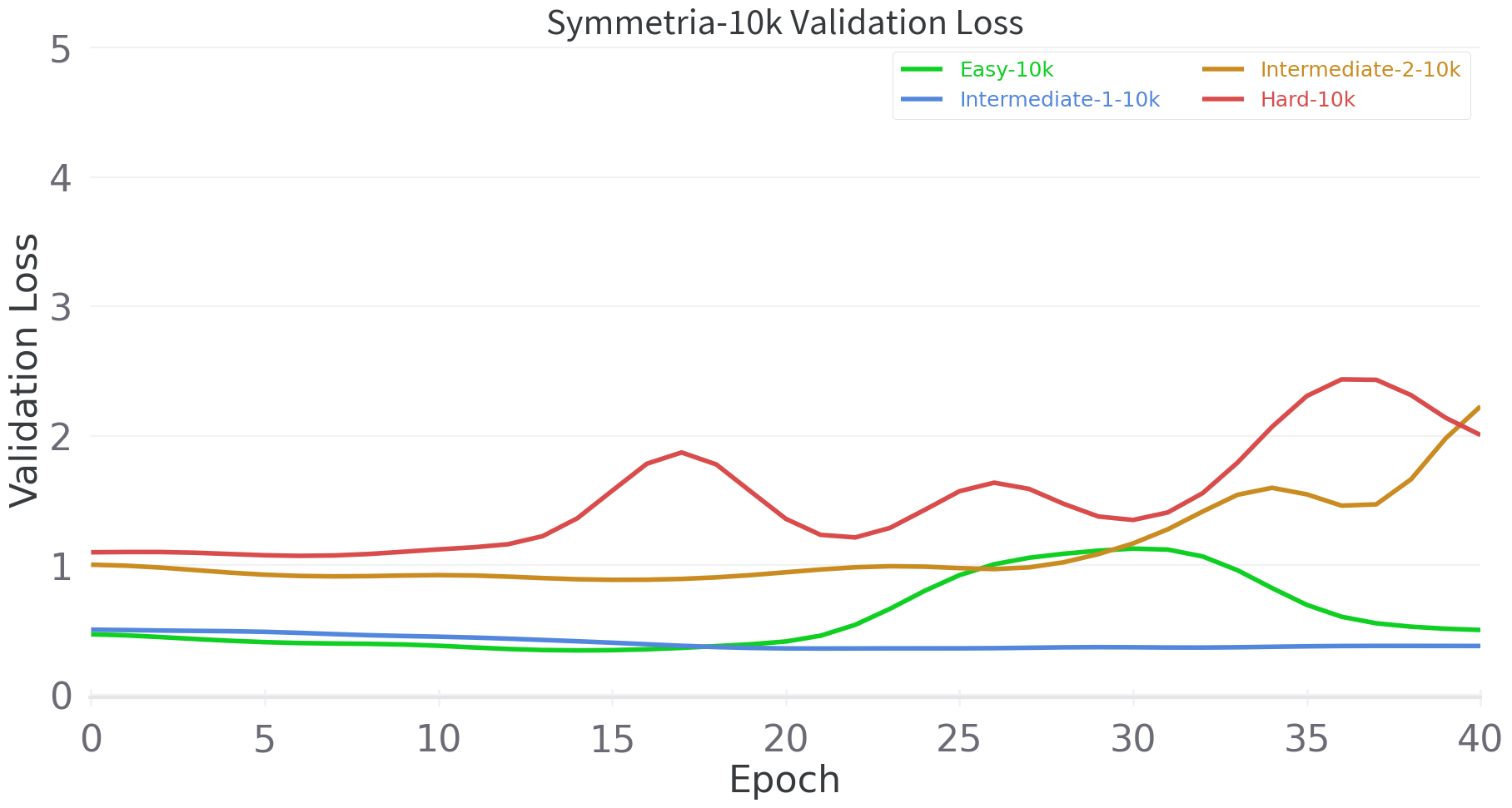} \\
    (a) & (b) \\
     & \\
    \includegraphics[width=0.475\textwidth,trim={0.0cm 0cm 0.0cm 0cm},clip]{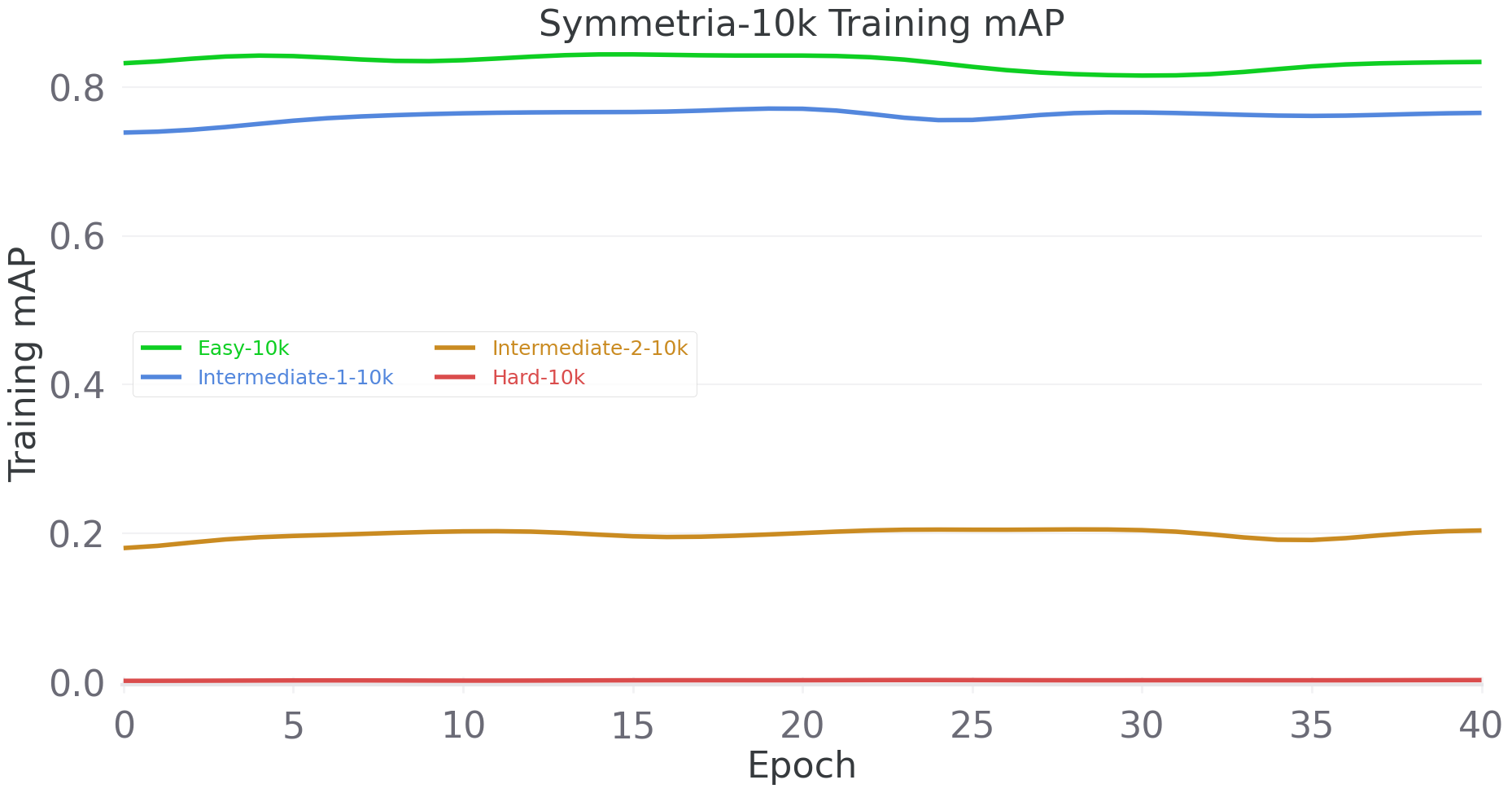} &
    \includegraphics[width=0.475\textwidth,trim={0.0cm 0cm 0.0cm 0cm},clip]{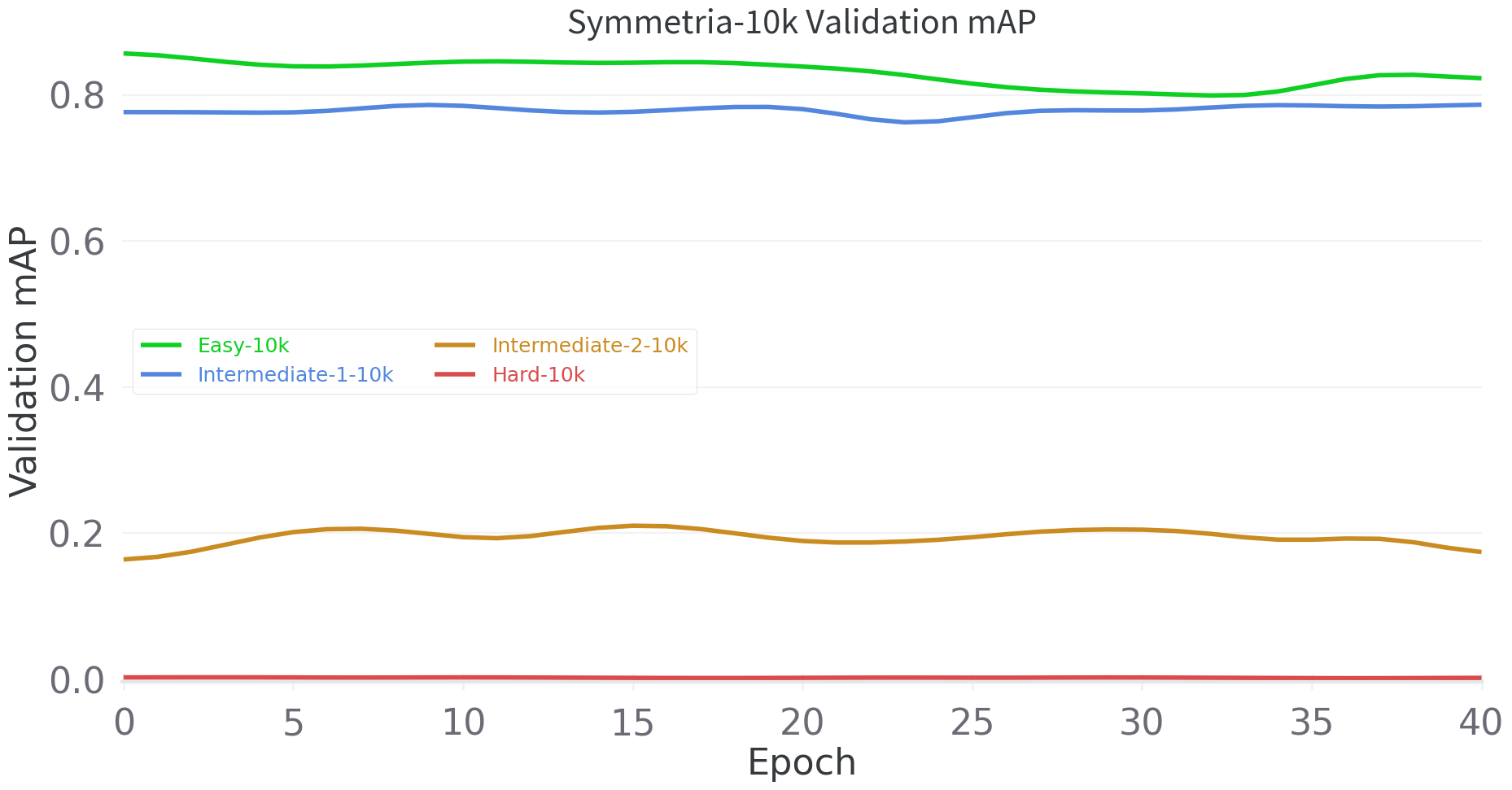} \\
    (c) & (d) \\
     & \\
    \includegraphics[width=0.475\textwidth,trim={0.0cm 0cm 0.0cm 0cm},clip]{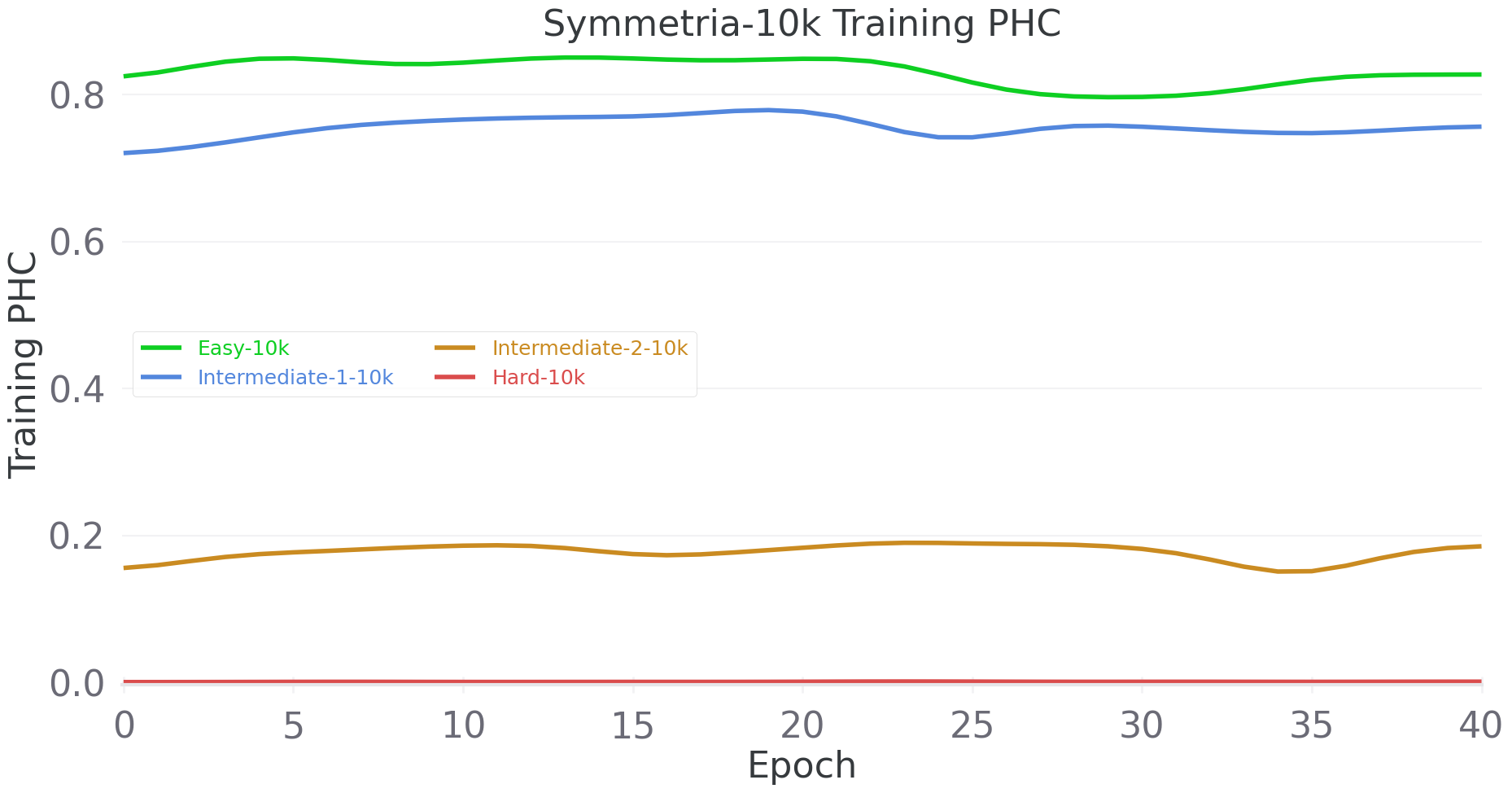} &
    \includegraphics[width=0.475\textwidth,trim={0.0cm 0cm 0.0cm 0cm},clip]{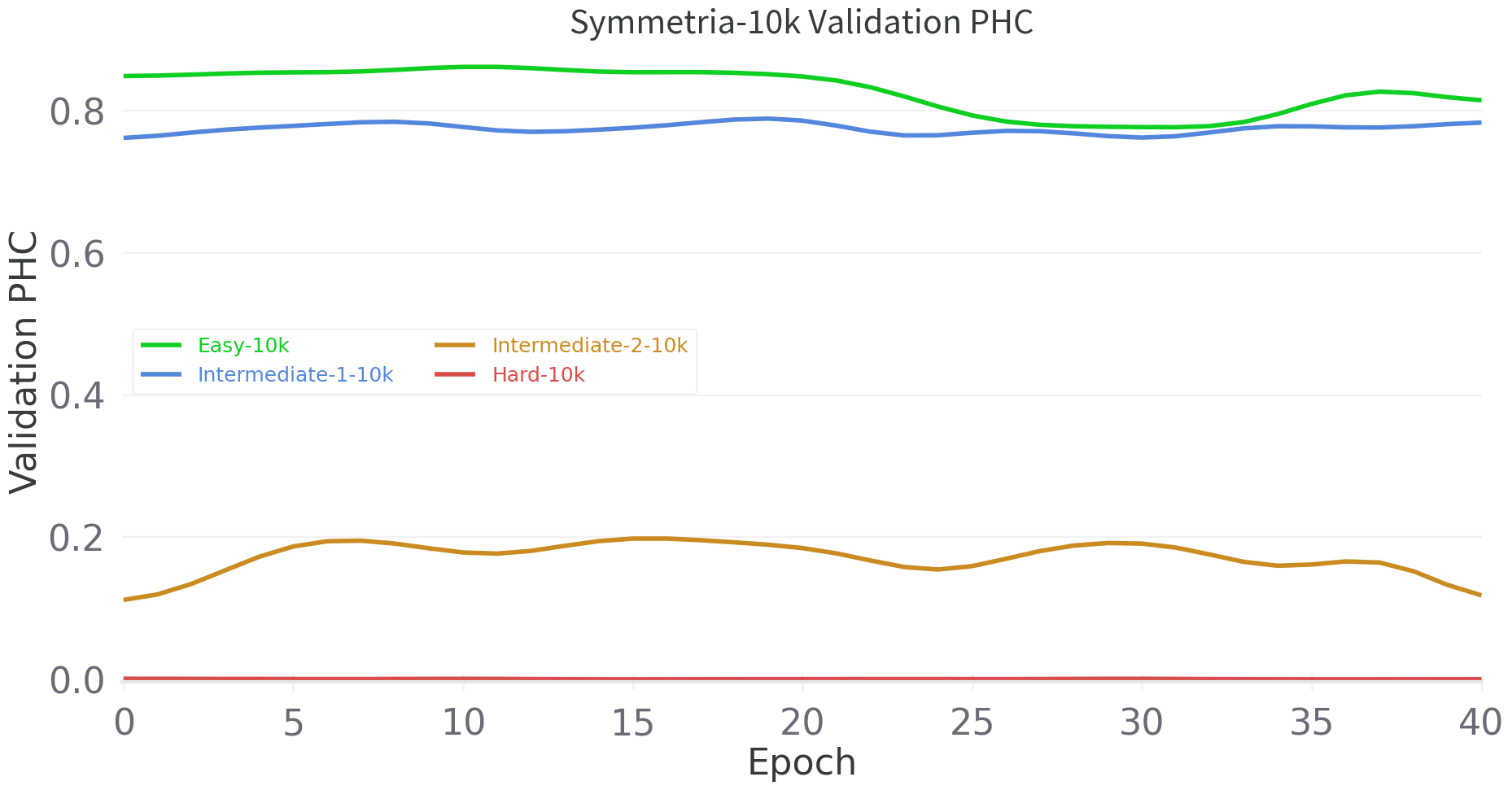} \\
    (e) & (f)  \\
\end{tabular}
\caption{Loss, mAP and PHC curves for the different Symmetria-10k training/validation runs. The $x$-axis represents the number of epochs, while the $y$-axis is the value of one of the three metrics.}
\label{tab:results_grouped_loss_map_phc_curves_10k}
\end{figure*}

where $\alpha$, $\beta$, $\gamma$, and $\delta$ are hyperparameters that balance the contributions of each loss component. For the experiments presented in Section \ref{sec:experimental_results_supervised}, we chose to set $\alpha$=1.0, $\beta$=1.0, $\delta$=1.0 and $\gamma$ = 0.1 due to this combination scaling the loss function to similar magnitudes.

\subsection{Extended analysis of the results}\label{sec:extendedanalysis}

In pursuit of exhaustive exploration of the vast hyperparameter and configuration space, we conducted a comprehensive sweep of model training, selectively presenting the most paradigmatic results from each sub-dataset comprising the Symmetria benchmark. 

We subjected the symmetry detection network, as detailed in Section \ref{sec:symmetrydetectionnetwork}, to rigorous training on the diverse Symmetria-10k sub-datasets for a minimum of 40 epochs. The converged performance is visualized in Figure \ref{tab:results_grouped_loss_map_phc_curves_10k}. Notably, on the 10k-Easy dataset, the model rapidly attains its peak validation mean average precision (mAP) of 0.868 at epoch 1, whereas on the 10k-Intermediate-1 and 10k-Intermediate-2 datasets, the maximum validation mAP is achieved at epochs 13 and 23, respectively, with values of 0.807 and 0.226. In stark contrast, the 10k-Hard sub-dataset poses a formidable challenge, with the network struggling to attain a mere 0.004 mAP at epoch 11.

\begin{figure*}[ht!]
\centering
\begin{tabular}{cc}
    \includegraphics[width=0.475\textwidth,trim={0.0cm 0cm 0.0cm 0cm},clip]{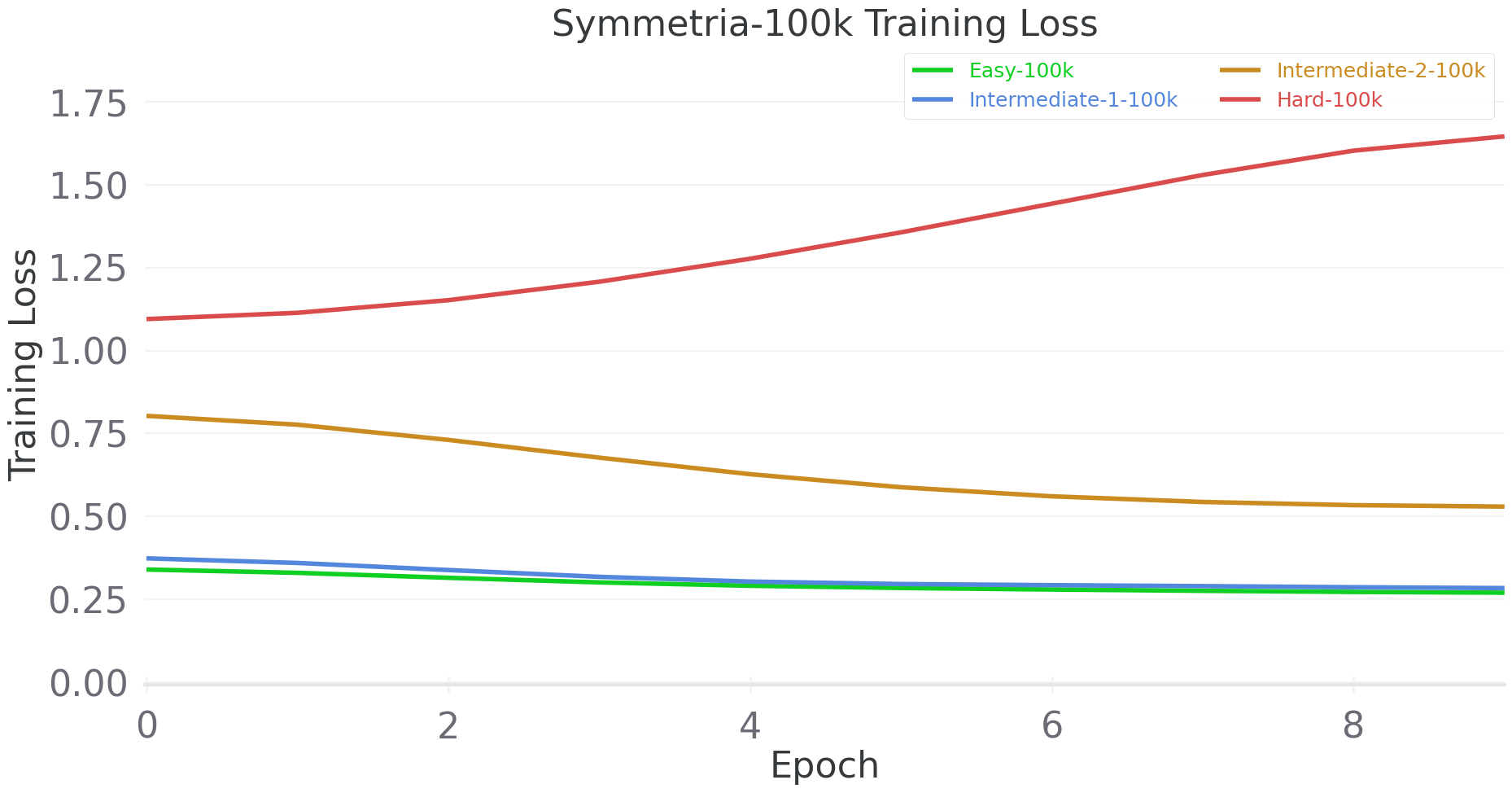} &
    \includegraphics[width=0.475\textwidth,trim={0.0cm 0cm 0.0cm 0cm},clip]{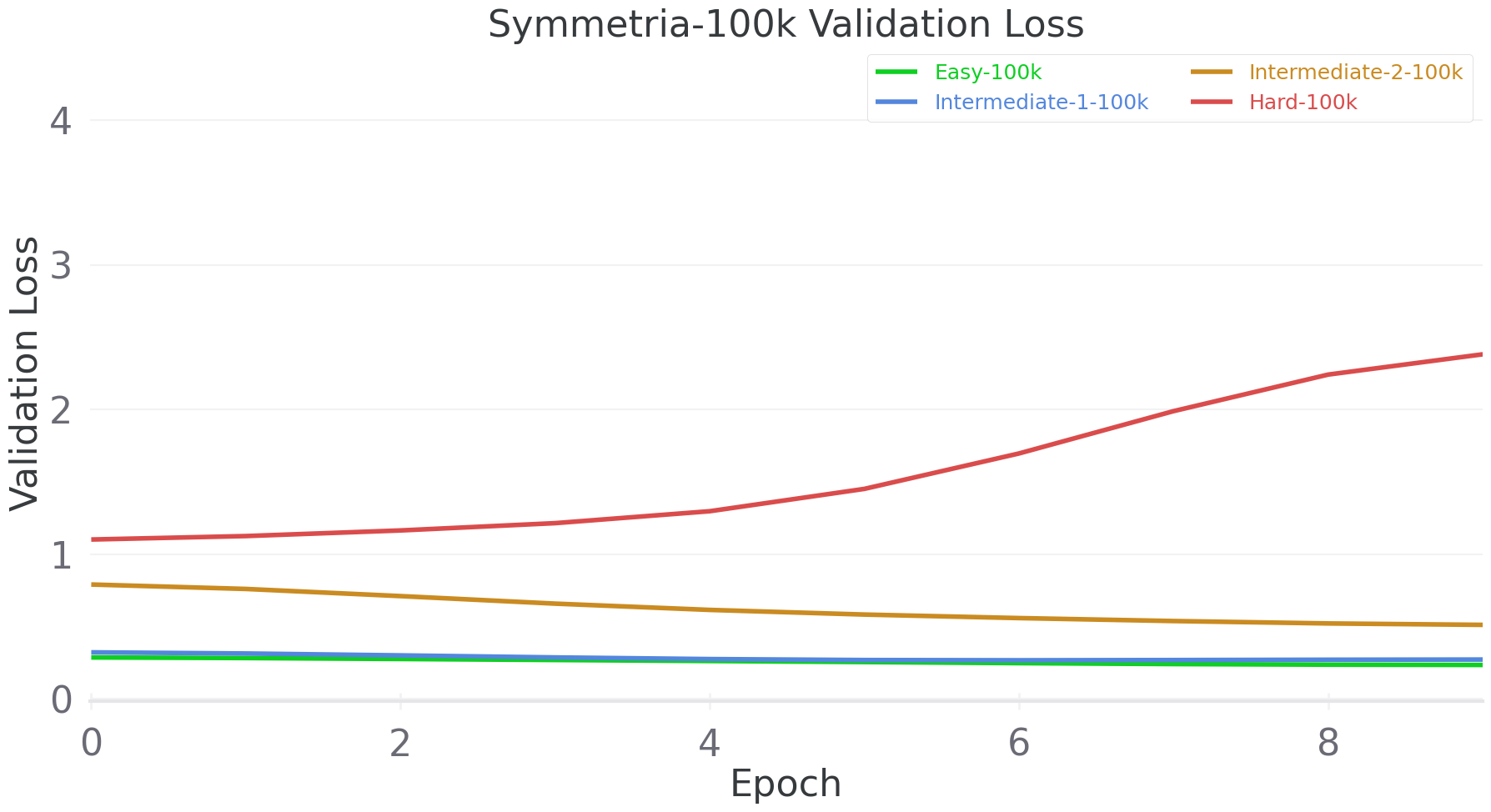} \\
    (a) & (b) \\
     & \\
    \includegraphics[width=0.475\textwidth,trim={0.0cm 0cm 0.0cm 0cm},clip]{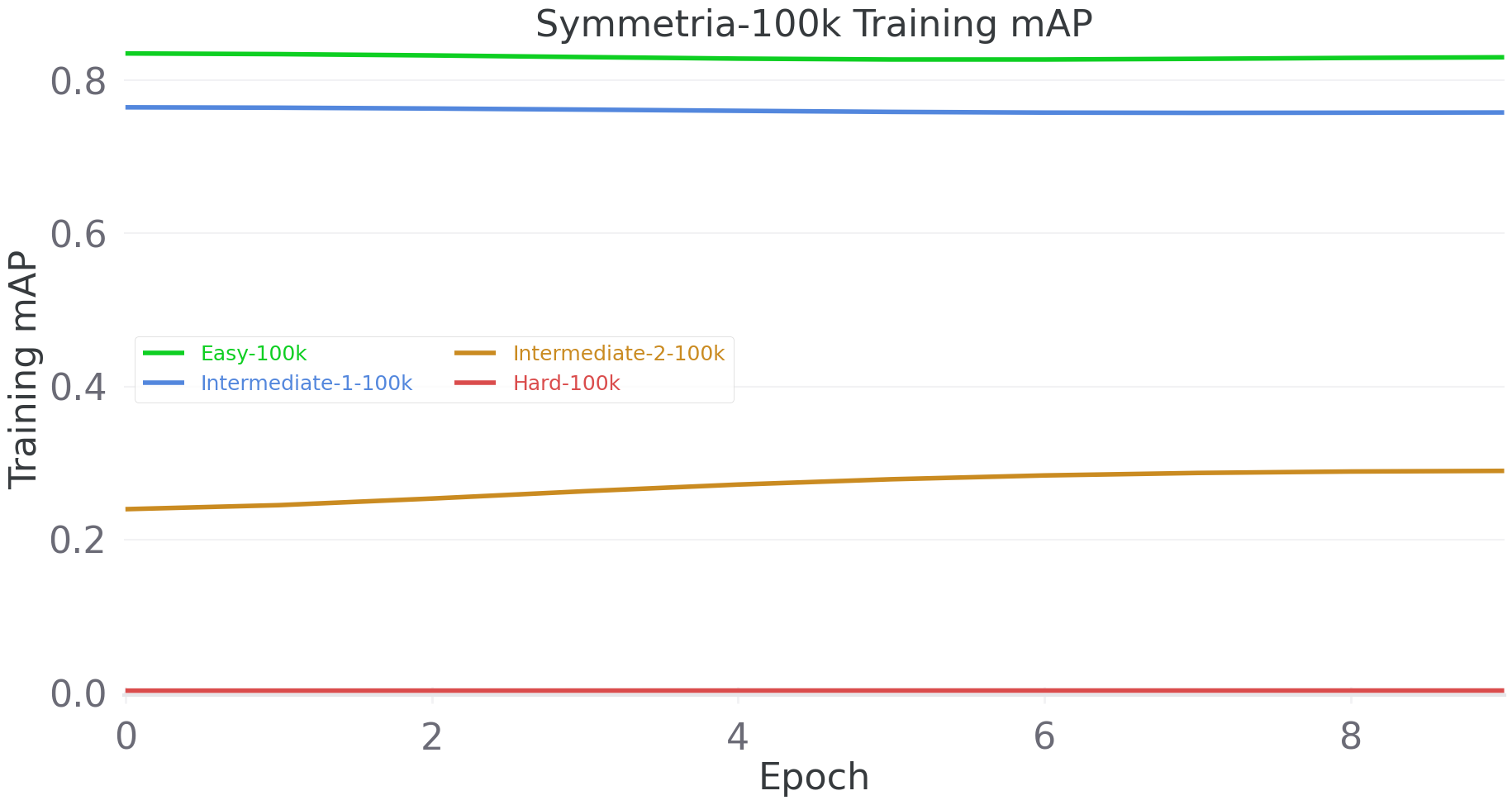} &
    \includegraphics[width=0.475\textwidth,trim={0.0cm 0cm 0.0cm 0cm},clip]{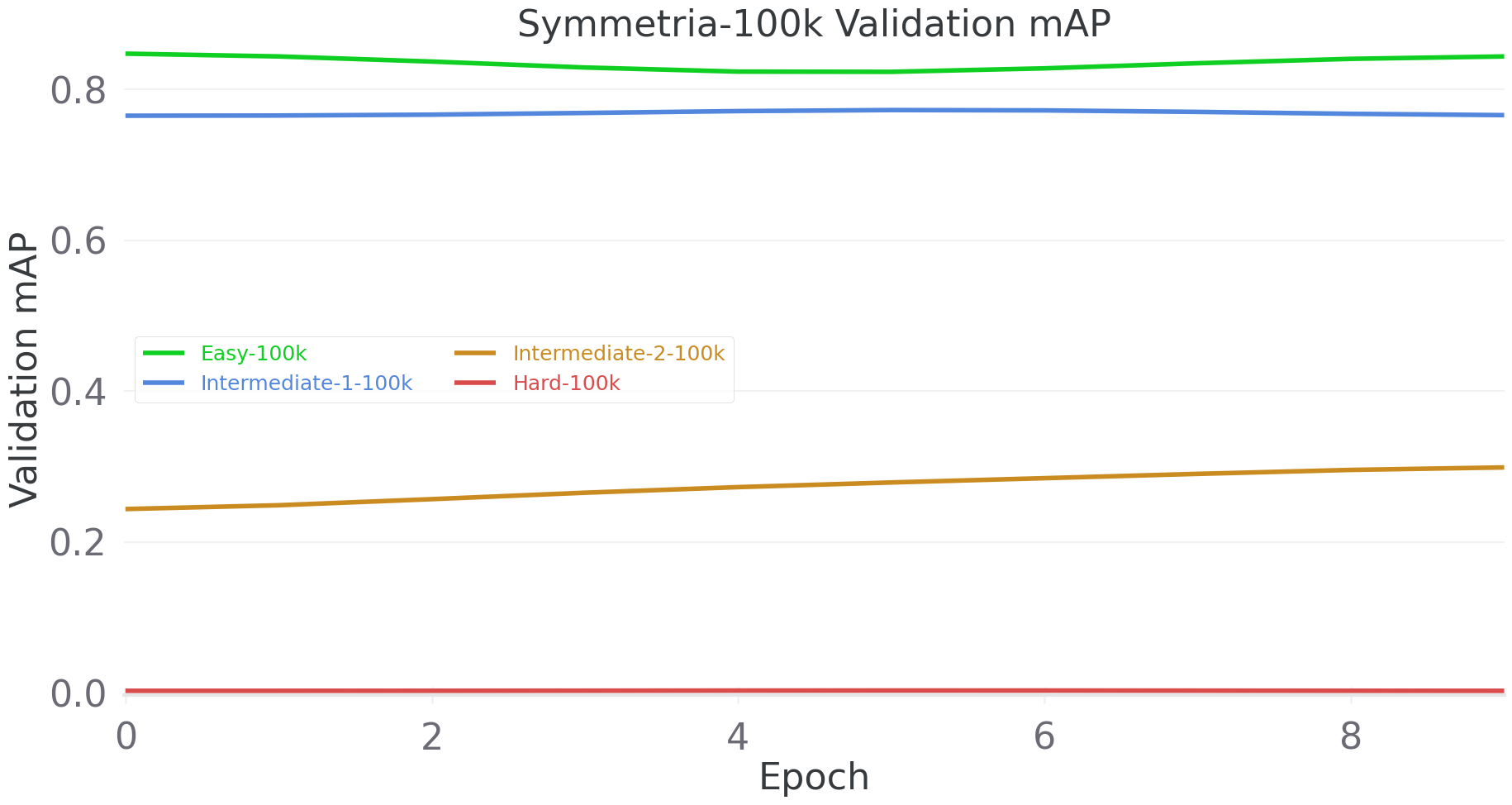} \\
    (c) & (d) \\
     & \\
    \includegraphics[width=0.475\textwidth,trim={0.0cm 0cm 0.0cm 0cm},clip]{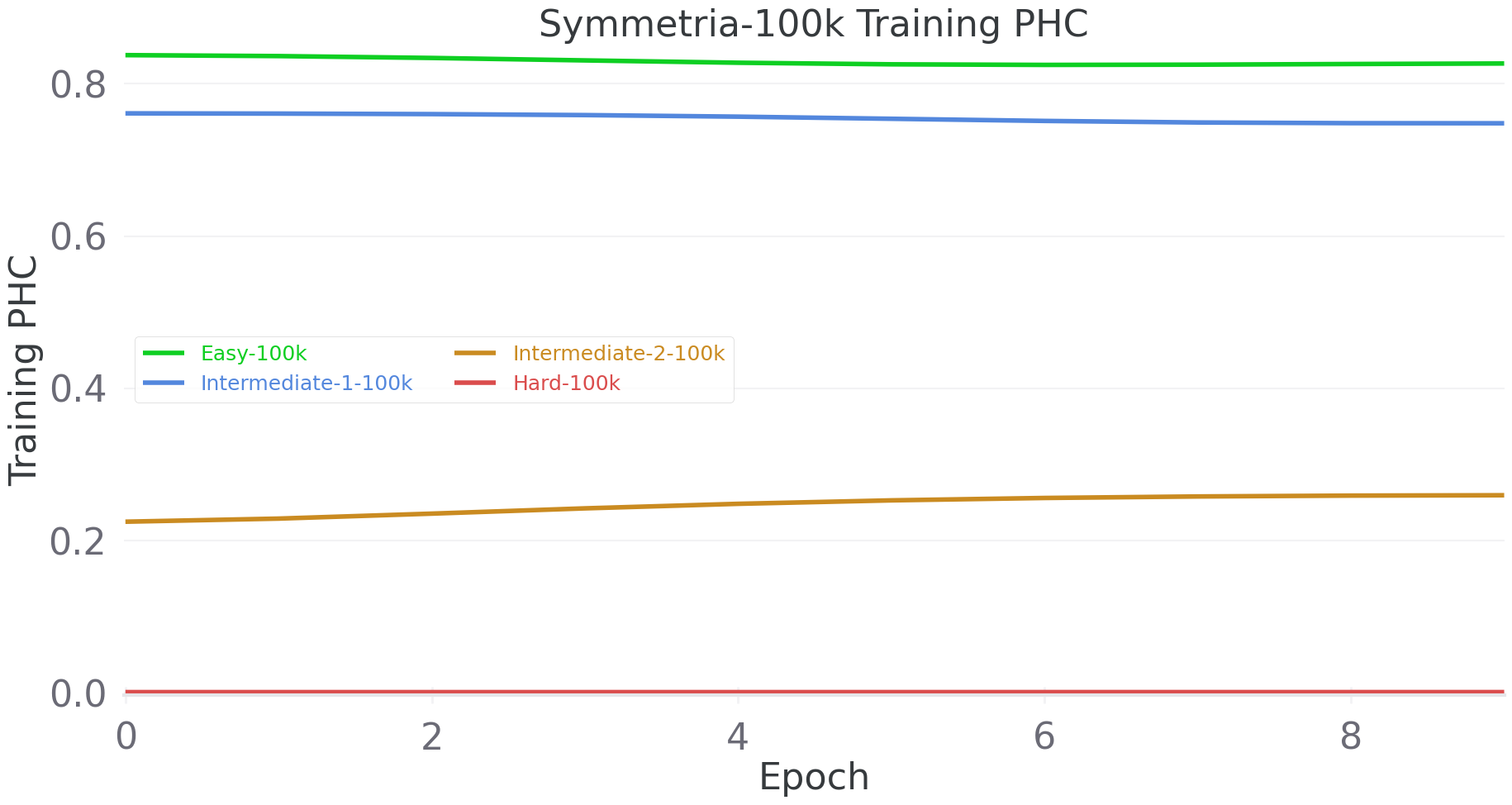} &
    \includegraphics[width=0.475\textwidth,trim={0.0cm 0cm 0.0cm 0cm},clip]{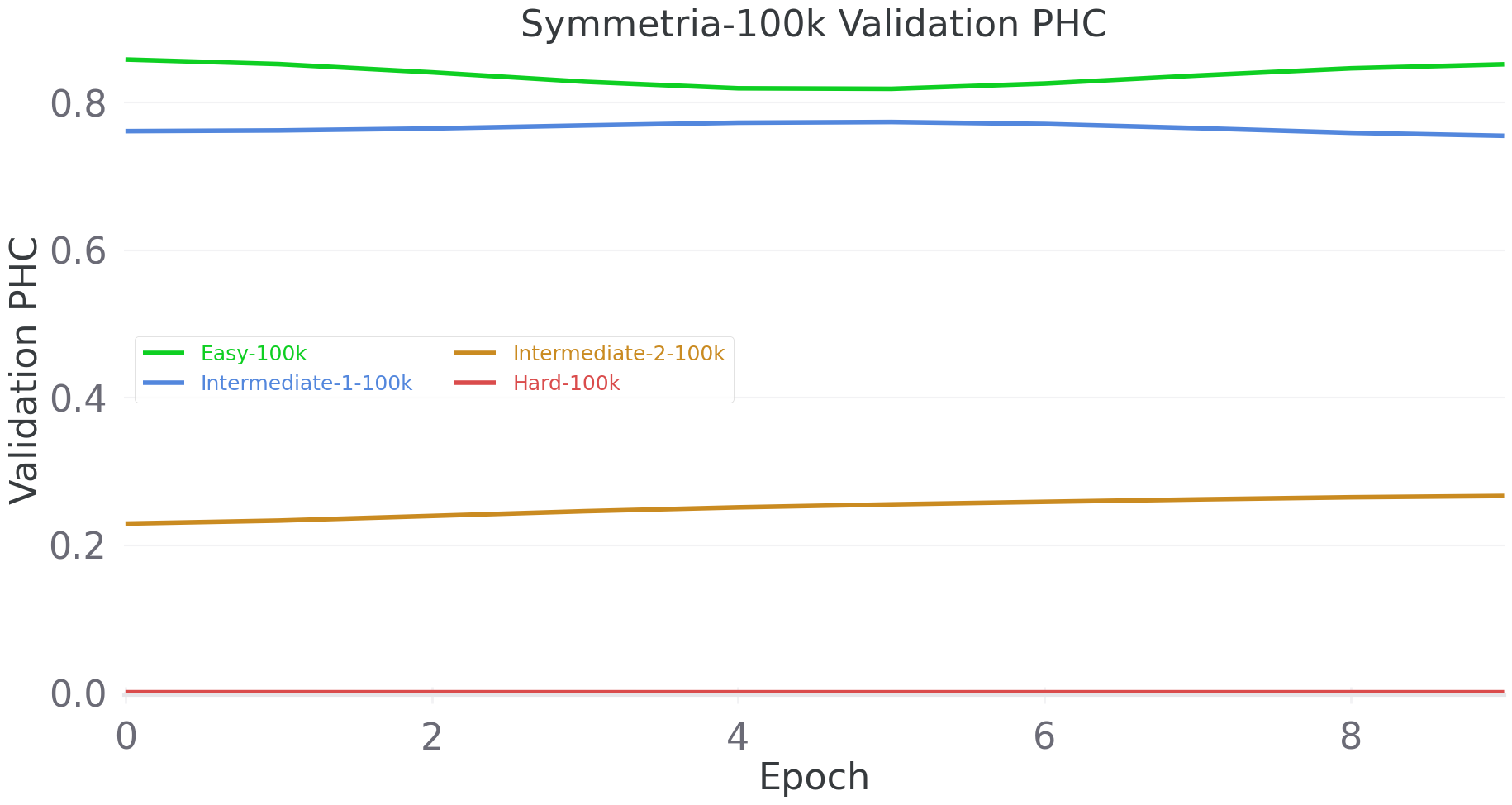} \\
    (e) & (f)  \\
\end{tabular}
\caption{Loss, mAP, and PHC curves for the different Symmetria-100k training/validation runs. The $x$-axis represents the number of epochs, while the $y$-axis is the value of one of the three metrics.}
\label{tab:results_grouped_loss_map_phc_curves_100k}
\end{figure*}

Our empirical investigation on the \textit{Symmetria-100k} sub-datasets unveils intriguing learning dynamics mirroring those observed in their downscaled counterparts. Given the extensive nature of these datasets, we trained the diverse models for a maximum of 10 epochs. The \textit{Easy-100k} model exhibits rapid convergence, achieving peak validation mAP of 0.852 within the first epoch, indicating swift assimilation of the inherent dataset features (Figure \ref{tab:results_grouped_loss_map_phc_curves_100k}). Similarly, models trained on \textit{Symmetria-100k} \textit{Intermediate-1} and \textit{Intermediate-2} attain their maximum validation mAP (0.788 and 0.320, respectively) at epochs 4 and 9, suggesting a positive correlation between task complexity and convergence time.

Conversely, the \textit{Hard-100k} model exhibits premature saturation, reaching its maximum validation mAP (0.004) at epoch 5, hinting at potential limitations in the current multitask problem formulation in effectively capturing the underlying data complexities.

\begin{figure*}[h!]
\centering
\begin{tabular}{cc}
    \includegraphics[width=0.475\textwidth,trim={0.0cm 0cm 0.0cm 0cm},clip]{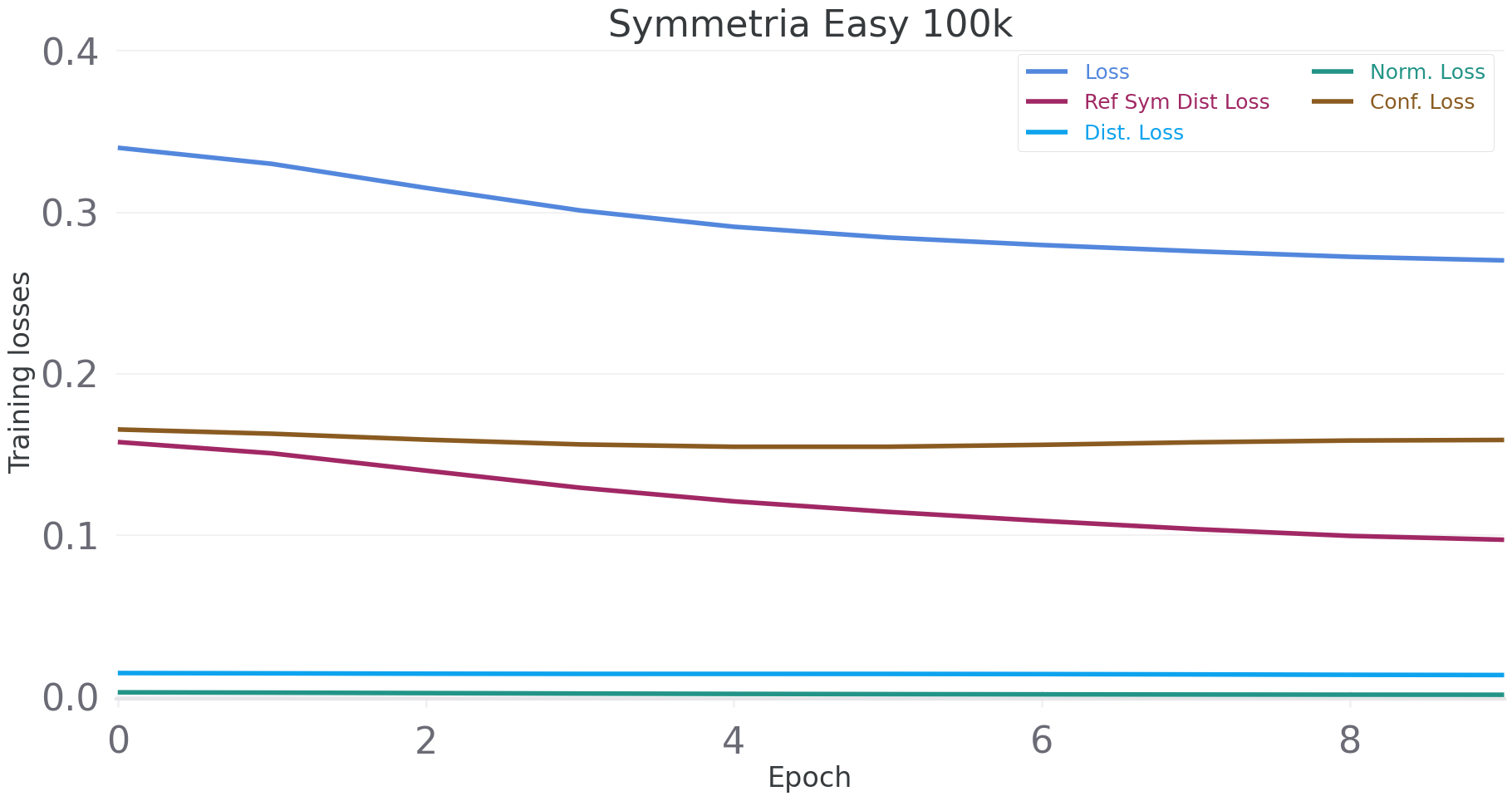} &
    \includegraphics[width=0.475\textwidth,trim={0.0cm 0cm 0.0cm 0cm},clip]{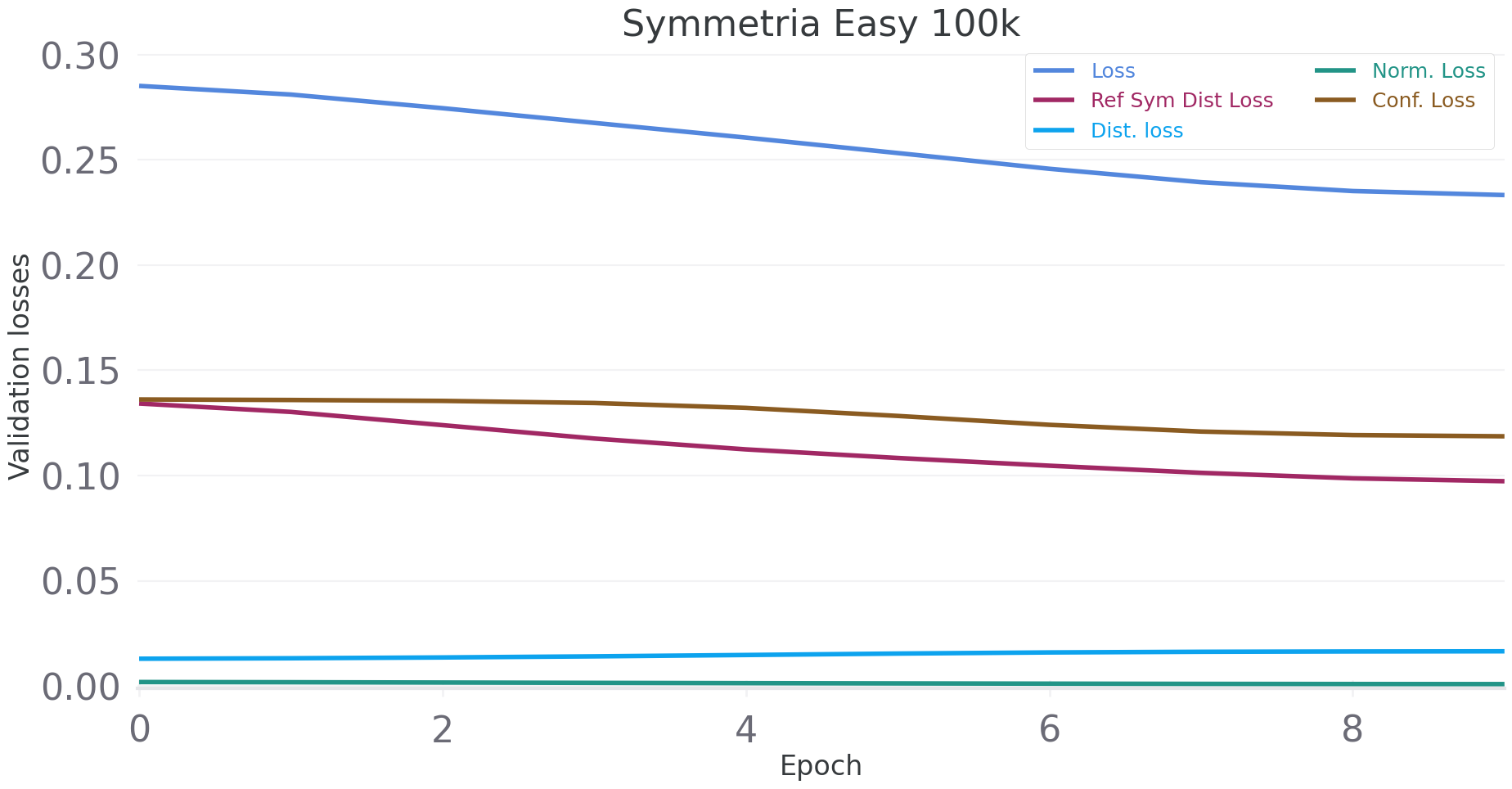} \\
    (a) & (b) \\
    \includegraphics[width=0.475\textwidth,trim={0.0cm 0cm 0.0cm 0cm},clip]{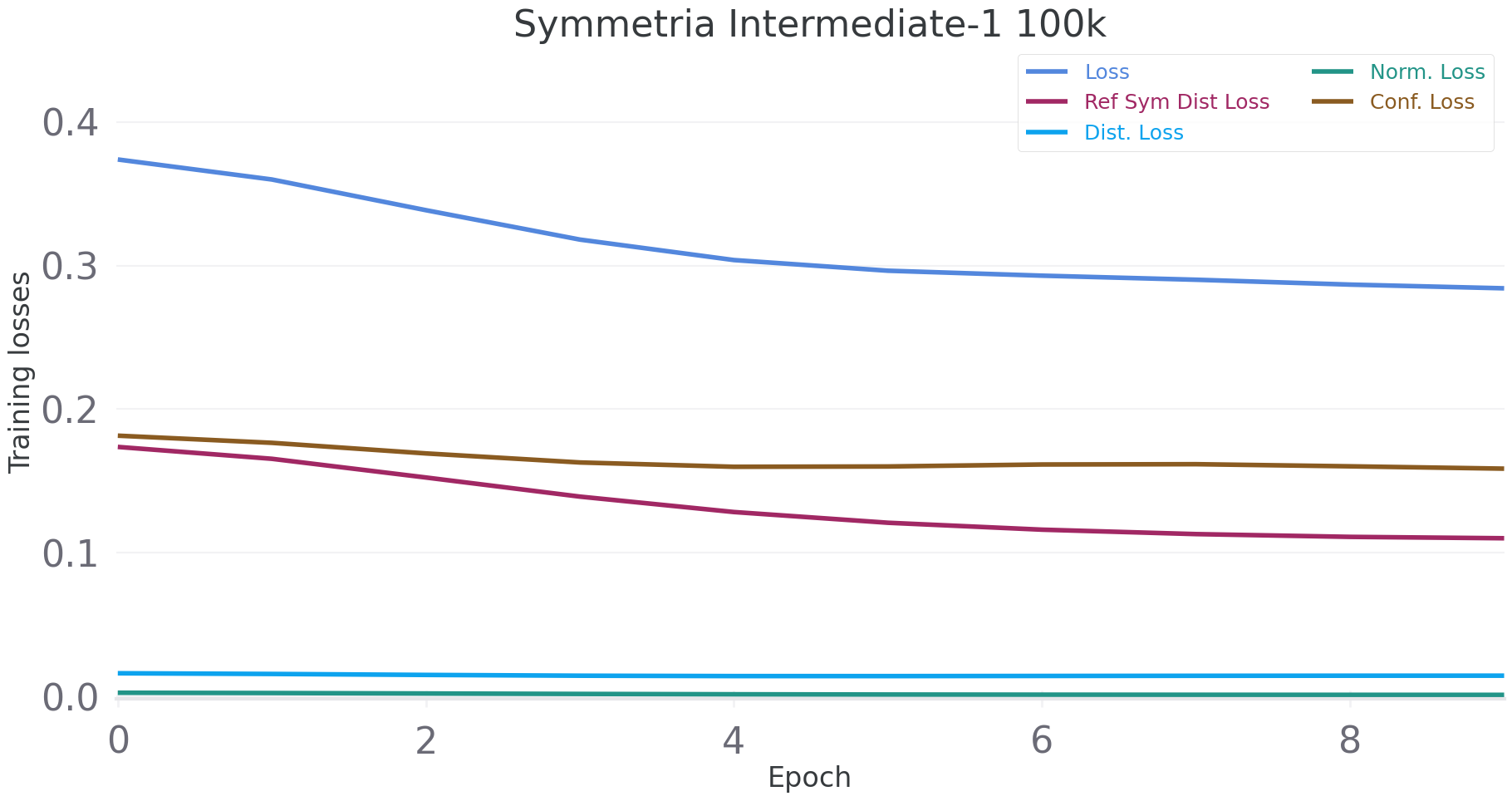} &
    \includegraphics[width=0.475\textwidth,trim={0.0cm 0cm 0.0cm 0cm},clip]{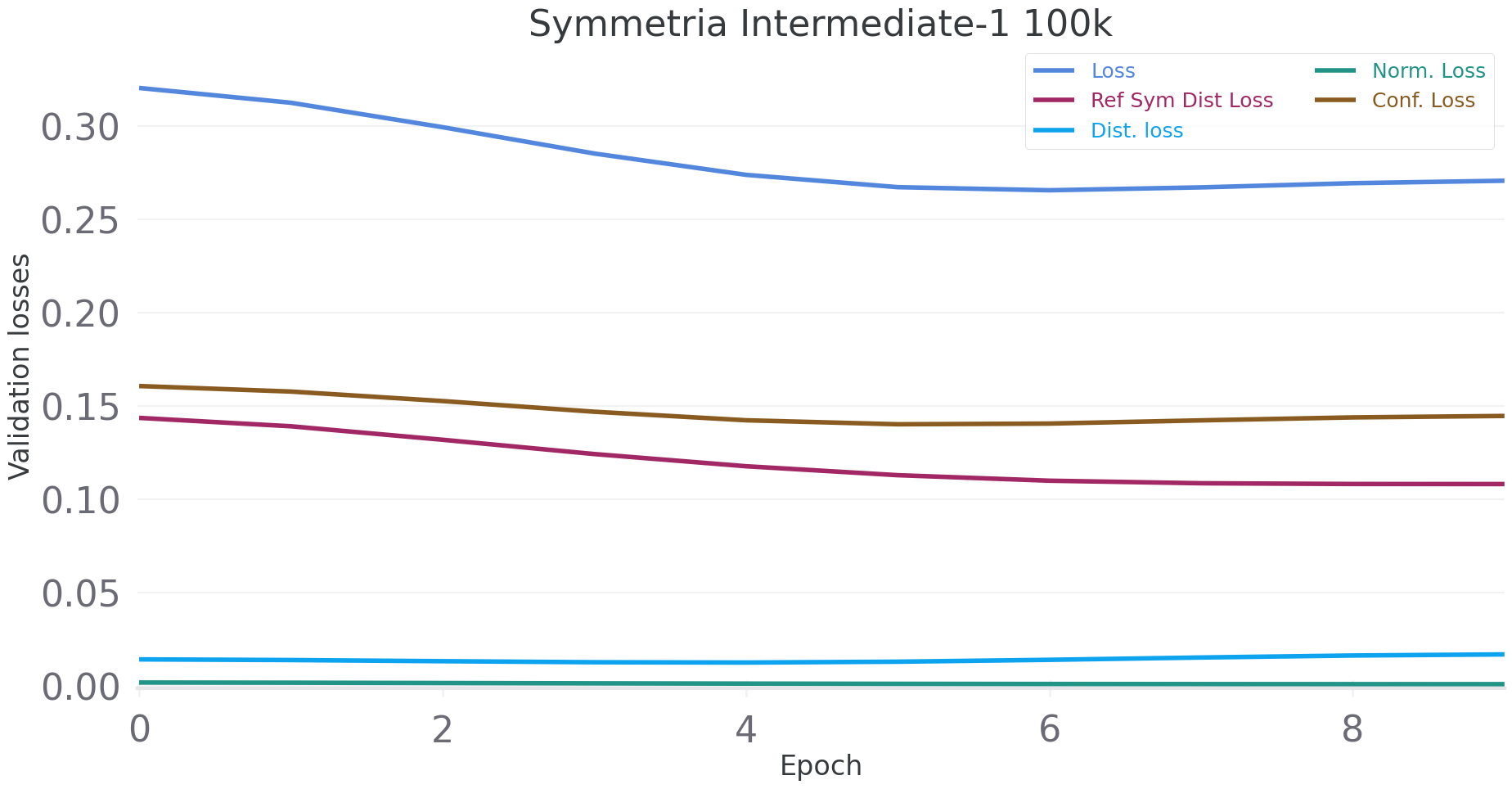} \\
    (c) & (d) \\
     & \\
    \includegraphics[width=0.475\textwidth,trim={0.0cm 0cm 0.0cm 0cm},clip]{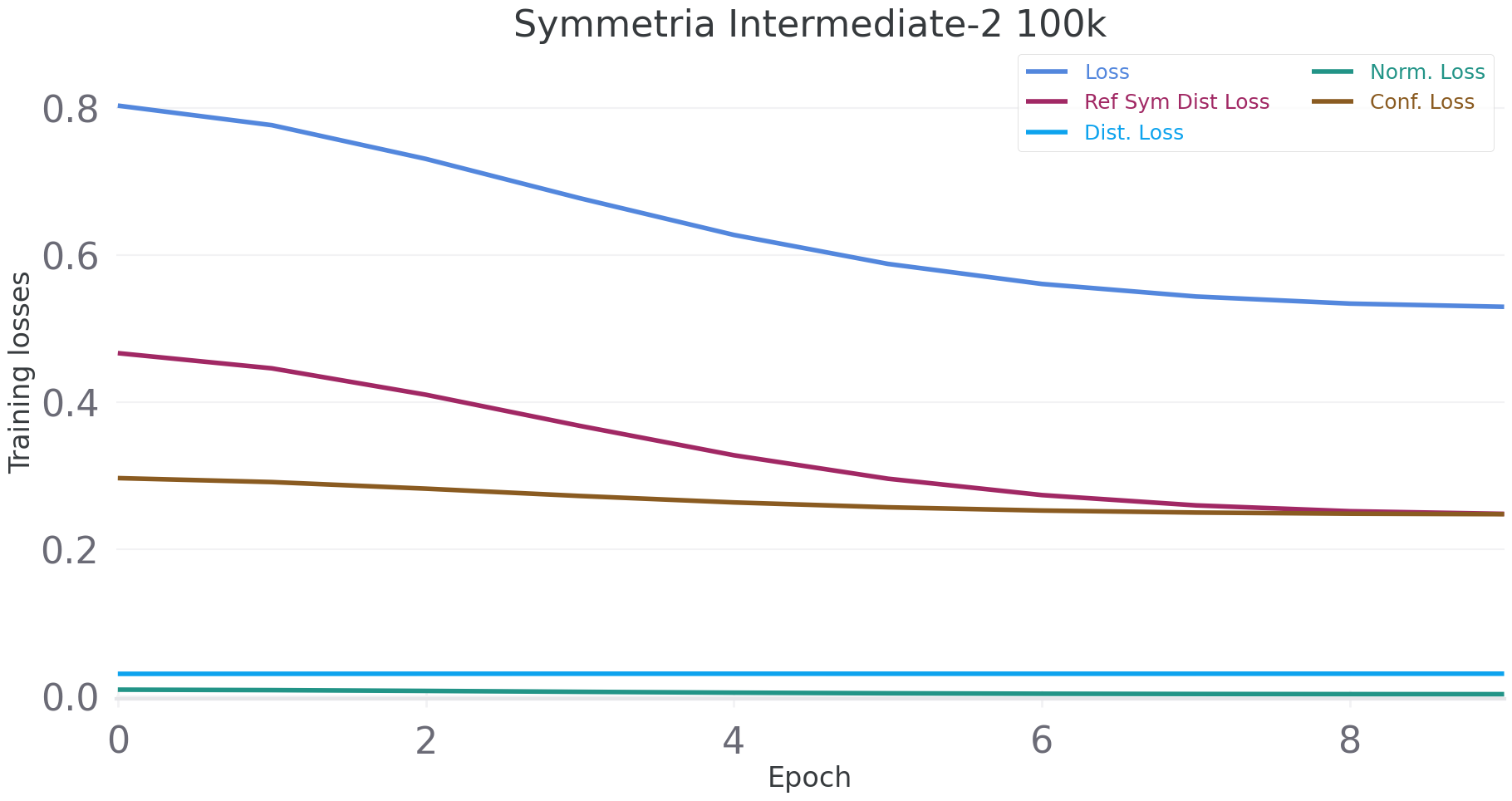} &
    \includegraphics[width=0.475\textwidth,trim={0.0cm 0cm 0.0cm 0cm},clip]{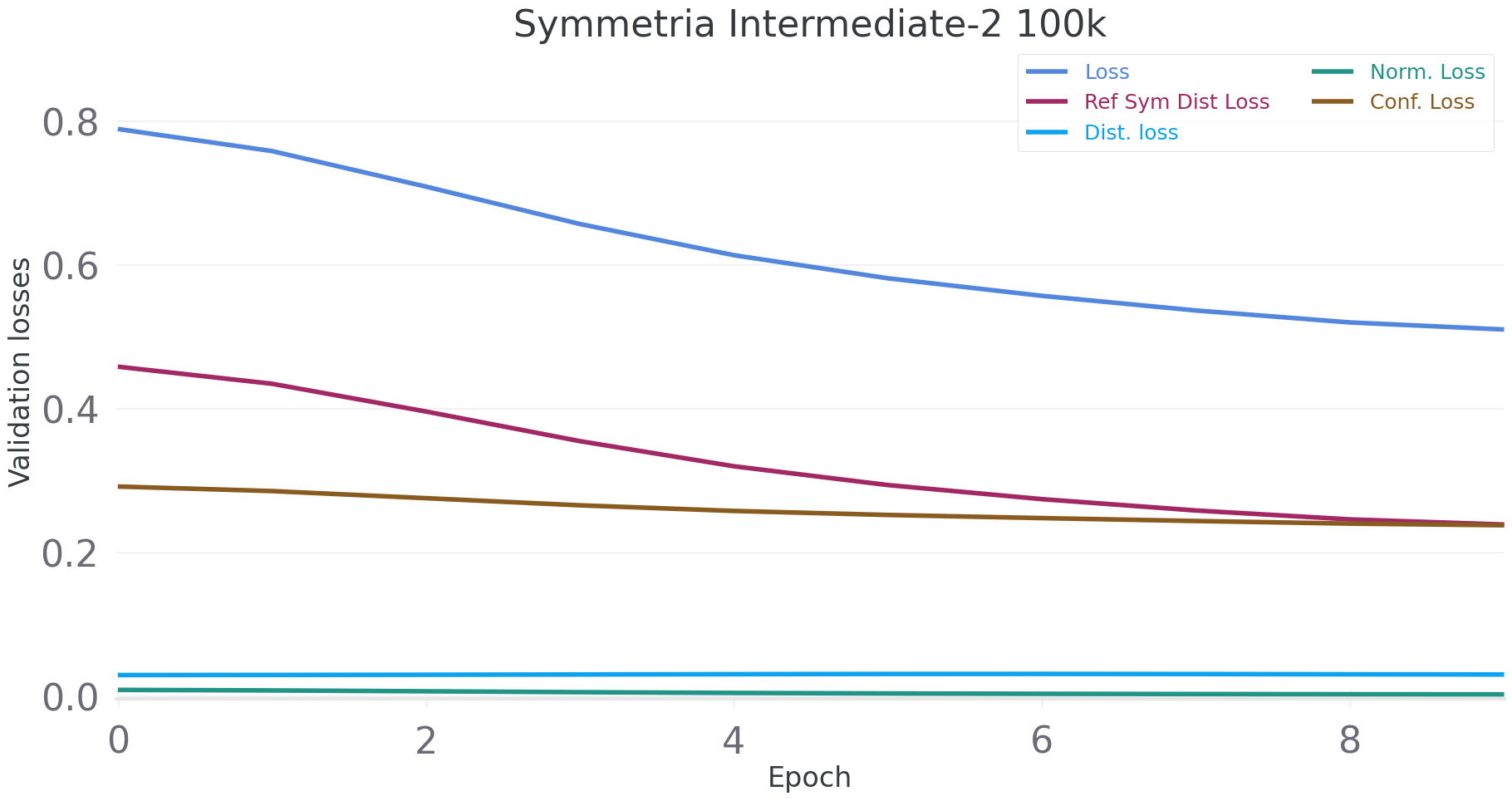} \\
    (e) & (f)  \\
     & \\
    \includegraphics[width=0.475\textwidth,trim={0.0cm 0cm 0.0cm 0cm},clip]{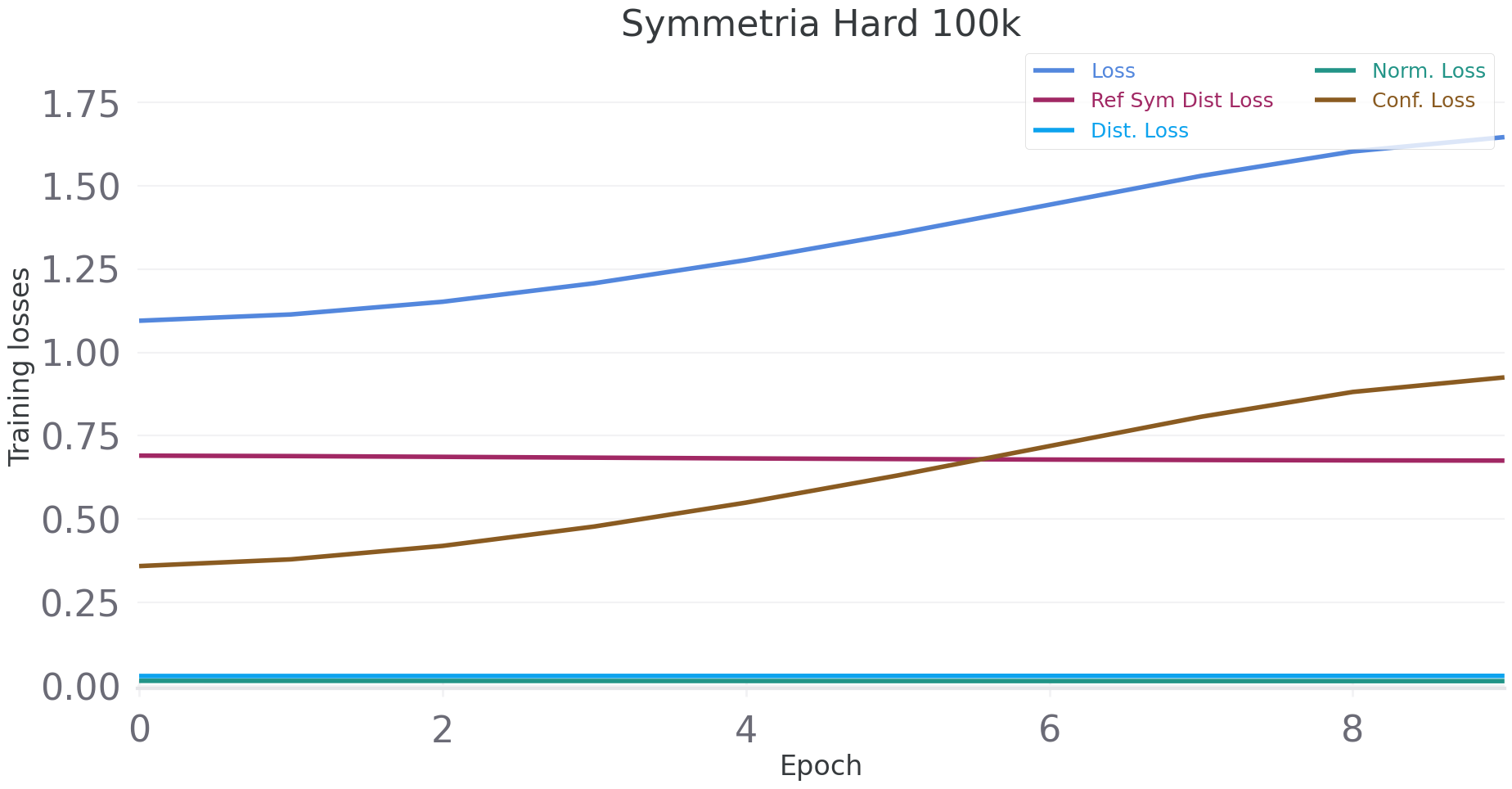} &
    \includegraphics[width=0.475\textwidth,trim={0.0cm 0cm 0.0cm 0cm},clip]{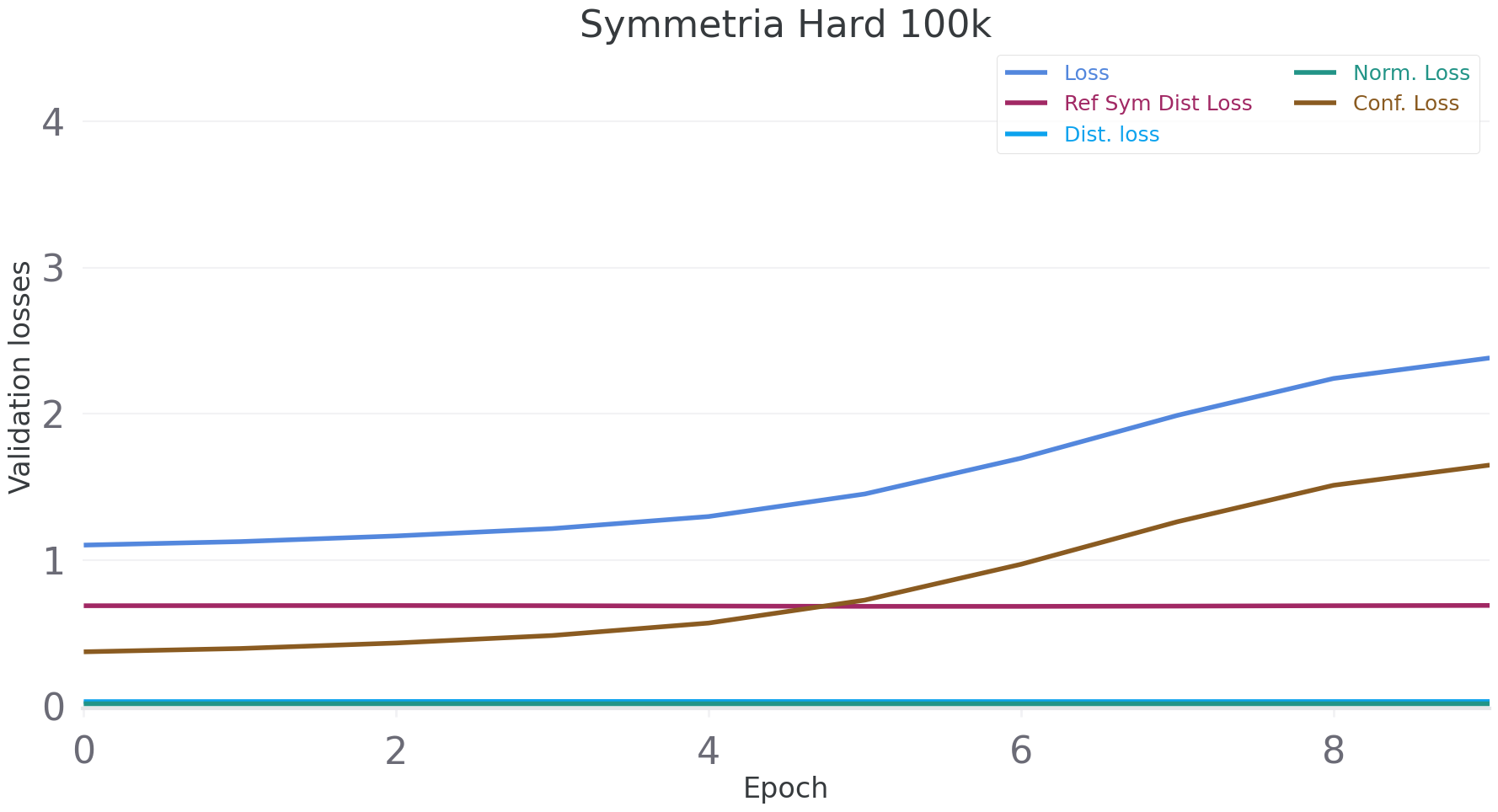} \\
    (g) & (h)  \\
\end{tabular}
\caption{Loss breakout curves for the different Symmetria-100k training/validation runs. The $x$-axis represents the number of epochs, while the $y$-axis shows the value of one of the five different losses composing the multitask learning process.}
\label{tab:results_loss_breakout_curves_100k}
\end{figure*}

To delve deeper into these learning behaviors, we analyze the constituent loss components of our multitask learning objective function (Section \ref{sec:sym_det_loss_decomposition}), visualized as breakout graphs in Figure \ref{tab:results_loss_breakout_curves_100k}.

These visualizations reveal the dominance of \textit{Reflection Symmetry Distance loss} and \textit{Confidence loss} in shaping the overall loss trajectory. For \textit{Easy}, \textit{Intermediate-1}, and \textit{Intermediate-2} models, at least one of these components consistently decreases over training epochs, contributing to the observed decrease in total loss and subsequent performance gains. In stark contrast, the \textit{Hard-100k} model exhibits a non-decreasing \textit{Reflection Symmetry Distance loss}, coupled with an increasing \textit{Confidence loss} from the first epoch onwards. This unfavorable trend suggests a struggle to minimize the designated loss components, leading to stagnation and even deterioration of performance metrics beyond the initial epoch.

\subsection{Information about the datasets and code}
In this section, we provide details about our datasets and source code provided as supplementary material of our paper. 

\subsubsection*{Access}
The proposed datasets are already available in the following link:
\url{http://deeplearning.ge.imati.cnr.it/symmetria/}. The source code and trained models for our experiments in self-supervised learning of point clouds can be found in this link: \url{https://github.com/ivansipiran/symmetria}. The source code and trained models for our experiments on symmetry detection can be found in this link: \url{https://github.com/QwagPerson/symmetria-symmetry-detection}.

The source code for the dataset generation is provided as supplementary material. This source code will be publicly available upon acceptance of our paper.

\subsubsection*{License}
The code and data generated in this paper are licensed under CC BY-NC-SA 4.0. This claim is already stated in the introduction of our paper.

\subsubsection*{Maintenance plan}
The datasets are hosted on a webpage at imati.cnr.it, which is the official domain of the IMATI institute. The project website will be hosted on that domain for a long time. In addition, the scripts that generate the datasets will be publicly available upon acceptance; therefore, anyone will be able to use the script to generate the dataset on demand.

\subsubsection*{Data format}
Our dataset is composed of files that store 3D point clouds. Each file contains a $N\times 3$ array with the point cloud. Additionally, we provide a Python class in our repository to create the Dataset and DataLoader in Pytorch. The Python class can be found here: \url{https://github.com/ivansipiran/symmetria/blob/main/datasets/SymmetryDataset.py}.

\end{document}